\documentclass{article}
\usepackage{ijcai18}

\usepackage{times}
\usepackage{xcolor}
\usepackage{soul}
\usepackage[utf8]{inputenc}
\usepackage[small]{caption}
\usepackage{algorithm}
\usepackage[noend]{algorithmic}
\usepackage{graphicx}
\usepackage{subfigure}
\usepackage{epstopdf}
\usepackage{verbatim}
\usepackage{amsmath}
\usepackage{float}
\usepackage{url}

\title{Towards Cooperation in Sequential Prisoner's Dilemmas: a
Deep Multiagent Reinforcement Learning Approach}

\author{
Weixun Wang$^1$,
Jianye Hao $^1$,
Yixi Wang $^1$,
Matthew Taylor$^2$,
\\
$^1$ Tianjin University, Tianjin, China \\
$^2$ Washington State University, Pullman, WA, USA\\
wxwang@tju.edu.cn,
jianye.hao@tju.edu.cn,
yixiwang2017@outlook.com,
taylorm@eecs.wsu.edu
}

\begin{document}

\maketitle

\begin{abstract}  
The Iterated Prisoner's Dilemma has guided research on social dilemmas for decades. However, it distinguishes between only two atomic actions: cooperate and defect. In real-world prisoner's dilemmas, these choices are temporally extended and different strategies may correspond to sequences of actions, reflecting grades of cooperation. We introduce a Sequential Prisoner's Dilemma (SPD) game to better capture the aforementioned characteristics. In this work, we propose a deep multiagent reinforcement learning approach that investigates the evolution of mutual cooperation in SPD games. Our approach consists of two phases. The first phase is offline: it synthesizes policies with different cooperation degrees and then trains a cooperation degree detection network. The second phase is online: an agent adaptively selects its policy based on the detected degree of opponent cooperation. The effectiveness of our approach is demonstrated in two representative SPD 2D games: the Apple-Pear game and the Fruit Gathering game. Experimental results show that our strategy can avoid being exploited by exploitative opponents and achieve cooperation with cooperative opponents.
\end{abstract}

\section{Introduction}
Learning is the key to achieving coordination with others in multiagent environments \cite{stone2000multiagent}. Over the last couple of decades, a large body of multiagent learning techniques have been proposed that aim to coordinate on various solutions (e.g., Nash equilibrium) in different settings, e.g., minimax Q-learning \cite{littman1994markov}, Nash Q-learning \cite{hu2003nash}, and Conditional-JAL \cite{banerjee2007reaching}, to name just a few.

One commonly investigated class of games is the Prisoner's Dilemma (PD), in which an Nash equilibrium solution is not a desirable learning target. Until now, a large body of work \cite{axelrod1984evolution,nowak1993strategy,banerjee2007reaching,crandall2005learning,damer2008achieving,hao2015introducing,mathieu2015new} has been devoted to incentivize rational agents towards mutual cooperation in repeated matrix PD games.
However, all the above works focus on the classic repeated PD games, which ignores several key aspects of real world prisoner's dilemma scenarios.
In repeated PD games, the moves are atomic actions and can be easily labeled as cooperative or uncooperative or learned from the payoffs \cite{busoniu2008comprehensive}. In contrast, in real world PD scenarios, cooperation/defection behaviors are temporally extended and the payoff signals are usually delayed (available after a number of steps of interactions).

Crandall \shortcite{crandall2012just} proposes the Pepper framework for repeated stochastic PD games (e.g., two-player gate entering problem), which can extend strategies originally proposed for classic repeated matrix games. Later some techniques extend Pepper in different scenarios, e.g., stochastic games with a large state space under tabular based framework \cite{elidrisi2014fast} and playing against the switching opponents \cite{Hernandez2017Towards}. However, these approaches rely  on hand-crafted state inputs and tabular Q-learning techniques to learn optimal policies. Thus, they cannot be directly applied to more realistic environments whose states are too large and complex to be analyzed beforehand.

Leibo et al.~\shortcite{leibo2017multi} introduce a 2D Fruit Gathering game to better capture the real world social dilemma characteristics, while also maintaining the characteristics of classical iterated PD games. In this game, at each time step, an agent selects its action based on its image observation and cannot directly observe the actions of the opponent. Different policies represent different levels of cooperativeness, which is a graded quantity. They investigate the cooperation/defection emergence problem by leveraging the power of deep reinforcement learning \cite{mnih2013playing,mnih2015human} from the descriptive point of view: how do multiple selfish independent agents' behaviors evolve when agents update their policy using deep Q-learning? In contrast, this paper takes a prescriptive and non-cooperative perspective and considers the following question: {\it how should an agent learn effectively in real world social dilemma environments when it is faced with different opponents?}

To this end, in the paper, we first formally introduce the general notion of sequential prisoner's dilemma (SPD) to model real world PD problems. We propose a multiagent deep reinforcement learning approach for mutual cooperation in SPD games. Our approach consists of two phases: offline and online phases. The offline phase generates policies with varying cooperation degrees and trains a cooperation degree detection network.  To generate policies,  we propose using the weighted target reward and two schemes, IAC and JAC, to train the baseline policies with varying cooperation degrees. Then we propose using the policy generation approach to synthesize the full range of policies from these baseline policies. Lastly, we propose a cooperation degree detection network implemented as an LSTM-based structure with an encoder-decoder module and generate a training dataset. The online phase extends the Tit-for-Tat principle into sequential prisoner's dilemma scenarios. Our strategy adaptively selects its policy with the proper cooperation degree from a continuous range of candidates based on the detected cooperation degree of an opponent. Intuitively, on one hand, our overall algorithm is cooperation-oriented and seeks for mutual cooperation whenever possible; on the other hand, our algorithm is also robust against selfish exploitation and resorts to a defection strategy to avoid being exploited whenever necessary. We evaluate the performance of our deep multiagent reinforcement approach using two 2D SPD games (the Fruit Gathering and Apple-Pear games).
Our experiments show that our agent can efficiently achieve mutual cooperation under self-play and also perform well against opponents with changing stationary policies.

\section{Background}

\subsection{Matrix Games and the Prisoner's Dilemma}

A matrix game can be represented as a tuple $ \langle N, \{A_i\}_{i \in N},\{R_i\}_{i \in N} \rangle$, where $N$ is the set of agents, $A_i$ is the set of actions available to agent $i$ with $\mathcal{A}$ being the joint action space $A_1 \times \ldots \times A_n$, and $R_i$ is the reward function for agent $i$. One representative class of matrix games is the prisoner's dilemma game, as shown in Table 1. In this game, each agent has two actions: cooperate (C) and defect (D), and is faced with four possible rewards: $R$, $P$, $S$, and $T$. The four payoffs satisfy the following four inequalities under a prisoner's dilemma game:
\begin{itemize}
\item $R > P$: mutual cooperation is preferred to mutual defection.
\item $R > S$: mutual cooperation is preferred to being exploited by a defector.
\item $2R > S + T$: mutual cooperation is preferred to an equal probability of unilateral cooperation and defection.
\item $T > R$: exploiting a cooperator is preferred over mutual cooperation.
\item $P > S$: mutual defection is preferred over being exploited.
\end{itemize}

\subsection{Markov Game}
Markov games combine matrix games and Markov Decision Processes and can be considered as an extension of Matrix games to multiple states.
A Markov game $\mathcal{M}$ is defined by a tuple $\langle N, S,\{A_i\}_{i \in N},\{R_i\}_{i \in N}, \mathcal{T} \rangle$, where $S$ is the set of states and $N$ is the number of agents, $\{A_i\}_{i\in N}$ is the collection of action sets, with $A_i$ being the action set of agent $i$, and $\{R_i\}_{i\in N}$ is the set of reward functions, $R_i: S \times A_1 \times \ldots \times A_n \rightarrow \mathcal{R}$ is the reward function for agent $i$. $\mathcal{T}$ is the state transition function: $S \times A_1 \times \ldots \times A_n \rightarrow \Delta (S)$, where $\Delta (S)$ denotes the set of discrete probability distributions over $S$. Matrix games are the special case of Markov games when $|S| = 1$.

\begin{table}
\centering
\caption{Prisoner's Dilemma}
\begin{tabular}{|c|c|c|}
\hline
\  & C & D  \\
\hline
C & R, R & S, T \\
\hline
D & T, S & P, P \\
\hline
\end{tabular}
\vspace{-10pt}
\end{table}

Next we formally introduce SPD by extending the classic iterated PD game to multiple states.

\subsection{Definition of Sequential Prisoner's Dilemma}
A two-player SPD is a tuple $\langle \mathcal{M}, \Pi \rangle$, where $\mathcal{M}$ is a 2-player Markov game with state space $S$. $\Pi$ is the set of policies with varying cooperation degrees. The empirical payoff matrix $(R(s), P(s), S(s), T(s))$ can be induced by policies $(\pi^C, \pi^D \in \Pi)$, where $\pi^C$ is more cooperative than $\pi^D$. Given two policies, $\pi^C$ and $\pi^D$, the corresponding empirical payoffs $(R, P, S, T)$ under any starting state s with respect to the payoff matrix in Section 2.1 can be defined as $(R(s), P(s), S(s), T(s))$ through their long-term expected payoff, where
\begin{eqnarray}
R(s) := V_1^{\pi^C, \pi^C} (s) & = V_2^{\pi^C, \pi^C}(s), \\
P(s) := V_1^{\pi^D, \pi^D} (s) & = V_2^{\pi^D, \pi^D}(s), \\
S(s) := V_1^{\pi^C, \pi^D} (s) & = V_2^{\pi^D, \pi^C}(s), \\
T(s) := V_1^{\pi^D, \pi^C} (s) & = V_2^{\pi^C, \pi^D}(s),
\end{eqnarray}
We can define the long-term payoff $V_i^{\vec \pi}(s)$ for agent $i$ when the joint policy $\vec \pi = (\pi_1, \pi_2)$ is followed starting from $s$.
\begin{eqnarray}
V_i^{\vec \pi}(s) = E_{\vec a_t \sim \vec \pi(s_t),s_{t+1} \sim \mathcal{T}(s_t,\vec a_t)}[\sum_{t=0}^{\infty} \gamma^t r_i(s_t,\vec a_t)]
\end{eqnarray}
A Markov game is an SPD when there exists a state $s \in S$ for which the induced empirical payoff matrix satisfies the five inequalities in Section 2.1. Since SPD is more complex than PD, the existing approaches addressing learning in matrix PD games cannot be directly applied in SPD.

\subsection{Deep Reinforcement Learning}

\textbf{Q-Learning and Deep Q-Networks}: Q-learning and Deep Q-Networks (DQN) \cite{mnih2013playing,mnih2015human} are value-based reinforcement learning approaches to learn optimal policies in Markov environments. Q-learning makes use of an action-value function for policy $\pi$ as $Q^{\pi}(s,a)=E_{s'}[r(s,a)+\gamma E_{a' \sim \pi}[Q^{\pi}(s',a')]]$. DQN uses a deep convolutional neural network to estimate Q-values and the optimal Q-values are learned by minimizing the following loss function:
\begin{eqnarray}
y = r + \gamma max_{a'}\bar{Q}(s',a'), \\
L(\theta) = E_{s,a,r,s'}[(Q(s,a|\theta)-y)^2]
\end{eqnarray}
where $\bar{Q}$ is a target $Q$ network whose parameters are periodically updated with the most recent $\theta$.

\textbf{Policy Gradient and Actor-Critic Algorithms}:  Policy Gradient methods are for a variety of RL tasks \cite{williams1992simple,sutton2000policy}. Their objective is to maximize $J(\theta)=E_{s \sim p^{\pi},a \sim \pi_{\theta}}[R]$ by taking steps in the direction of $\bigtriangledown _{\theta}J(\theta)$, where
\begin{eqnarray}
\bigtriangledown _{\theta}J(\theta) = E_{s \sim p^{\pi},a \sim \pi_{\theta}}[\bigtriangledown_{\theta}log\pi_{\theta}(a|s)Q^{\pi}(s,a)]
\end{eqnarray}
where $p^{\pi}$ is the state transition distribution. Practically, the value of $Q^{\pi}$ can be estimated in different ways. For example, $Q^{\pi}(s,a)$ serves as a critic to guide the updating direction of $\pi_{\theta}$, which leads to a class of actor-critic algorithms \cite{schulman2015high,wang2016sample}.

\section{Deep RL: Towards Mutual Cooperation}

Algorithm 1 describes our deep multiagent reinforcement learning approach, which consists of two phases, as discussed Section 1.
In the offline phase, we first seek to generate policies with varying cooperation degrees. Since the number of policies with different cooperation degrees is infinite, it is computationally infeasible to train all the policies from scratch. To address this issue, we first train representative policies using Actor-Critic until convergence (i.e., cooperation and defection baseline policies) (Lines \ref{alline3}-\ref{alline5}) detailed in Section 3.1; second, we synthesize the full range of policies (Lines \ref{alline6}-\ref{alline7}) from the above baseline policies, which will be detailed in Section 3.2. Another task is to how to effectively detect the cooperation degree of the opponent. We divide this task into two steps: we first train an LSTM-based cooperation degree detection network offline (Lines \ref{alline8}-\ref{alline10}), which will be then used for real-time detection during the online phase, detailed in Section 3.3. In the online phase, our agent plays against any opponent by reciprocating with a policy of a slightly higher cooperation degree than that of the opponent we detect (Lines \ref{alline11}-\ref{alline18}), detailed in Section 3.4.
Intuitively, on one hand, our algorithm is cooperation-oriented and seeks for mutual cooperation whenever possible; on the other hand, our algorithm is also robust against selfish exploitation and resorts to defection strategy to avoid being exploited whenever necessary.

\begin{algorithm}[tp]
\caption{The Approach of Deep Multiagent Reinforcement Learning Towards Mutual Cooperation}
\begin{algorithmic}[1]
\STATE \emph{//offline training}
\STATE initialize the size $N_{t}$ of training policy set, the size $N_{g}$ of generation policy set, the number $N_{d}$ of training data set,
	 the episode number $N_e$ and the step number $N_r$ of each episode
\FOR {training policy set index t = 1 to $N_{t}$} \label{alline3}
\STATE set agents' attitudes
\STATE train agents' policy set $P_{t}$ using weighted target reward\label{alline5}
\ENDFOR
\FOR {generation policy set index g = 1 to $N_{g}$} \label{alline6}
\STATE use policy set $\{P_{t}\}_{t \in N_t}$ to generate policy set $P_{g}$ \label{alline7}
\ENDFOR
\FOR {training data set index d = 1 to $N_{d}$ } \label{alline8}
\STATE generate training data set $D_{d}$ as $\{P_{t}\}_{t \in N_t} \cup \{P_{g}\}_{g \in N_g}$
\ENDFOR
\STATE use data set $\{D_{d}\}_{d \in N_d}$ to train cooperation degree detection network \label{alline10}
\STATE \emph{//adjust the policy online}
\STATE initialize $agent_1's$ cooperation degree \label{alline11}
\FOR {episode index e = 1 to $N_e$} \label{alline13}
\FOR {step index r = 1 to $N_r$}
\STATE $agent_1$ and $agent_2$ take actions and get rewards
\STATE $agent_1$ uses $n$-state trajectory $\langle s_{r-n+1}, \ldots, s_r \rangle$ to detect the cooperation degree $cd_2^r$ of $agent_2$
\STATE $agent_1$ updates $cd_1^r$ incrementally based on $cd_2^r$
\STATE $agent_1$ synthesizes a policy with $cd_1^r$ using policy generation \label{alline18}
\ENDFOR
\ENDFOR
\end{algorithmic}
\end{algorithm}

\subsection{Train Baseline Policies with Different Cooperation Degrees}

One way of generating policies with different cooperation degrees is by directly changing the key parameters of the environments. For example, Leibo et al.~\cite{leibo2017multi} investigate the influence of the resource abundance degree on the learned policy's cooperation tendency in sequential social dilemma games where agents compete for limited resources. It is found that when both agents employ deep Q-learning algorithms, more cooperative behaviors can be learned when resources are plentiful and vice versa. We may leverage similar ideas of modifying game settings to generate policies with different cooperation degrees. However, this type of approach requires a perfect understanding of the environment as a prior, and also may not be practically feasible when on cannot modify the underlying game engine.

Another more generalized way of generating policies with different cooperation degrees is to modify agents' reward signals during learning. Intuitively agents with the reward of the sum of all agents' immediate rewards would learn towards cooperation policies to maximize the expected accumulated social welfare eventually, and agents maximizing only their own reward would learn more selfish (defecting) policies. Formally, for a two-player environment (agent $i$ and $j$), agent $i$ computes a weighted target reward $r_i'$ as follows:
\begin{eqnarray}
r_i' = r_i + att_{ij} \times r_j
\end{eqnarray}
where $att_{ij} \in [0,1]$ is agent $i$'s attitude towards agent $j$, which reflects the relative importance of agent $j$ in agent $i's$ perceived reward. By setting the values of $att_{ij}$ and $att_{ji}$ to 0, agents would update their strategies in the direction of maximizing their own accumulated discounted rewards. By setting the values of $att_{ij}$ and $att_{ji}$ to 1, agents would update their strategies in the direction of maximizing the overall accumulated discounted reward. The greater the value of agent one's attitude towards agent two is, the higher the cooperation degree of agent one's learned policy would be.

\begin{figure}[t]
\centering
\includegraphics[height=1.5in,width=3.5in,angle=0]{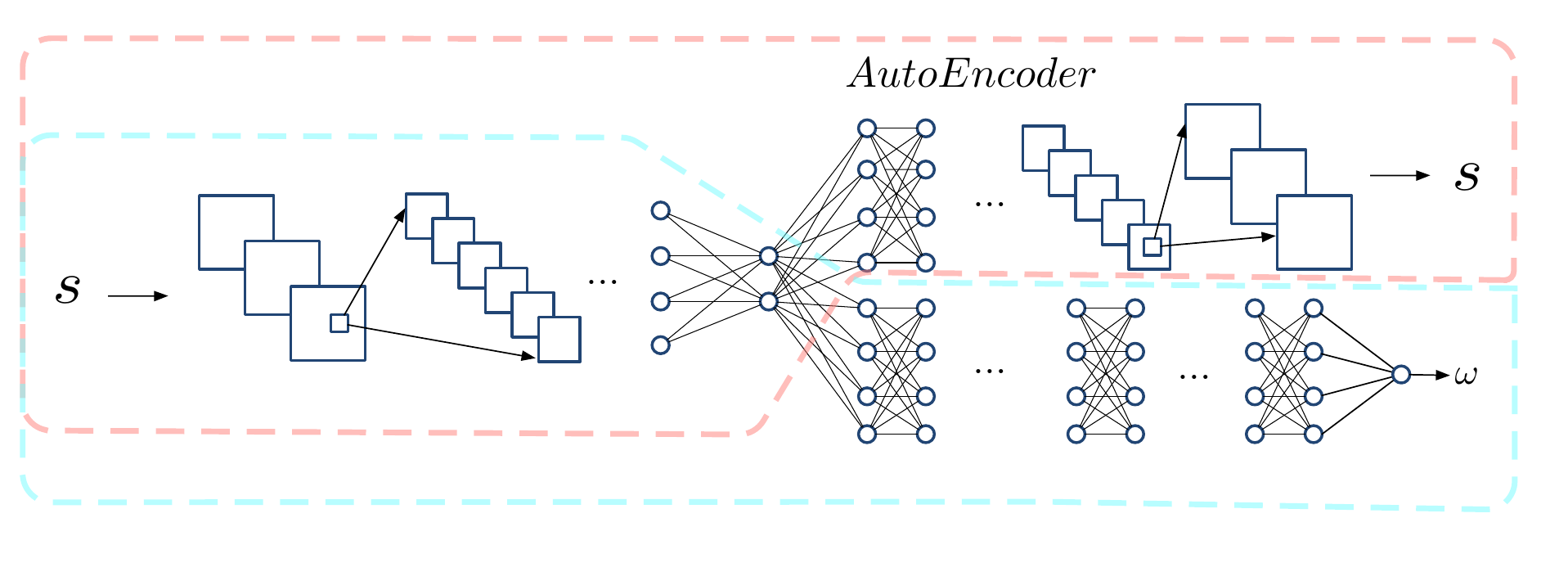}
\caption{ Cooperation Degree Detection Network }
\label{detectionnetwork}
\end{figure}

Given the modified reward signal for each agent, the next question is how agents should learn to effectively converge to the expected behaviors. One natural way is to resort to the the independent Actor-Critic (IAC) or some other independent deep reinforcement learning, i.e., equipping each agent with an individual deep Q-learning algorithm (IDQL) with the modified reward. However, the key element to the success of DQL,  experience replay memory, might prohibit effective learning in deep multiagent Q-learning environments \cite{foerster2017stabilising,sunehag2017value}. The nonstationarity introduced by the coexistence of multiple IACs means that data in the replay memory may no longer reflect the current dynamics in which the agent is learning. Thus IACs may frequently get confused by obsolete experience and impede the learning process. A number of methods have been proposed to remedy this issue \cite{foerster2017stabilising,lowe2017multi,foerster2017counterfactual} and we omit the details which are out of the scope of this paper.

Since the baseline policy training step is performed offline, another way of improving training is to use the joint Actor-Critic (JAC) by treating both agents as a single learner during training.
Note that we use JAC only for the purpose of training baseline policies offline,
and we do not require that we can control the policies that an opponent may use online.
In JAC, both agents share the same underlying network that learns the optimal policy over the joint action space using a single reward signal. In this way, the aforementioned nonstationarity problem can be avoided. Besides, compared with IAC, the training efficiency can be improved significantly since the network parameters are shared across agents in JAC.

In JAC, the weighted target reward is defined as follows:


\begin{eqnarray}
\label{attequation}
r_{ { \emph{total} } } = \sum_{i \in N} att_{i} \times r_i
\end{eqnarray}

where $att_i\in [0,1]$ represents the relative importance of agent $i$ on the overall reward. The smaller the value of $att_i$, the higher the cooperation degree of agent $i$'s learned policy and vice versa. Given the learned joint policy $\pi^{joint}(a_1,a_2|s,att_1,att_2)$, agent $i$ can easily obtain its individual policy $\pi_i$ as follows:


\begin{eqnarray}
\pi_i = \sum_{a_j,~j \neq i}\pi^{{\emph{joint}}}(a_1,a_2|s,att_1,att_2)
\end{eqnarray}

As we mentioned previously, it is computationally prohibitive to train a large number of policies with different cooperation degrees due to the high training cost of deep Q-learning, and because the policy space is infinite. To alleviate this issue, here we propose that only two policies, cooperation policy $\pi_c$ and defection policy $\pi_d$, need to be trained. Other policies with cooperation degree between the baselines can be synthesized efficiently, which will be introduced in Section \ref{policygeneration}.

\begin{figure}[t]
\includegraphics[height=2.0in,width=3.5in,angle=0]{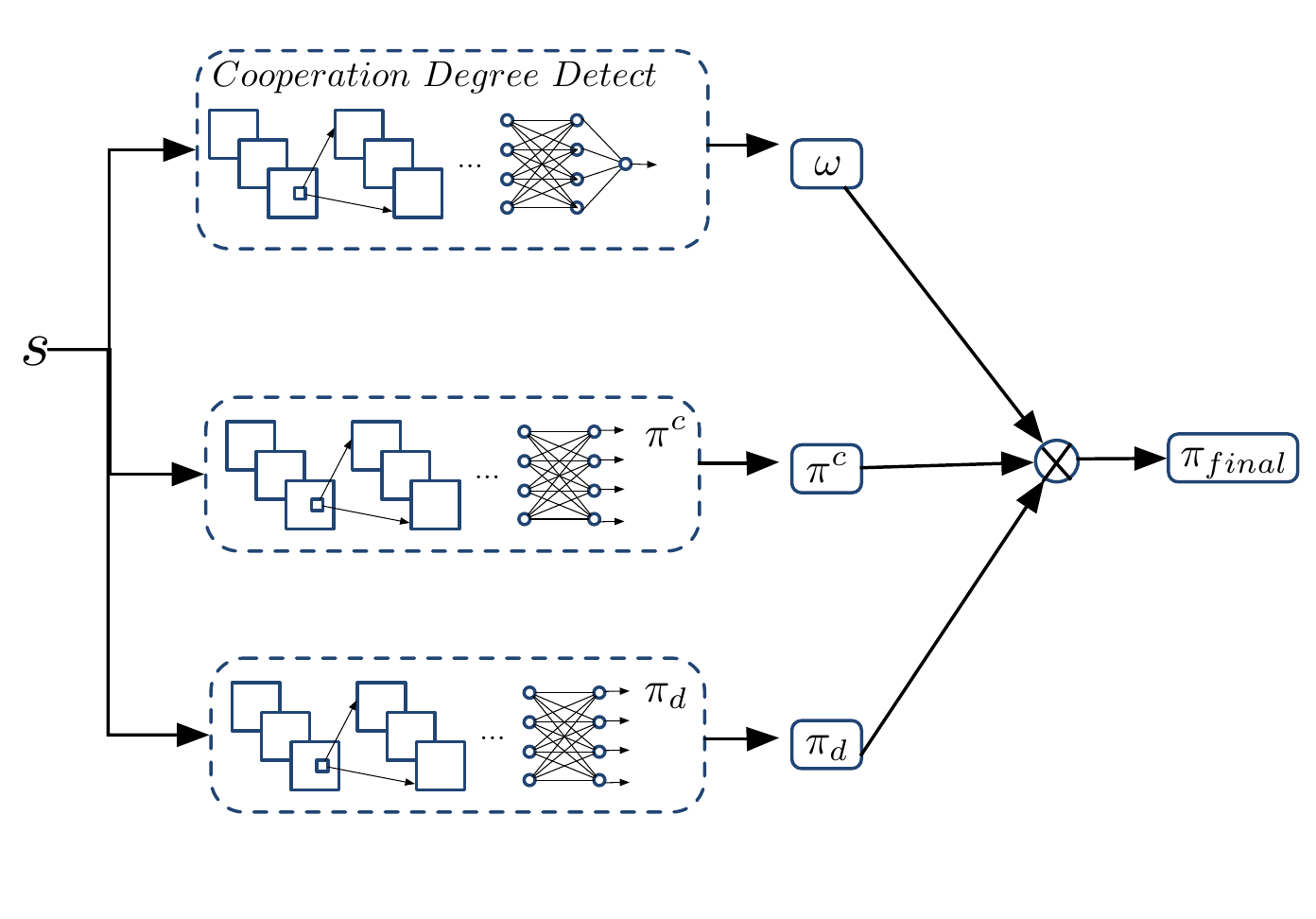}
\caption{ The Structure of Deep Reinforcement Learning
Approach Towards Mutual Cooperation }
\label{deepStructure}
\end{figure}
\subsection{Policy Generation}
\label{policygeneration}

Given the baseline policies $\pi_i^c$ and $\pi_i^d$, we synthesize multiple policies. Each continuous weighting factor $w_c\in [0,1]$ corresponds to a new policy $\pi_i^{w_c}$ defined as follows:
\begin{eqnarray}
\label{policygenerationequation}
\pi_i^{w_c} = w_c \times \pi_i^c + (1 - w_c) \times \pi_i^d
\end{eqnarray}
The weighting factor $w_c$ is defined as policy $\pi_i^{w_c}$'s cooperation degree. Specially, the linear combination of two policies has two advantages --- it 1) generates policies with varying cooperation degrees and 2) ensures low computational cost. Any synthesized policy $\pi_i^{w_c}$ should be more cooperative than $\pi_i^d$ and more defecting than $\pi_i^c$. The higher the value of $w_c$ is, the more cooperative the corresponding policy $\pi_i^{w_c}$ is. It is important to mention that the cooperation degrees of synthesized policies are ordinal, i.e., the cooperation degree of policies only reflect their relative cooperation ranking. For example, considering two synthesized policies $\pi_i^{0.6}$ and $\pi_i^{0.3}$, it only implies that policy $\pi_i^{0.6}$ is more cooperative than policy $\pi_i^{0.3}$. However, we cannot say that $\pi_i^{0.6}$ is twice as cooperative as $\pi_i^{0.3}$. Our way of synthesizing new policies can be understood as synthesizing policies over expert policies \cite{he2016opponent}. The previous work applies a similar idea to generate policies to better respond with different opponents in competitive environments, however, our goal here is to synthesize policies with varying cooperation degrees in sequential Prisoner's dilemmas.

\subsection{Opponent Cooperation Degree Detection}
\begin{figure}[t]
\includegraphics[height=0.9in,width=1.8in,angle=0]{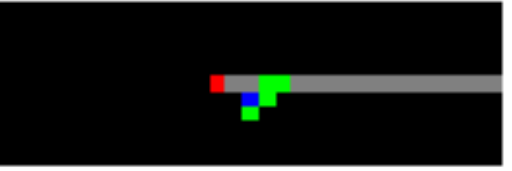}
\includegraphics[height=0.9in,angle=0]{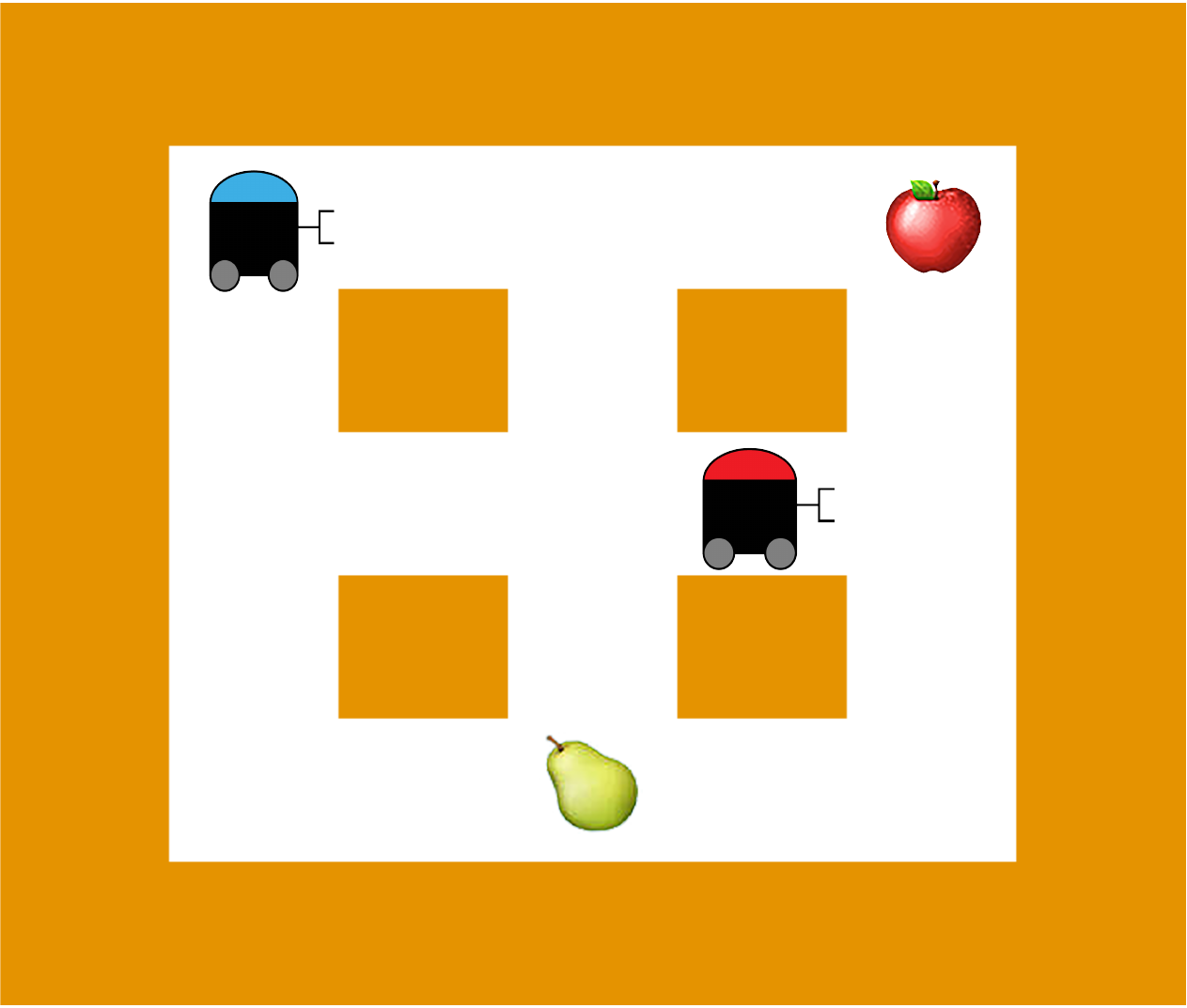}
\caption{The Fruit Gathering game (left) and the Apple-Pear game. }
\label{game}
\end{figure}

In classical iterated PD games, a number of techniques have been proposed to estimate the opponent's cooperation degree, e.g., counting the cooperation frequency when action can be observed; using a partial filter or a Bayesian approach otherwise \cite{damer2008achieving,hernandez2016bayesian,Hernandez2016Identifying,leibo2017multi}. However, in SPD, the actions are temporally extended, and the opponent's information (actions and rewads) cannot be observed directly. Thus the previous works cannot be directly applied. We need a way of accurately predicting the cooperation degree of the opponents from the observed sequence of moves in a qualitative manner. In SPD, given the sequential observations (time-series data), we propose an LSTM-based cooperation degree detection network.

In previous sections, we have introduced a way of synthesizing policies of any cooperation degree, thus we can easily prepare a large dataset of agents' behaviors with varying cooperation degrees. Based on this, we can transform the cooperation degree detection problem into a supervised learning problem: given a sequence of moves of an opponent, our task is to detect the cooperation degree (label) of this opponent. We propose a recurrent neural network, which combines an autoencoder and a recurrent classifier, as shown in Figure~\ref{detectionnetwork}.  Combing an autoencoder with a recurrent classifier brings two major benefits here. First, the classifier and autoencoder share underlying layer parameters of the neural network. This ensures that the classification task is based on the effective feature extraction of the observed moves, which improves the classification detection accuracy. Second, concurrent training of the autoencoder also helps to accelerate the training speed of the classifier and reduces fluctuation during training.

The network is trained on experiences collected by agents. Both agents $i$ and $j$ interact with the environment starting with initialized policies $\pi_i$ and $\pi_j$, yielding a training set $D$:

\begin{figure}[tp]
\centering
\subfigure {
\includegraphics[height=1.3in,width=1.7in,angle=0]{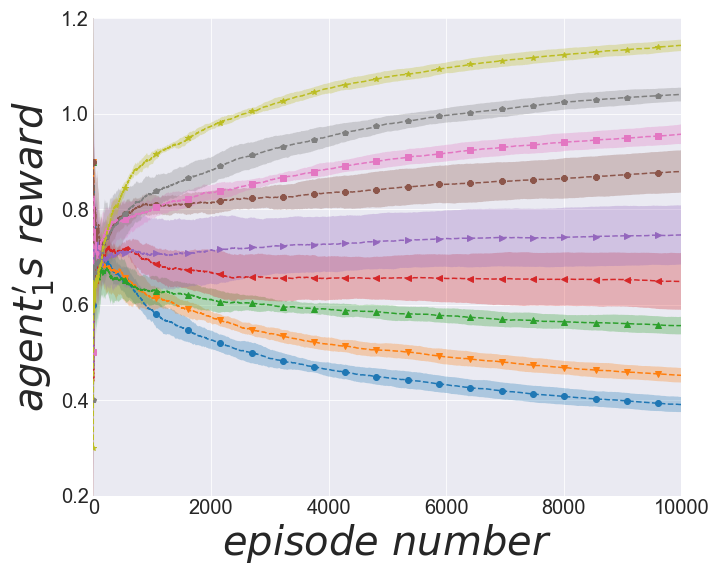}
\includegraphics[height=1.3in,width=1.7in,angle=0]{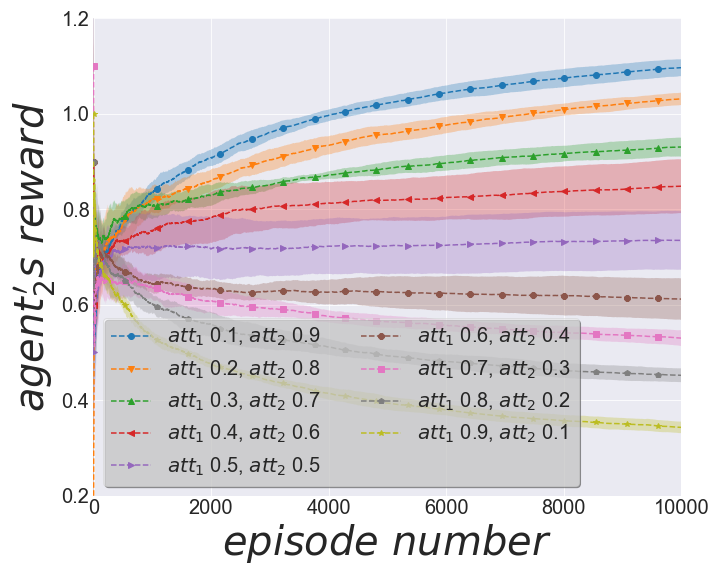}
}
\caption{ Agents' average rewards under policies trained with different cooperation attitudes. }
\label{baseline}
\end{figure}

\begin{equation}
\begin{split}
D = \{(X, label) | X = env(\pi_i,\pi_j),\\
\pi_i \in \{\pi_i^c, \pi_i^d\},\pi_j \in \Pi_j,\\
\mbox{if} \ \pi_i = \pi_i^c, \ label = 1 \ \mbox{else} \ 0 \}_{i \in N}
\end{split}
\end{equation}

where $\pi_i^c$ and $\pi_i^d$ are baseline policies for agent $i$. $\Pi_j$ is the learned policy set of its opponent $j$. $label$ is the relative cooperation degree of policy $\pi_i$, and $X$ is the set of trajectories under the joint policy $(\pi_i,\pi_j)$. For each trajectory $x = \langle s_1,s_2,\ldots,s_d \rangle \in X$, its label is the cooperation degree of agent $i$'s policy. The network is trained to minimize the following weighted cross entropy loss function as follows:

\begin{eqnarray}
\label{cd}
\begin{split}
L(f_c(\vec x),label) = - (w_1 \times label_1 \times log(p_1) + \\
w_2 \times (1 - lable_1) \times log(1 - p_1))
\end{split}
\end{eqnarray}
where $w_1$ is the weight of $label_1$. $p_1$ is the network output, which is the probability of $\pi^c$.

\subsection{Play Against Different Opponents}
Once we have detected the cooperation degree of the opponent, the final question arises as to how an agent should select proper policies to play against that opponent.  A self-interested approach would be to simply play the best response policy toward the detected policy of the opponent. However, as we mentioned before, we seek a
solution that can allow agents to achieve cooperation while avoiding being exploited.

Figure~\ref{deepStructure} shows our overall approach playing with opponents towards mutual cooperation. At each time step $t$, agent $i$ uses its previous $n$-step sequence of observations (from time step $t-n$ to $t$) as the input of the detection network, and obtain the detected cooperation degree $cd_j^t$. However, the one-shot detection of the opponent's cooperation degree might be misleading due to either the detection error of our classifier or the stochastic behaviors of the opponent. Thus this may lead to high variance of our detection outcome and our response policy thereafter. To reduce the variance, agent $i$ uses the exponential smoothing to update its current cooperation degree estimation of its opponent $j$ as follows:
\begin{eqnarray}
cd_i^t = (1 - \alpha) \times cd_i^{t-1} + \alpha \times cd_j^t
\end{eqnarray}
where $cd_i^{t-1}$ is agent $i$'s cooperation degree in the last time step, and $\alpha$ is the cooperation degree changeing factor.

Finally, agent $i$ sets its own cooperation degree equal to $cd_i^t$ plus its reciprocation level, and then synthesizes a new policy with the updated cooperation degree following Equation (\ref{policygenerationequation}) as its next-step strategy to play against its opponent. Note that the reciprocation level can be quite low and still produce cooperation. The benefit is that it will not lead to a significant loss if the opponent is not cooperative at all, while full cooperation can be reached if the opponent reciprocates in the same way. Also, note that here we only provide a way of responding to the opponent with changing policies. However, our overall approach is general and any existing multiagent strategy selection approaches can be applied here.

\begin{figure*}[htbp]
\centering
\subfigure[The Apple-Pear game] {\includegraphics[height=1.5in,width=2.2in,angle=0]{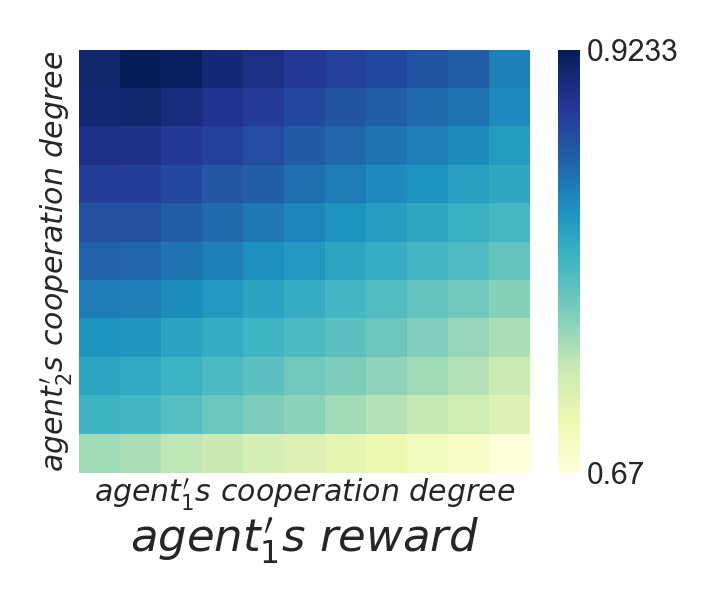}
\includegraphics[height=1.5in,width=2.2in,angle=0]{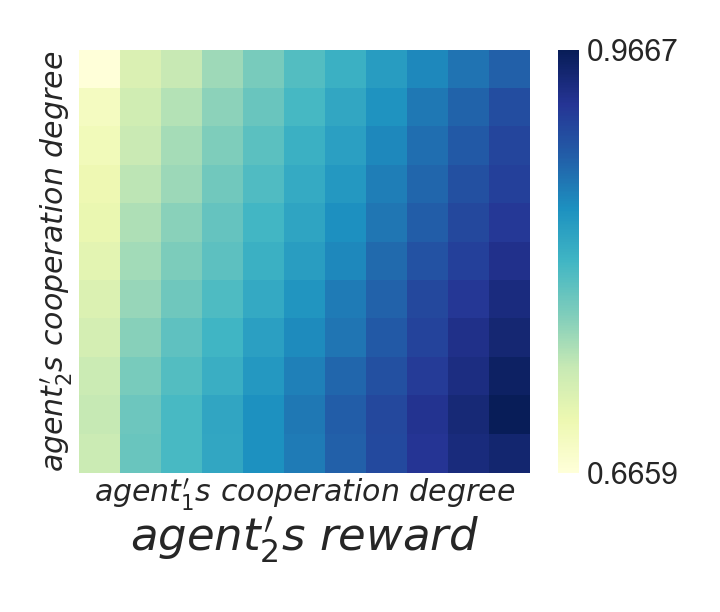}
\includegraphics[height=1.5in,width=2.2in,angle=0]{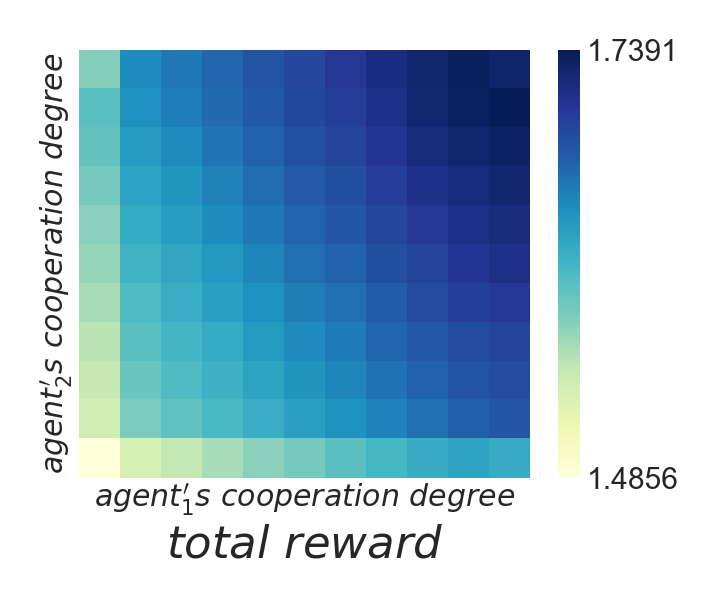}}

\subfigure[The Fruit Gathering game] {\includegraphics[height=1.5in,width=2.2in,angle=0]{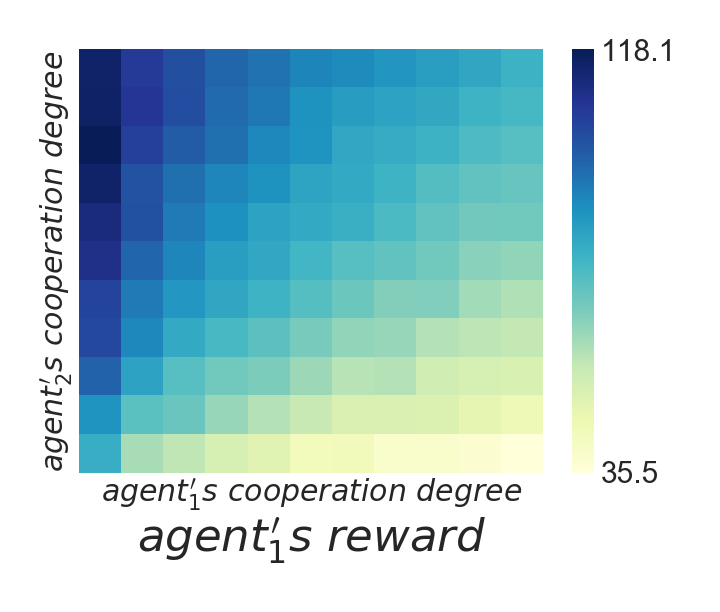}
\includegraphics[height=1.5in,width=2.2in,angle=0]{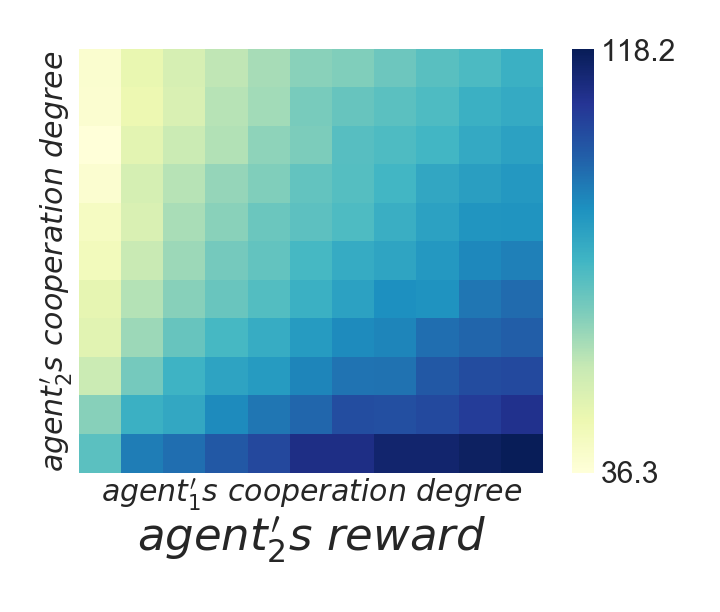}
\includegraphics[height=1.5in,width=2.2in,angle=0]{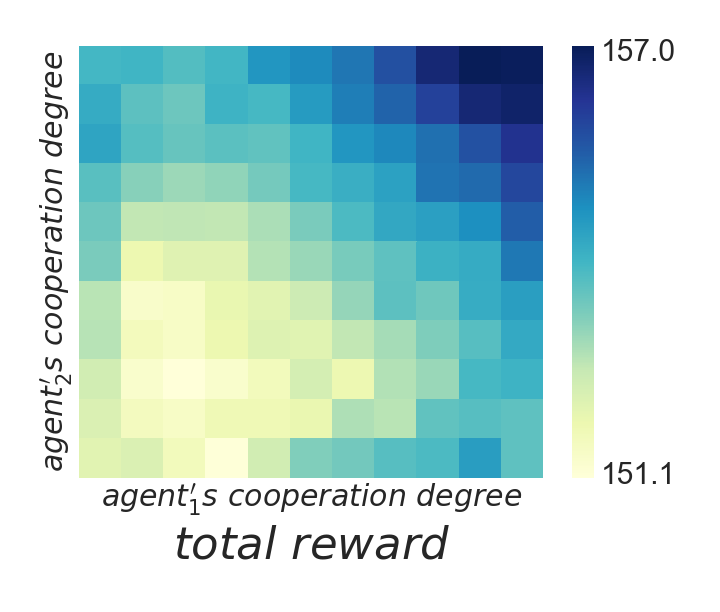}}

\caption{ Average and total rewards under different cooperation degrees. The cooperation degrees of $agent_1$ and $agent_2$ increase from left to right and from bottom to top respectively. Each cell corresponds the rewards of different policy pairs. }
\label{rewardheatmap}
\end{figure*}

\section{Simulation and Results}
\subsection{SPD Game Descriptions}
In this section, we adopt the Fruit Gathering game \cite{leibo2017multi} to evaluate the effectiveness of our approach. We also propose another game Apple-Pear, which also satisfies real world social dilemma conditions we mentioned before. Each game involves two agents (in blue and red). The task of an agent in the Fruit Gathering game is to collect as many apples, represented by green pixels, as possible (see Figure~\ref{game} (left)). An agent's action set is: step forward, step backward, step left, step right, rotate left, rotate right, use beam, and stand still.
The agent obtains the corresponding fruit when it steps on the same square as the fruit is located.
When an agent collects an apple, it will receive a reward 1. And the apple will be removed from the environment and respawn after 40 frames. Each agent can also emit a beam in a straight line along its current orientation. An agent is removed from the map for 20 frames if it is hit by the beam twice. Intuitively, a defecting policy in this game is one which frequently tags the rival agent to remove it from the game. A cooperation policy is one that rarely tags the other agent.
For the Apple-Pear game, there is a red apple and a green pear (see Figure~\ref{game} (right)). The blue agent prefers apple while the red agent prefers pear. Each agent has four actions: step right, step left, step backward, step forward, and each step of moving incurs a cost of 0.01.
The fruit is collected when the agent steps on the same square as it.
When the blue (red) agent collects an apple (pear) individually, it receives a higher reward 1. When the blue agent collects a pear individually, it receives a lower reward 0.5. The situation is the opposite for the red agent. One exception is that they both receive a half of their corresponding rewards when they share a pear or an apple. In this game, a fully defecting policy is to collect both fruits whenever the fruit-collecting reward exceeds the moving cost, while a cooperative one is to only collect the fruit it prefers to maximize the social welfare of agents. In Section 4.4, we find that the two games satisfy the definition of SPD games in Section 2.3 by using policies with different cooperation degrees to play with each other.

\subsection{Network Architecture and Parameter Settings}
In both games, our network architectures for training the baseline policies follow standard AC networks, except that we allow both actor and critic to share the same underlying network to reduce the parameter space. For the underlying network, the first hidden layer convolves $32$ filters of $8 \times 8$ with stride $4$ with the input image and applies a rectifier nonlinearity. The second hidden layer convolves $64$ filters of $4 \times 4$ with stride $2$, again followed by a rectifier nonlinearity. This is followed by a third convolutional layer that convolves $64$ filters of $3 \times 3$ with stride $1$ followed by a rectifier. For the actor, on the basis of sharing network, the next layer includes $128$ units with rectifier nonlinearity, and the final softmax layer has as many units as the number of actions. The critic is similar to the actor, but with only one scalar output.

The recurrent cooperation degree detection network is shown in Figure~\ref{detectionnetwork}. The autoencoder and the detection network share the same underlying network. The first hidden layer convolves $10$ filters of $3 \times 3$ with stride $2$ with the input image and applies a rectifier nonlinearity. The second hidden layer is the same as the first one and the third hidden layer convolves $10$ filters of $3 \times 3$ with stride $3$ and applies a rectifier nonlinearity. The autoencoder is followed by a fourth hidden layer that deconvolves $10$ filters of $3 \times 3$ with stride $3$ and applies a sigmoid and its output shape is $21 \times 21 \times 10$. The next layer deconvolves $10$ filters of $3 \times 3$ with stride $2$ and applies a sigmoid and the output shape is $42 \times 42 \times 10$. The final layer deconvolves $10$ filters of $3 \times 3$ with stride $2$ and applies a sigmoid and the output shape is $84 \times 84 \times 3$. Cooperation degree detection network is followed by two LSTM layers of $256$ units. The final layer is an output node.\footnote{The code and network architectures will be available soon: \url{https://goo.gl/3VnFHj}.}

For the Apple-Pear game, each episode has at most $100$ steps. The exploration rate is annealed linearly from $1$ to $0.1$ over the first $20000$ steps. The weight parameters are updated by soft target updates \cite{lillicrap2015continuous} every 4 steps to avoid the update fluctuation as follows:

\begin{eqnarray}
\theta' \leftarrow 0.05\theta + 0.95\theta'
\end{eqnarray}
where $\theta$ is the parameter of the policy network and $\theta'$ is the parameter of the target network. The learning rate is set to $0.0001$ and memory is set to $25000$. The batch size is $128$. For the loss function in the cooperation degree detection network, we set $w_1$ and $w_2$ as 1 and 2 (see Equation~\ref{cd}).
When agents play with different opponents online,
we assign the number of states visited to the length $n$ of
the state sequence which is the input of cooperation detection network.
The cooperation degree changing factor $\alpha$ is set to 1.

The Fruit Gathering game uses the same detection network architecture. The actor-critic uses independent policy networks. Each episode has at most $100$ steps during training. The exploration rate and the memory are the same as the Apple-Pear game. The weight parameters are updates the same as the Apple-Pear game:
\begin{eqnarray}
\theta' \leftarrow 0.001\theta + 0.999\theta'
\end{eqnarray}
When agents play with different opponents online, we set state sequence length $n$ as 50 and the changing factor $\alpha$ is set to 0.02.

\begin{figure*}[htbp]
\subfigure[Apple-Pear]{\includegraphics[height=1.5in,width=2.2in,angle=0]{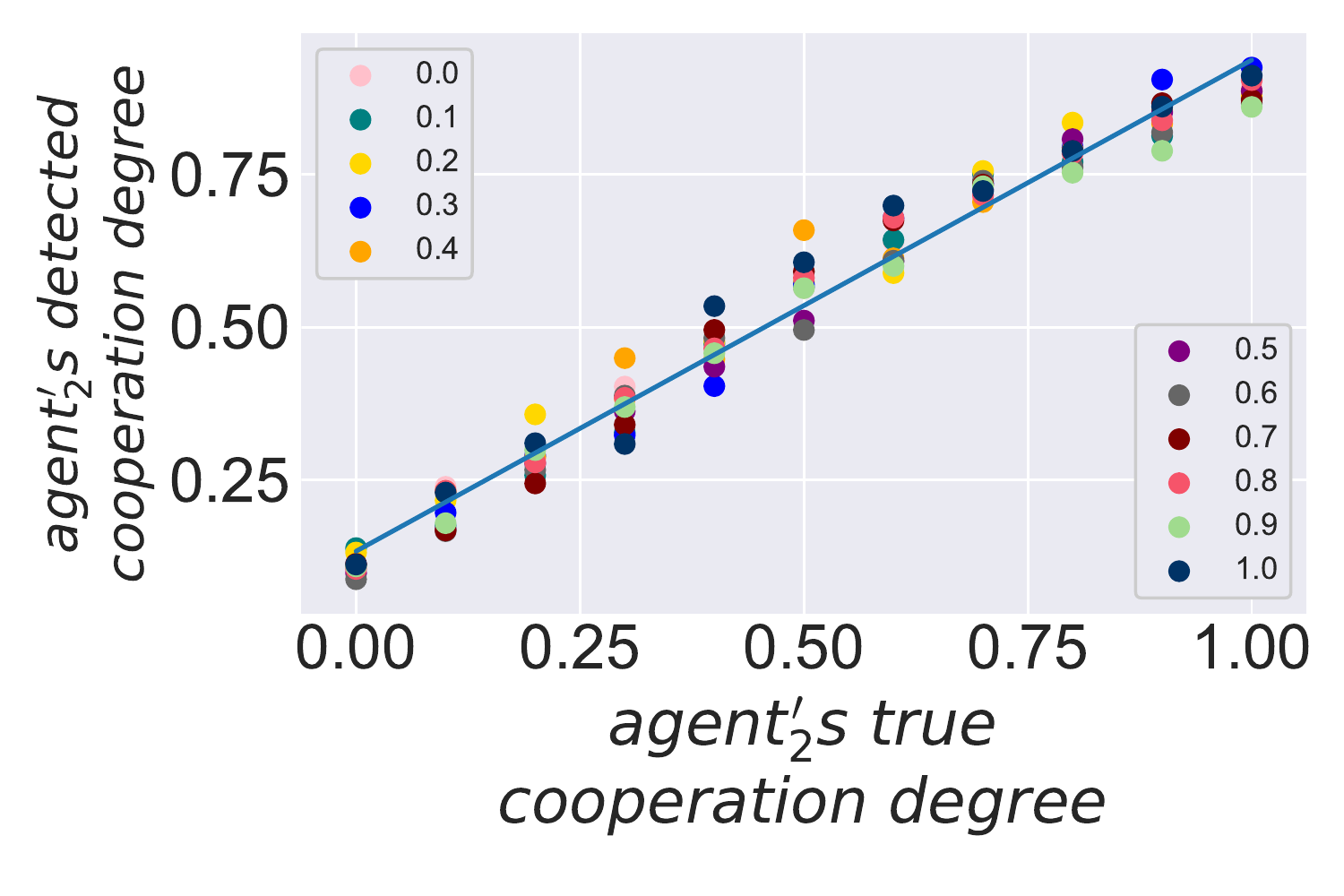}}
\subfigure[Fruit Gathering]{\includegraphics[height=1.5in,width=2.2in,angle=0]{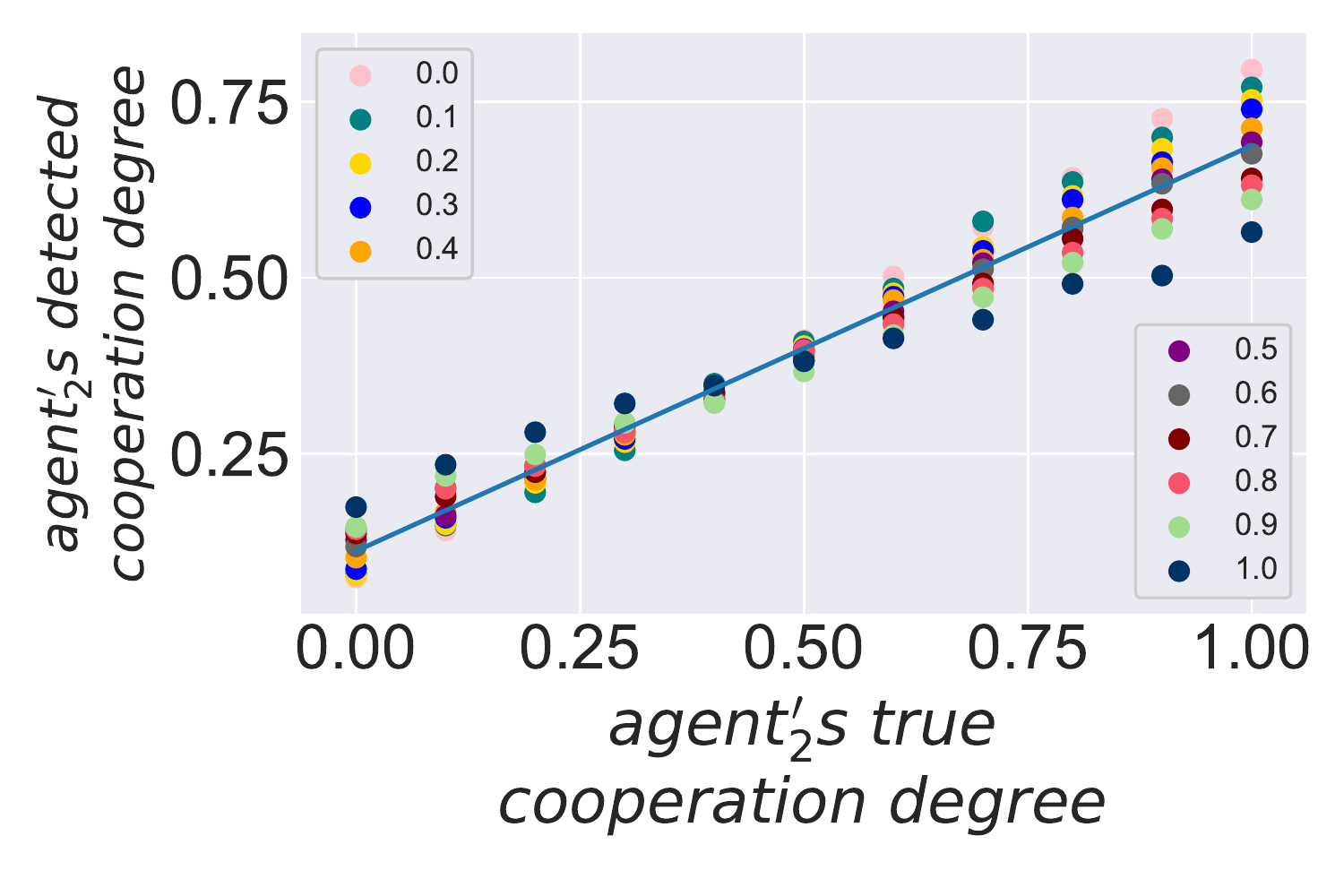}}
\subfigure[Fruit Gathering ($cd_{1} = 1$)] {\includegraphics[height=1.5in,width=2.2in,angle=0]{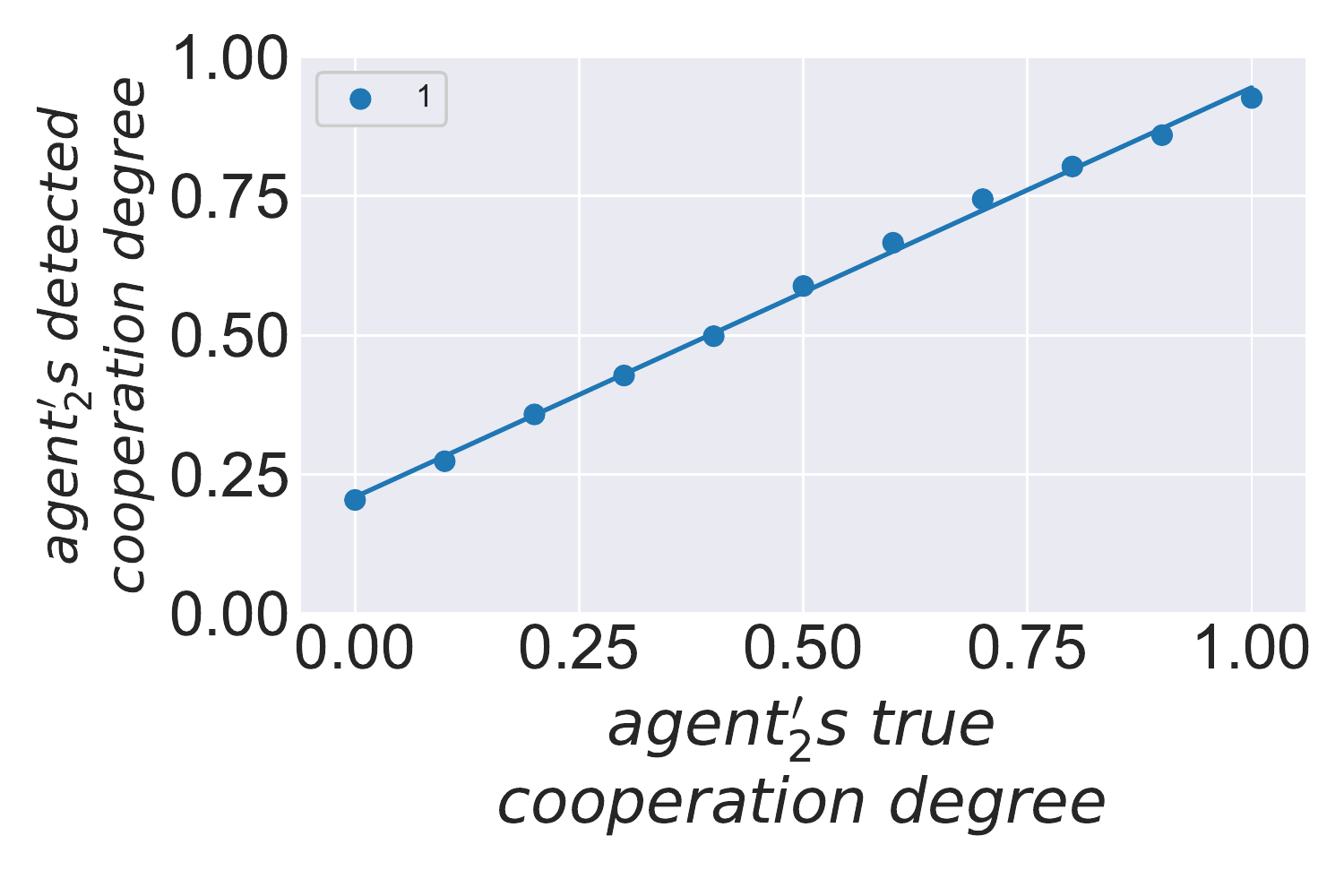}}
\caption{ The detection results for $agent_2$ under different cooperation degrees of $agent_1$: (a) Apple-Pear game; (b) Fruit Gathering game; (c)$agent_1$'s cooperation degree as 1 in Fruit Gathering game}
\label{detectionresult}
\end{figure*}

\subsection{Effect of Baseline Policy Generation}

For the Apple-Pear game, the baseline policies are trained using the JAC scheme. The individual rewards of each agent increase gradually as its attitude increases and the other agent's attitude decreases. The results also indicate that an agent can learn a defecting policy when its attitude
that represents its relative importance on overall reward
increases and vice versa. We set the attitude in Equation (\ref{attequation}) between 0.1 and 0.9 to train baseline policies. The reason that attitudes are not set 0 and 1 is that these settings will cause an agent with meaningless policy,
e.g, the agent whose attitude equals $0$ makes no influence to total reward in Equation (\ref{attequation}) and hence will always avoid collecting fruit.
This would, in turn, affect the learned policy quality of the other agent (i.e., lazy agent problem \cite{perolat2017multi}). Figure~\ref{baseline} shows the average rewards of agents under policies trained with different weighted target rewards. We also evaluate the IAC scheme and similar results can be obtained and we omit it here.

For the Fruit Gathering game, the learning policies are synthesized based on IAC. IAC is more efficient for training baseline policies in this game, relative to JAC, since the rewards of collecting the apple for both agents are the same. On the other hand, if we adopt a JAC approach, it might lead to the consequence that the agent with a higher weight will collect all the apples, while the other agent will not collect any apples. Similar results as the Apple-Pear game can be observed here and we omit it here.
\subsection{Effect of Policy Generation}

For the Apple-Pear game, the baseline cooperation policies $\pi_1^c$ and $\pi_2^c$ are trained under the setting of $\{att_1 = att_2 = 0.5\}$. For $agent_1$, the baseline defection policy $\pi_1^d$ has $\{att_1 = 0.75, att_2 = 0.25\}$. For $agent_2$, the baseline defection policy $\pi_2^d$ has $\{att_1 = 0.25, att_2 = 0.75\}$. Then we generate the policy set of $agent_1$ $\Pi_1 = \{\pi = w_1 \times \pi_1^c + (1 - w_1) \times \pi_1^d | w_1 = 0.0, 0.1, \ldots, 1.0\}$ and the policy set of $agent_2$ $\Pi_2 = \{\pi = w_2 \times \pi_2^c + (1 - w_2) \times \pi_2^d | w_2 = 0.0, 0.1, \ldots, 1.0\}$. After that, two policies $\pi_1$ and $\pi_2$ are sampled from $\Pi_1$ and $\Pi_2$ and matched against each other in the games for 200,000 episodes. The average rewards are assigned to individual cells of different attitude pairs, in which $\pi_1$ correspond to policies with varying cooperation degrees $w_1$ for agent $1$, and $\pi_2$, policies with varying cooperation degrees $w_2$ for agent $2$ (see Figure~\ref{rewardheatmap} (a)). In Figure~\ref{rewardheatmap} (a), we can observe that when agents' cooperation degrees decrease, their rewards decrease. When both of their cooperation degrees increase, which means they are more cooperative, the sum of their rewards increase. Besides, given a fixed cooperation degree of $agent_1$, $agent_2$'s reward is increased as its cooperation degree decreases, and vice versa. Similar pattern can be observed for agent 1 as well. Therefore, it indicates that we can successfully synthesize policies with a continuous range of cooperation degrees. It also confirms that the Apple-Pear game can be seen as an SPD following the definition in Section 2.3.

For the Fruit Gathering game, we use policies with both attitudes equal to 0 and 0.5 as the baseline defection and cooperation policy respectively. Figure~\ref{rewardheatmap} (b) shows the results of their rewards, which is similar to the Apple-Pear game.

\subsection{Effect of Cooperation Degree Detection}
This section evaluates the detection power of the cooperation degree detection network. First, we train the cooperation degree detection network using datasets which include only the data labeled as full cooperation or full defection. The training data $(\langle s_0, s_1, s_2, \ldots, s_n \rangle,label)$ in the simulation is obtained based on baseline policies of $agent_2$. We set its label as 1 when $agent_2$ uses its baseline cooperation policy and 0 when $agent_2$ uses its baseline defection policy. Then we use this dataset to train the detection network. After training we evaluate the detection accuracy by applying it to detecting the cooperation degree of agent 2 when it uses policies with cooperation degrees from 0 to 1 and the degree interval is 0.1. The policy pair $(\pi_1, \pi_2)$ is sampled from $\Pi_1$ and $\Pi_2$ and matched against with each other for each episode. After the cooperation degree average value is stable, we view the output value as the cooperation degree of $agent_2$.

For the Apple-Pear game, we collect 10000 data for state sequence lengths \{3, 4, 5, 6, 7, 8\}. The cooperation detection results of $agent_2$ are shown in Figure~\ref{detectionresult} (a). We can see that the cooperation detection values can well approximate the true values with slight variance. Besides, the network can clearly detect the order of different cooperation degrees, which allows us to calculate the cooperation degree of $agent_2$ accurately. For the Fruit Gathering game, we collect 4000 data for each state sequence lengths \{40, 50, 60, 70\}, which includes 2000 data labeled as 1 and 2000 data labeled as 0. The network is trained in a similar way and the cooperation degree detection accuracy is high (see Figure~\ref{detectionresult} (b)). From Figure~\ref{detectionresult}, We observe that the detection results are almost linear with the true values. Fig 6(c) shows the detection results of $agent_2$ when $agent_1$'s policy is fixed with cooperation degree 1. We can see that the true cooperation degree of $agent_2$ can be easily obtained by fitting a linear curve between the predicted and true values. Thus for each policy of $agent_1$ in $\Pi_1$, we can fit a linear function to evaluate the true cooperation degrees of $agent_2$. During practical online learning, when $agent_1$ uses policies of varying cooperation degrees to play with $agent_2$, it firstly chooses the function whose corresponding policy is closest to the used policy, and then computes the cooperation degree of $agent_2$.

\begin{figure}[htbp]
\centering
\subfigure {\includegraphics[height=1.2in,width=1.7in,angle=0]{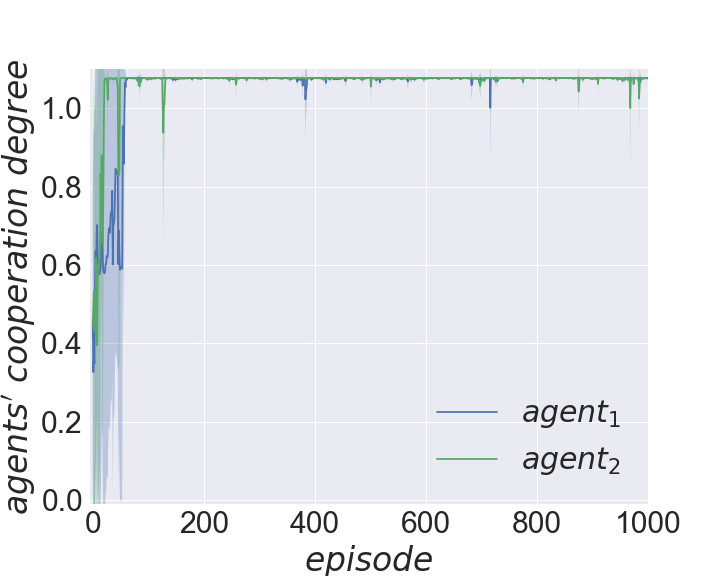}
\includegraphics[height=1.2in,width=1.7in,angle=0]{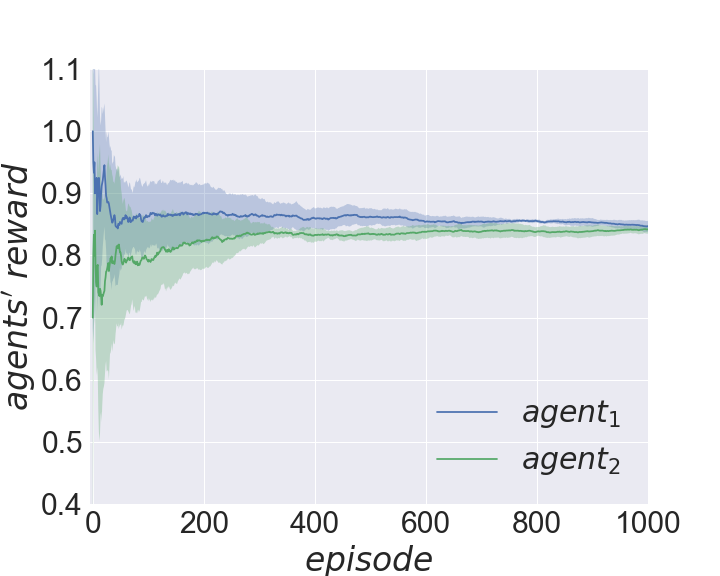}}
\caption{ Performance under self-play in the Apple-Pear game when both agents use our strategy. }
\label{aselfplay}
\end{figure}

\begin{figure}[htbp]
\centering
\subfigure {\includegraphics[height=1.2in,width=1.8in,angle=0]{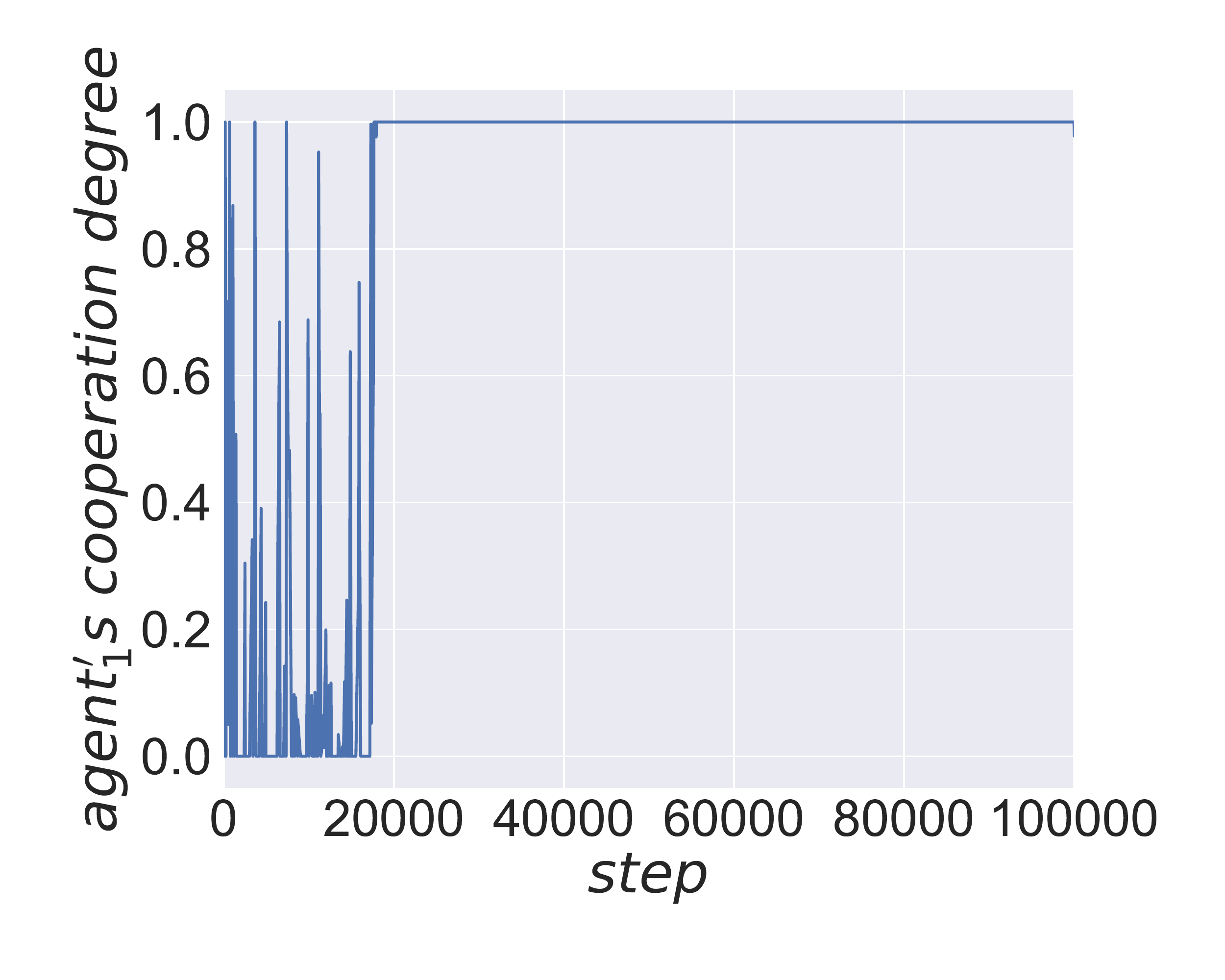}
\includegraphics[height=1.2in,width=1.8in,angle=0]{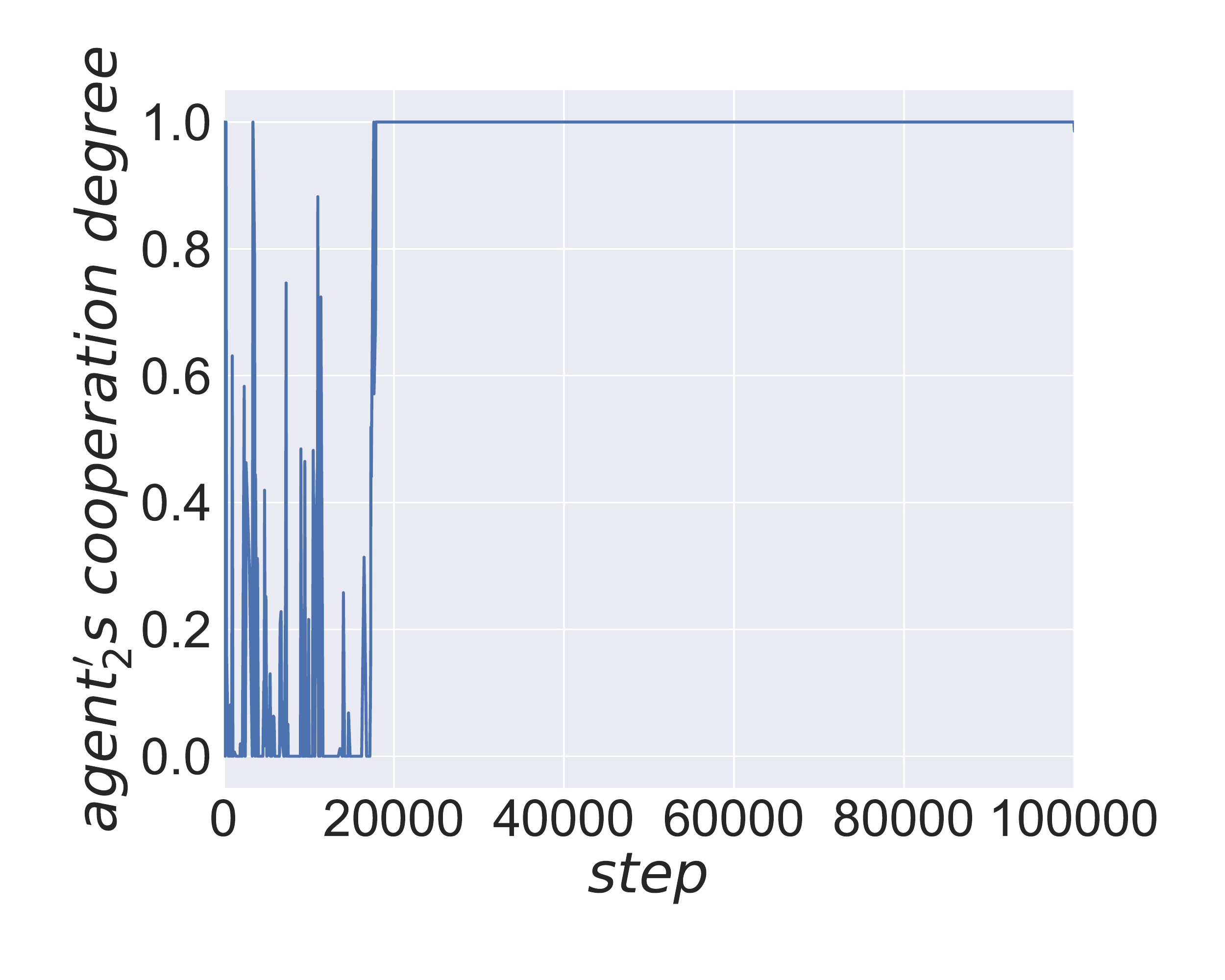}}
\caption{ Performance under self-play in the Fruit Gathering game when both agents use our strategy. }
\label{gselfplay}
\end{figure}

\begin{figure*}[htbp]
\centering
\subfigure{\includegraphics[height=1.225in,width=1.75in,angle=0]{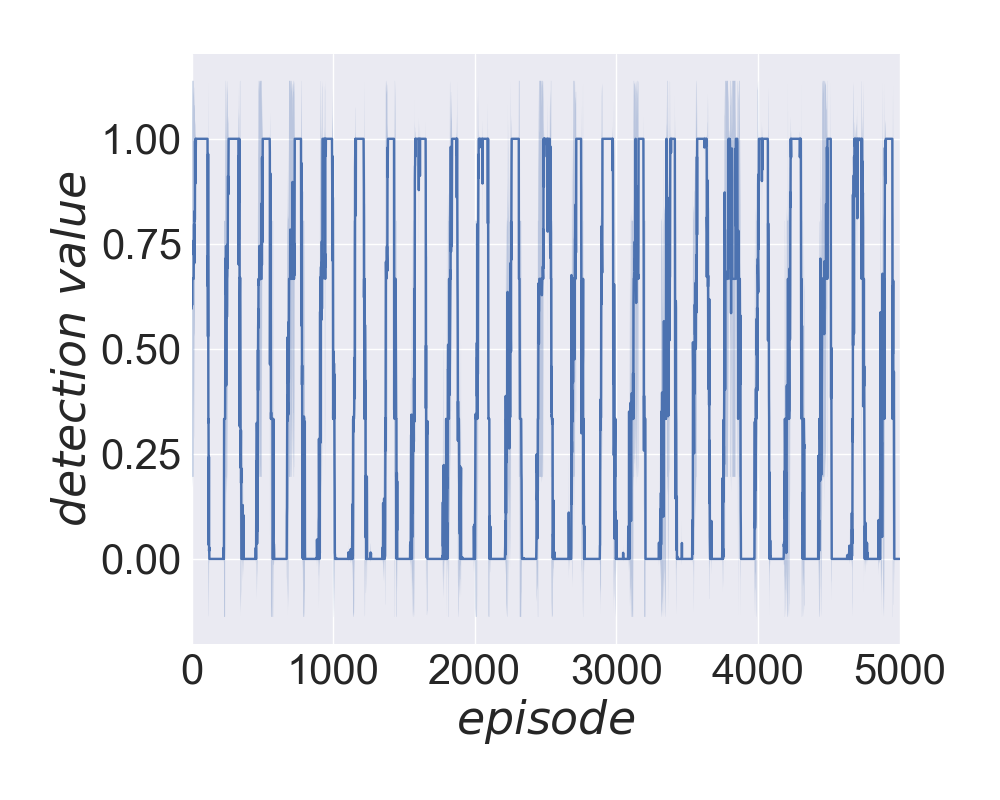}
\includegraphics[height=1.225in,width=1.75in,angle=0]{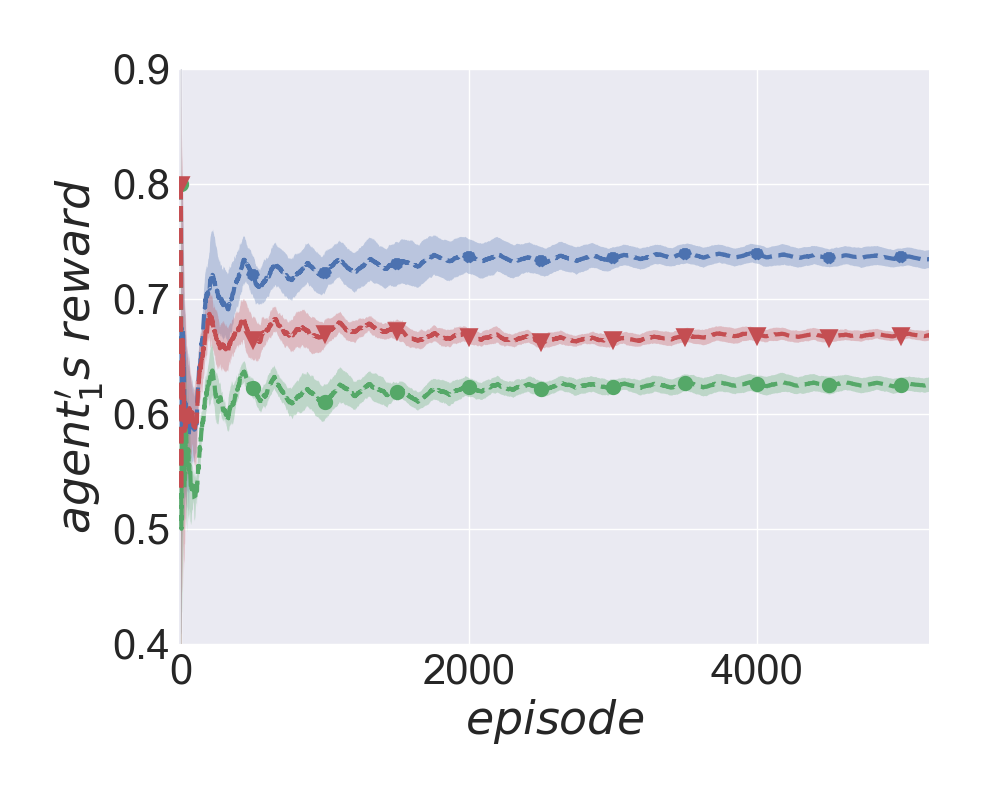}
\includegraphics[height=1.225in,width=1.75in,angle=0]{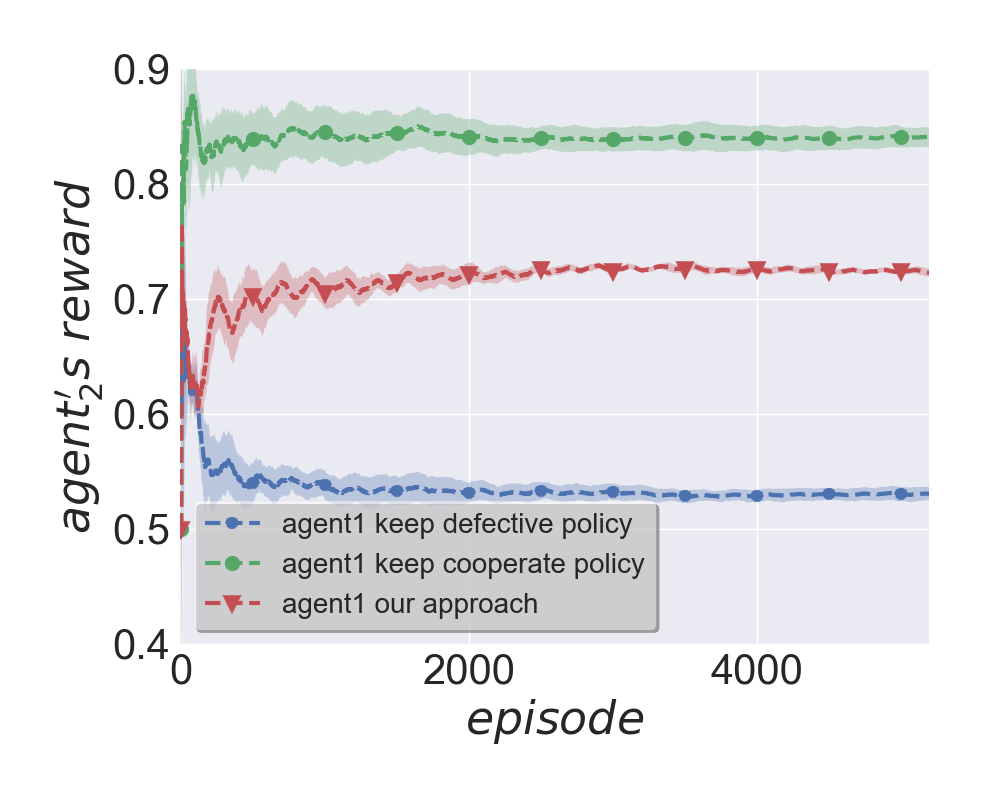}
\includegraphics[height=1.225in,width=1.75in,angle=0]{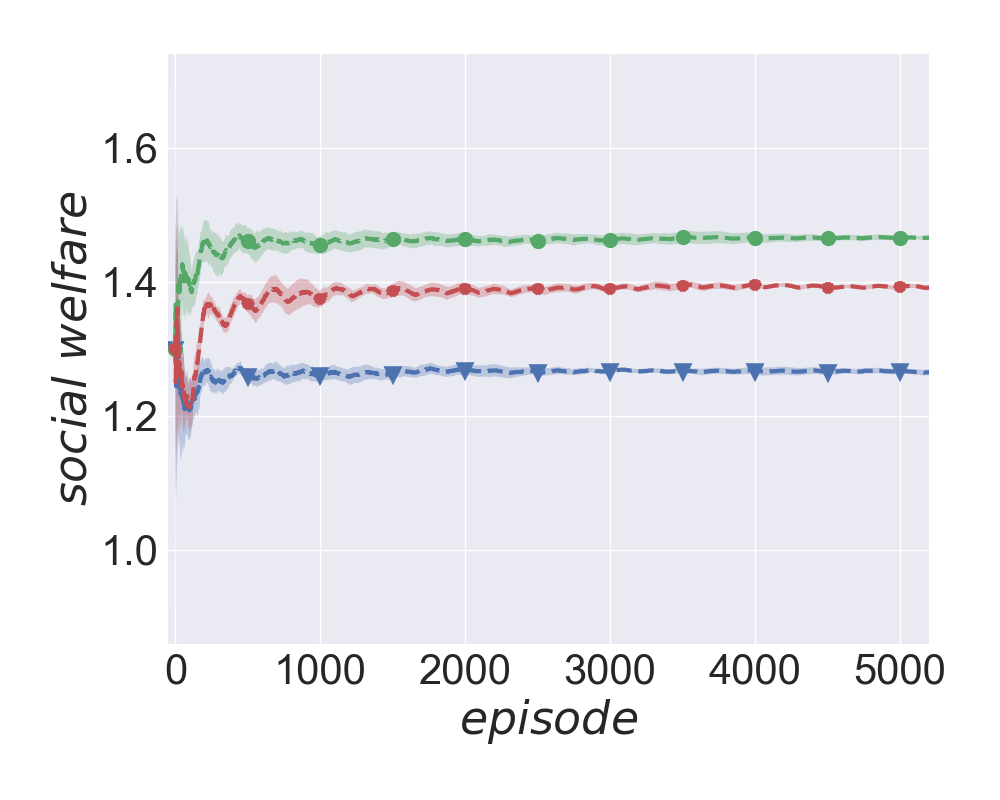}
}
\caption{Apple-Pear game: $agent_2$'s policy varies between $\pi^c$ and $\pi^d$ every 110 episodes.
The average rewards of the $agent_1$ are higher when using our approach than using $\pi^c$,
which means our approach can avoid being exploited by defective opponents.
The social welfare is higher than using $\pi^d$,
indicating that our approach can seek for cooperation against cooperative ones.}
\label{achange}
\end{figure*}

\begin{figure*}[htbp]
\centering
\subfigure{\includegraphics[height=1.225in,width=1.75in,angle=0]{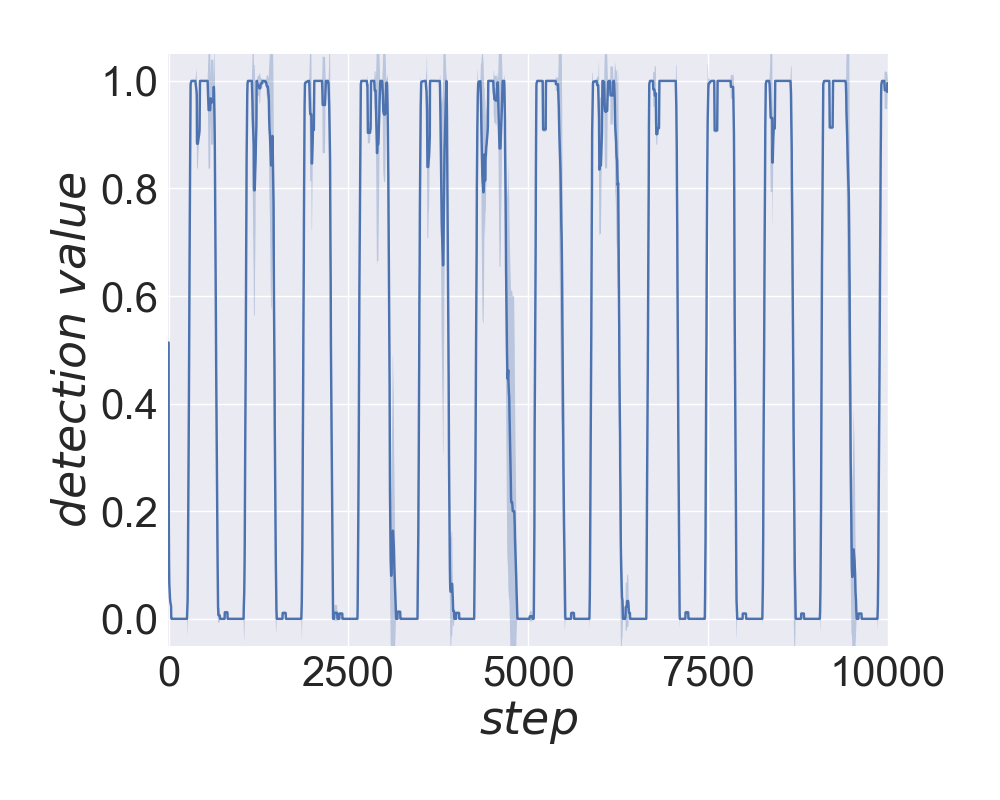}
\includegraphics[height=1.225in,width=1.75in,angle=0]{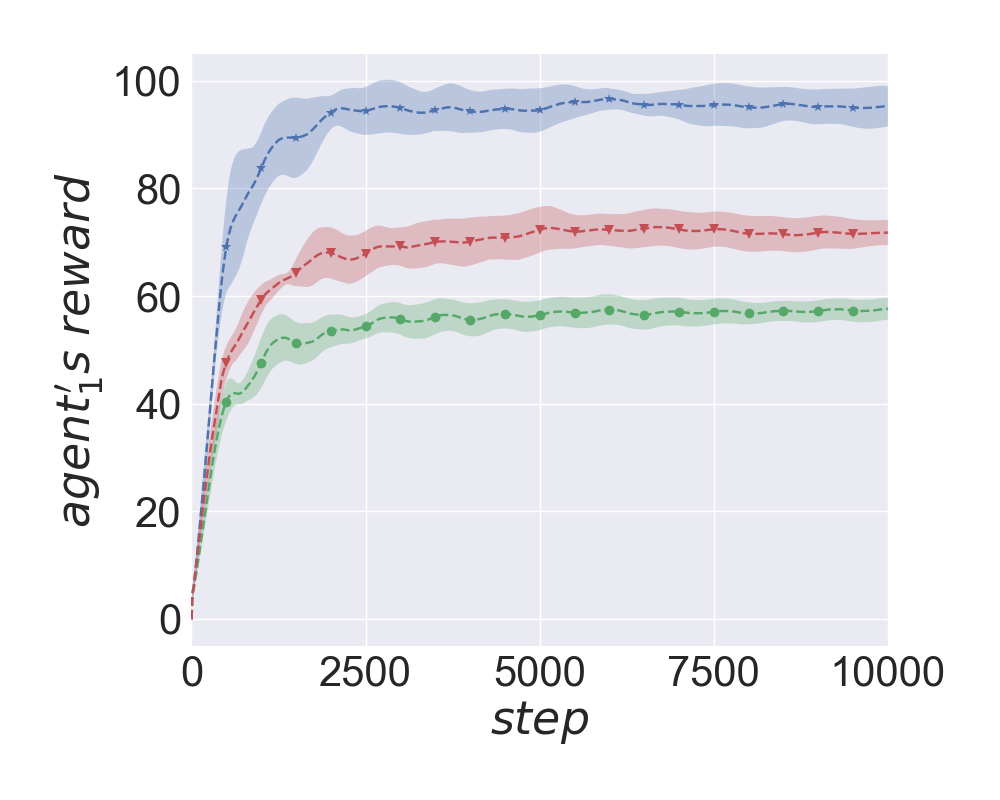}
\includegraphics[height=1.225in,width=1.75in,angle=0]{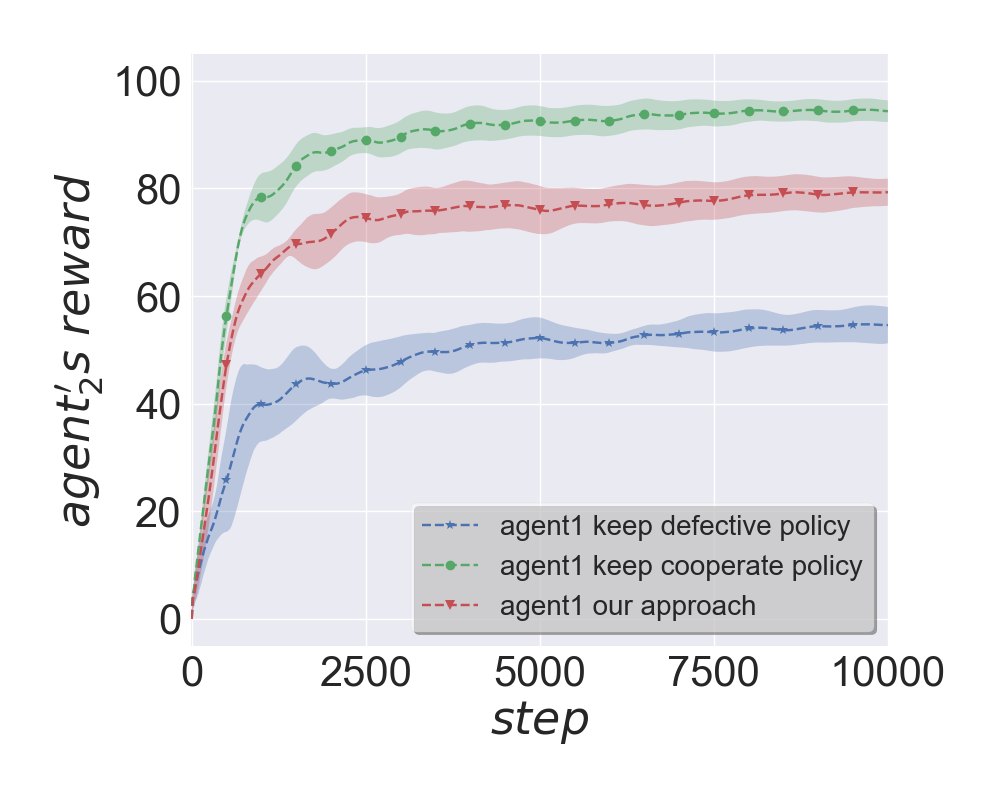}
\includegraphics[height=1.225in,width=1.75in,angle=0]{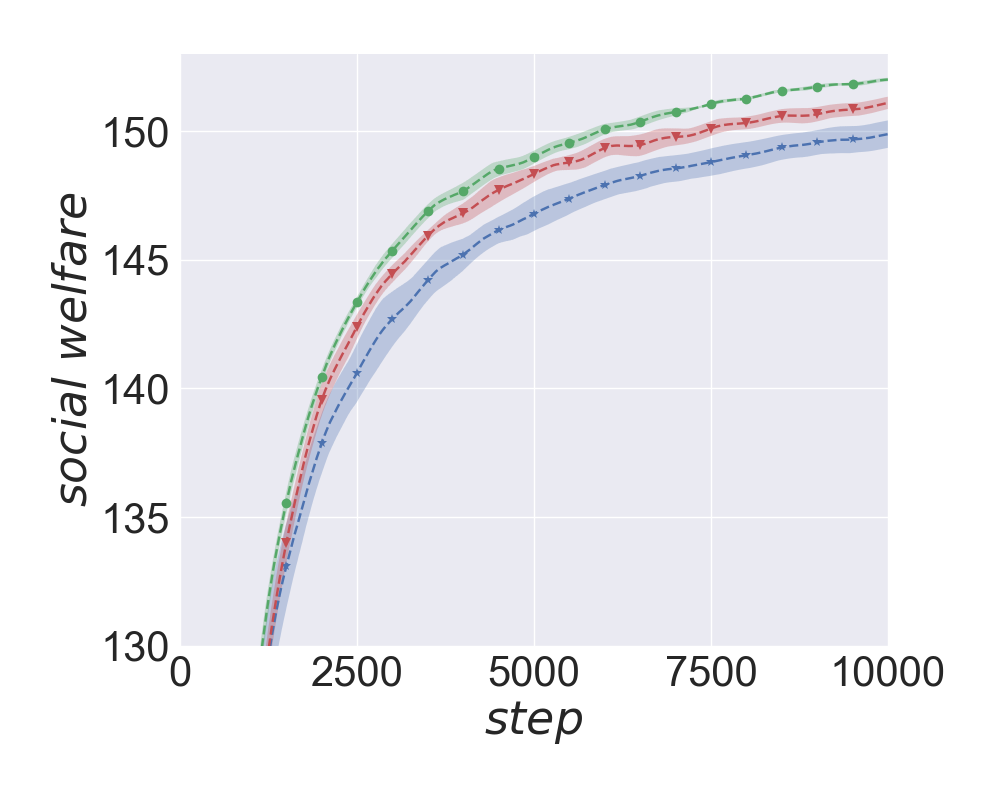}
}
\caption{Fruit Gathering game: $agent_2$'s policy varies between $\pi^c$ and $\pi^d$ every 300 steps.
Similar phenomenon can be observed as in Figure \ref{achange}.}
\label{gchange}
\end{figure*}

\subsection{Performance under Self-Play}
Next, we evaluate the learning performance of our approach under self-play. Since the initialized policies of agents can affect their behaviors in the game, we evaluate under all different initial conditions: a) $agent_1$ starts with cooperation policy and $agent_2$ starts with defecting policy; b) $agent_1$ starts with defecting policy and $agent_2$ starts with cooperation policy; c) both agents start with cooperation policies; d) both agents start with defecting policies. Agents converge to full cooperation for all four cases. We present the results for the last case, which is the most challenging one (see Figure~\ref{aselfplay} and Figure~\ref{gselfplay}).
When both agents start with a defection policy, it is more likely for them to model each other as defective and thus both play defect thereafter. Our approach enables agents to successfully detect the cooperation tendency of their opponents from sequential actions and converge to cooperation at last. In the Apple-Pear game, agents converge efficiently within few episodes. The reason is that when agents are close to fruits they prefer, they will collect fruits no matter whether their policies are cooperative or not. Thus, an agent is more likely to be detected as being cooperative, which induces its opponent to change policies towards cooperation. In contrast, for the Fruit Gathering game, agents need a relatively longer time before converging to full cooperation. This is because in the Fruit Gathering game, the main feature of detecting the cooperation degree is beam emitting frequency. When one agent emits a beam continuously, they change to defect. Only when both agents collect fruits without emitting beam would lead to mutual cooperation.

\subsection{Playing with Opponents with Changing Strategies}
Now, we evaluate the performance against switching opponents, during a repeated interaction of $T$ episodes, the opponent changes its policy after a certain number of episodes and the learning agent does not know when the switches happen. Through this, we can evaluate the cooperation degree detection performance  of our network, and verify whether our strategy can perform better than the fully cooperation or defection strategy from two aspects: 1) our approach seeks for mutual cooperation whenever
possible; 2) our approach is robust against selfish exploitation.
In the Apple-Pear game, we vary the value of $N_{change}$ in the range of $[50, 200]$ at the interval of 30,
and similarly in the Fruit Gathering game we vary it in the range of $[100,500]$ at the interval of 100. Only one set of results for each game are provided in Figure 9 and 10. The results of other values of $N_{change}$ are in the Appendix. Similar phenomenon can be observed for other values of $N_{change}$. From Figure 9 and 10, we observe that for both games, the average rewards of the learning agent ($agent_1$) are higher than its rewards using cooperation strategy $\pi^c$.
And the social welfare (the sum of both agents' rewards)
is higher than that using defection strategy $\pi^d$. This indicates that our approach can prevent agents from being exploited by defecting opponents and seek for cooperation against cooperative ones.  By comparing the results for different values of $N_{change}$, we find that the detection accuracy decreases when the opponent changes its policy quickly. Since the agent requires observing several episodes to detect the cooperation degree of the opponent, when the agent realizes its opponent changes the policy and adjusts its policy, the opponent may change the policy again. This problem becomes less severe when $T$ is comparatively large.

\section{Conclusions}

In this paper, we make the first step in investigating multiagent learning problem in large-scale PD games by leveraging the recent advance of deep reinforcement learning.
A deep multiagent RL approach is proposed towards mutual cooperation in SPD games to support adaptive end-to-end learning.
Empirical simulation shows that our agent can efficiently achieve mutual cooperation under self-play and also perform well against opponents with changing strategies.
As the first step towards solving multiagent learning problem in large-scale environments, we believe there are many interesting questions remaining for future work. One worthwhile direction is how to generalize our approach to other classes of large-scale multiagent games. Generalized policy detection and reuse techniques should be proposed, e.g., by extending existing approaches in traditional reinforcement learning contexts \cite{Hernandez2016Identifying,hernandez2016bayesian,Hernandez2017Towards}.

\bibliographystyle{named}
\bibliography{ijcai18}

\begin{thebibliography}{}

\bibitem[\protect\citeauthoryear{Axelrod}{1984}]{axelrod1984evolution}
Robert Axelrod.
\newblock The evolution of cooperation, 1984.

\bibitem[\protect\citeauthoryear{Banerjee and Sen}{2007}]{banerjee2007reaching}
Dipyaman Banerjee and Sandip Sen.
\newblock Reaching pareto-optimality in prisoner's dilemma using conditional
  joint action learning.
\newblock {\em Autonomous Agents and Multi-Agent Systems}, 15(1):91--108, 2007.

\bibitem[\protect\citeauthoryear{Busoniu \bgroup \em et al.\egroup
  }{2008}]{busoniu2008comprehensive}
Lucian Busoniu, Robert Babuska, and Bart De~Schutter.
\newblock A comprehensive survey of multiagent reinforcement learning.
\newblock {\em IEEE Transactions on Systems, Man, And Cybernetics-Part C:
  Applications and Reviews, 38 (2), 2008}, 2008.

\bibitem[\protect\citeauthoryear{Crandall and
  Goodrich}{2005}]{crandall2005learning}
Jacob~W Crandall and Michael~A Goodrich.
\newblock Learning to teach and follow in repeated games.
\newblock In {\em AAAI workshop on Multiagent Learning}, 2005.

\bibitem[\protect\citeauthoryear{Crandall}{2012}]{crandall2012just}
Jacob~W Crandall.
\newblock Just add pepper: extending learning algorithms for repeated matrix
  games to repeated markov games.
\newblock In {\em Proceedings of the 11th International Conference on
  Autonomous Agents and Multiagent Systems-Volume 1}, pages 399--406.
  International Foundation for Autonomous Agents and Multiagent Systems, 2012.

\bibitem[\protect\citeauthoryear{Damer and Gini}{2008}]{damer2008achieving}
Steven Damer and Maria~L Gini.
\newblock Achieving cooperation in a minimally constrained environment.
\newblock In {\em AAAI}, pages 57--62, 2008.

\bibitem[\protect\citeauthoryear{Elidrisi \bgroup \em et al.\egroup
  }{2014}]{elidrisi2014fast}
Mohamed Elidrisi, Nicholas Johnson, Maria Gini, and Jacob Crandall.
\newblock Fast adaptive learning in repeated stochastic games by game
  abstraction.
\newblock In {\em Proceedings of the 2014 international conference on
  Autonomous agents and multi-agent systems}, pages 1141--1148. International
  Foundation for Autonomous Agents and Multiagent Systems, 2014.

\bibitem[\protect\citeauthoryear{Foerster \bgroup \em et al.\egroup
  }{2017a}]{foerster2017counterfactual}
Jakob Foerster, Gregory Farquhar, Triantafyllos Afouras, Nantas Nardelli, and
  Shimon Whiteson.
\newblock Counterfactual multi-agent policy gradients.
\newblock {\em arXiv preprint arXiv:1705.08926}, 2017.

\bibitem[\protect\citeauthoryear{Foerster \bgroup \em et al.\egroup
  }{2017b}]{foerster2017stabilising}
Jakob Foerster, Nantas Nardelli, Gregory Farquhar, Philip Torr, Pushmeet Kohli,
  Shimon Whiteson, et~al.
\newblock Stabilising experience replay for deep multi-agent reinforcement
  learning.
\newblock {\em arXiv preprint arXiv:1702.08887}, 2017.

\bibitem[\protect\citeauthoryear{Hao and Leung}{2015}]{hao2015introducing}
Jianye Hao and Ho-fung Leung.
\newblock Introducing decision entrustment mechanism into repeated bilateral
  agent interactions to achieve social optimality.
\newblock {\em Autonomous Agents and Multi-Agent Systems}, 29(4):658--682,
  2015.

\bibitem[\protect\citeauthoryear{He \bgroup \em et al.\egroup
  }{2016}]{he2016opponent}
He~He, Jordan Boyd-Graber, Kevin Kwok, and Hal Daum{\'e}~III.
\newblock Opponent modeling in deep reinforcement learning.
\newblock In {\em International Conference on Machine Learning}, pages
  1804--1813, 2016.

\bibitem[\protect\citeauthoryear{Hernandez-Leal and
  Kaisers}{2017}]{Hernandez2017Towards}
Pablo Hernandez-Leal and Michael Kaisers.
\newblock Towards a fast detection of opponents in repeated stochastic games.
\newblock In {\em The Workshop on Transfer in Reinforcement Learning}, 2017.

\bibitem[\protect\citeauthoryear{Hernandez-Leal \bgroup \em et al.\egroup
  }{2016a}]{hernandez2016bayesian}
Pablo Hernandez-Leal, Benjamin Rosman, Matthew~E Taylor, L~Enrique Sucar, and
  Enrique Munoz~de Cote.
\newblock A bayesian approach for learning and tracking switching,
  non-stationary opponents.
\newblock In {\em Proceedings of the 2016 International Conference on
  Autonomous Agents \& Multiagent Systems}, pages 1315--1316, 2016.

\bibitem[\protect\citeauthoryear{Hernandez-Leal \bgroup \em et al.\egroup
  }{2016b}]{Hernandez2016Identifying}
Pablo Hernandez-Leal, Matthew~E Taylor, Benjamin Rosman, L~Enrique Sucar, and
  Enrique Munoz~De Cote.
\newblock Identifying and tracking switching, non-stationary opponents: a
  bayesian approach.
\newblock In {\em Multiagent Interaction without Prior Coordination Workshop at
  AAAI}, 2016.

\bibitem[\protect\citeauthoryear{Hu and Wellman}{2003}]{hu2003nash}
Junling Hu and Michael~P Wellman.
\newblock Nash q-learning for general-sum stochastic games.
\newblock {\em Journal of machine learning research}, 4(Nov):1039--1069, 2003.

\bibitem[\protect\citeauthoryear{Leibo \bgroup \em et al.\egroup
  }{2017}]{leibo2017multi}
Joel~Z Leibo, Vinicius Zambaldi, Marc Lanctot, Janusz Marecki, and Thore
  Graepel.
\newblock Multi-agent reinforcement learning in sequential social dilemmas.
\newblock In {\em Proceedings of the 16th Conference on Autonomous Agents and
  MultiAgent Systems}, pages 464--473. International Foundation for Autonomous
  Agents and Multiagent Systems, 2017.

\bibitem[\protect\citeauthoryear{Lillicrap \bgroup \em et al.\egroup
  }{2015}]{lillicrap2015continuous}
Timothy~P Lillicrap, Jonathan~J Hunt, Alexander Pritzel, Nicolas Heess, Tom
  Erez, Yuval Tassa, David Silver, and Daan Wierstra.
\newblock Continuous control with deep reinforcement learning.
\newblock {\em arXiv preprint arXiv:1509.02971}, 2015.

\bibitem[\protect\citeauthoryear{Littman}{1994}]{littman1994markov}
Michael~L Littman.
\newblock Markov games as a framework for multi-agent reinforcement learning.
\newblock In {\em Proceedings of the eleventh international conference on
  machine learning}, volume 157, pages 157--163, 1994.

\bibitem[\protect\citeauthoryear{Lowe \bgroup \em et al.\egroup
  }{2017}]{lowe2017multi}
Ryan Lowe, Yi~Wu, Aviv Tamar, Jean Harb, Pieter Abbeel, and Igor Mordatch.
\newblock Multi-agent actor-critic for mixed cooperative-competitive
  environments.
\newblock {\em arXiv preprint arXiv:1706.02275}, 2017.

\bibitem[\protect\citeauthoryear{Mathieu and Delahaye}{2015}]{mathieu2015new}
Philippe Mathieu and Jean-Paul Delahaye.
\newblock New winning strategies for the iterated prisoner's dilemma.
\newblock In {\em Proceedings of the 2015 International Conference on
  Autonomous Agents and Multiagent Systems}, pages 1665--1666. International
  Foundation for Autonomous Agents and Multiagent Systems, 2015.

\bibitem[\protect\citeauthoryear{Mnih \bgroup \em et al.\egroup
  }{2013}]{mnih2013playing}
Volodymyr Mnih, Koray Kavukcuoglu, David Silver, Alex Graves, Ioannis
  Antonoglou, Daan Wierstra, and Martin Riedmiller.
\newblock Playing atari with deep reinforcement learning.
\newblock {\em arXiv preprint arXiv:1312.5602}, 2013.

\bibitem[\protect\citeauthoryear{Mnih \bgroup \em et al.\egroup
  }{2015}]{mnih2015human}
Volodymyr Mnih, Koray Kavukcuoglu, David Silver, Andrei~A Rusu, Joel Veness,
  Marc~G Bellemare, Alex Graves, Martin Riedmiller, Andreas~K Fidjeland, Georg
  Ostrovski, et~al.
\newblock Human-level control through deep reinforcement learning.
\newblock {\em Nature}, 518(7540):529--533, 2015.

\bibitem[\protect\citeauthoryear{Nowak and Sigmund}{1993}]{nowak1993strategy}
Martin Nowak and Karl Sigmund.
\newblock A strategy of win-stay, lose-shift that outperforms tit-for-tat in
  the prisoner's dilemma game.
\newblock {\em Nature}, 364(6432):56--58, 1993.

\bibitem[\protect\citeauthoryear{Perolat \bgroup \em et al.\egroup
  }{2017}]{perolat2017multi}
Julien Perolat, Joel~Z Leibo, Vinicius Zambaldi, Charles Beattie, Karl Tuyls,
  and Thore Graepel.
\newblock A multi-agent reinforcement learning model of common-pool resource
  appropriation.
\newblock {\em arXiv preprint arXiv:1707.06600}, 2017.

\bibitem[\protect\citeauthoryear{Schulman \bgroup \em et al.\egroup
  }{2015}]{schulman2015high}
John Schulman, Philipp Moritz, Sergey Levine, Michael Jordan, and Pieter
  Abbeel.
\newblock High-dimensional continuous control using generalized advantage
  estimation.
\newblock {\em arXiv preprint arXiv:1506.02438}, 2015.

\bibitem[\protect\citeauthoryear{Stone and Veloso}{2000}]{stone2000multiagent}
Peter Stone and Manuela Veloso.
\newblock Multiagent systems: A survey from a machine learning perspective.
\newblock {\em Autonomous Robots}, 8(3):345--383, 2000.

\bibitem[\protect\citeauthoryear{Sunehag \bgroup \em et al.\egroup
  }{2017}]{sunehag2017value}
Peter Sunehag, Guy Lever, Audrunas Gruslys, Wojciech~Marian Czarnecki, Vinicius
  Zambaldi, Max Jaderberg, Marc Lanctot, Nicolas Sonnerat, Joel~Z Leibo, Karl
  Tuyls, et~al.
\newblock Value-decomposition networks for cooperative multi-agent learning.
\newblock {\em arXiv preprint arXiv:1706.05296}, 2017.

\bibitem[\protect\citeauthoryear{Sutton \bgroup \em et al.\egroup
  }{2000}]{sutton2000policy}
Richard~S Sutton, David~A McAllester, Satinder~P Singh, and Yishay Mansour.
\newblock Policy gradient methods for reinforcement learning with function
  approximation.
\newblock In {\em Advances in neural information processing systems}, pages
  1057--1063, 2000.

\bibitem[\protect\citeauthoryear{Wang \bgroup \em et al.\egroup
  }{2016}]{wang2016sample}
Ziyu Wang, Victor Bapst, Nicolas Heess, Volodymyr Mnih, Remi Munos, Koray
  Kavukcuoglu, and Nando de~Freitas.
\newblock Sample efficient actor-critic with experience replay.
\newblock {\em arXiv preprint arXiv:1611.01224}, 2016.

\bibitem[\protect\citeauthoryear{Williams}{1992}]{williams1992simple}
Ronald~J Williams.
\newblock Simple statistical gradient-following algorithms for connectionist
  reinforcement learning.
\newblock {\em Machine learning}, 8(3-4):229--256, 1992.

\end{thebibliography}

\onecolumn
\appendix
\section{Playing with Opponents with Changing Strategies}
\subsection{The Apple-Pear Game}

\begin{figure}[H]
\centering
\subfigure{\includegraphics[height=1.225in,width=1.75in,angle=0]{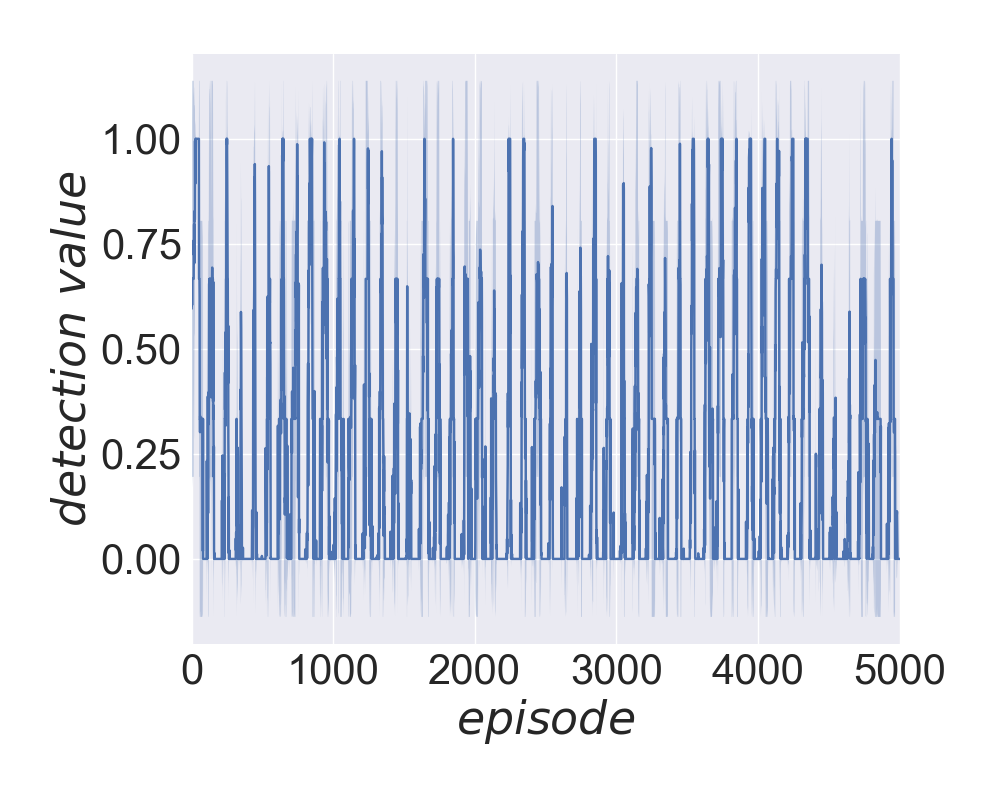}
\includegraphics[height=1.225in,width=1.75in,angle=0]{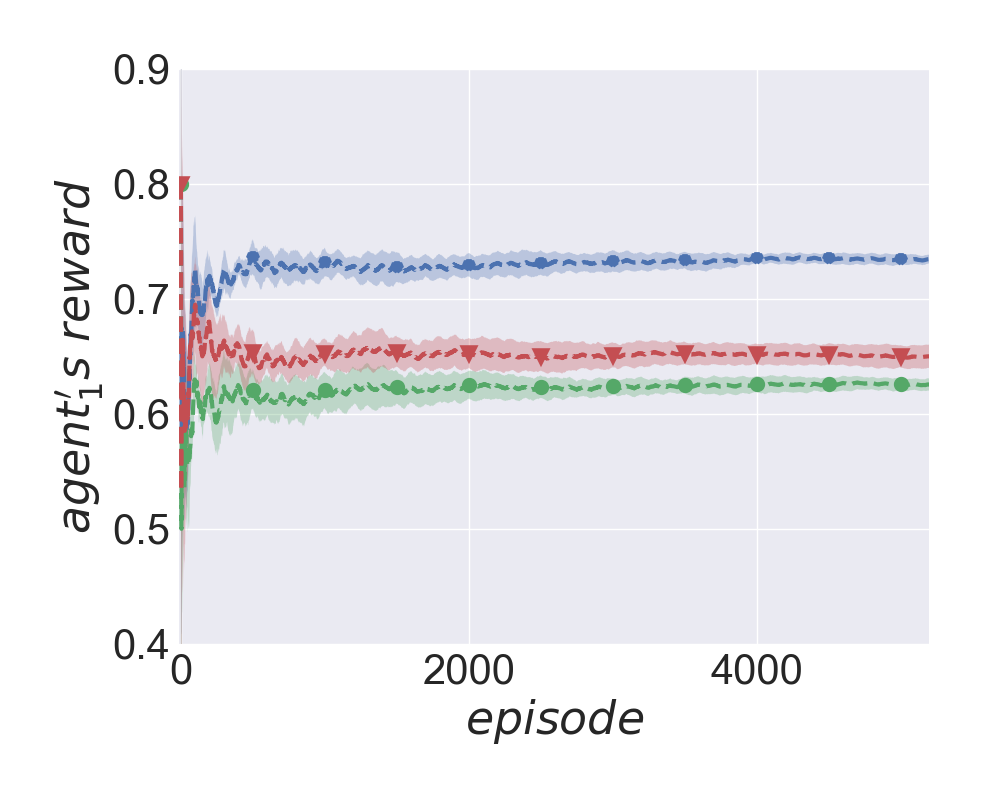}
\includegraphics[height=1.225in,width=1.75in,angle=0]{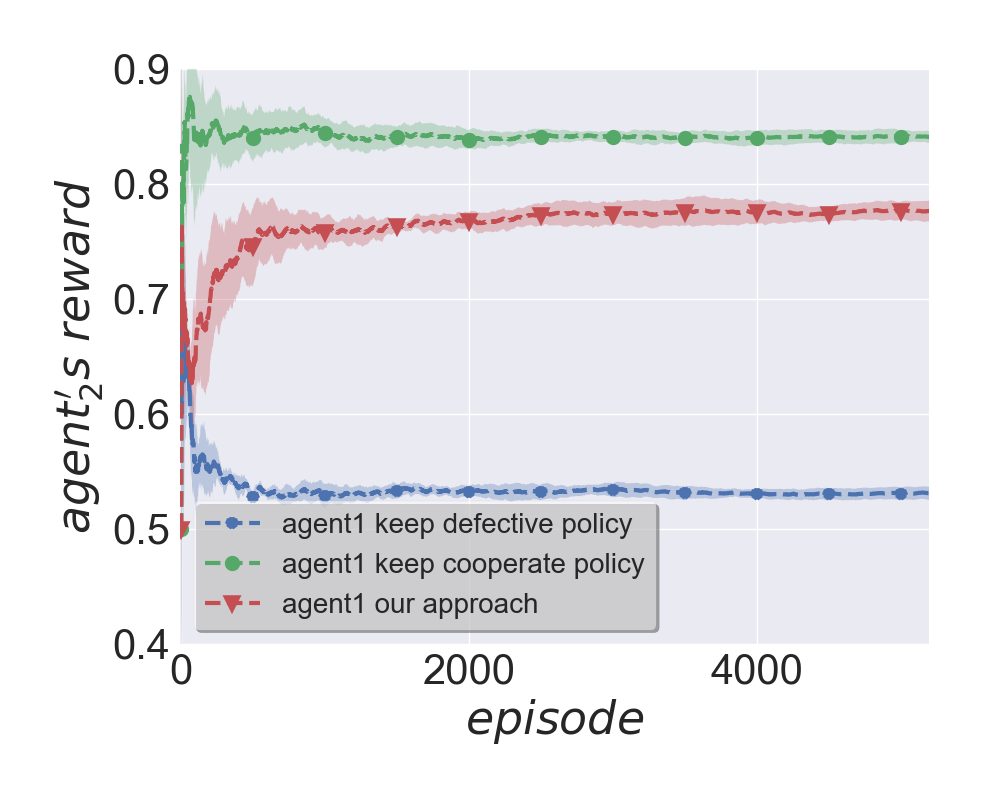}
\includegraphics[height=1.225in,width=1.75in,angle=0]{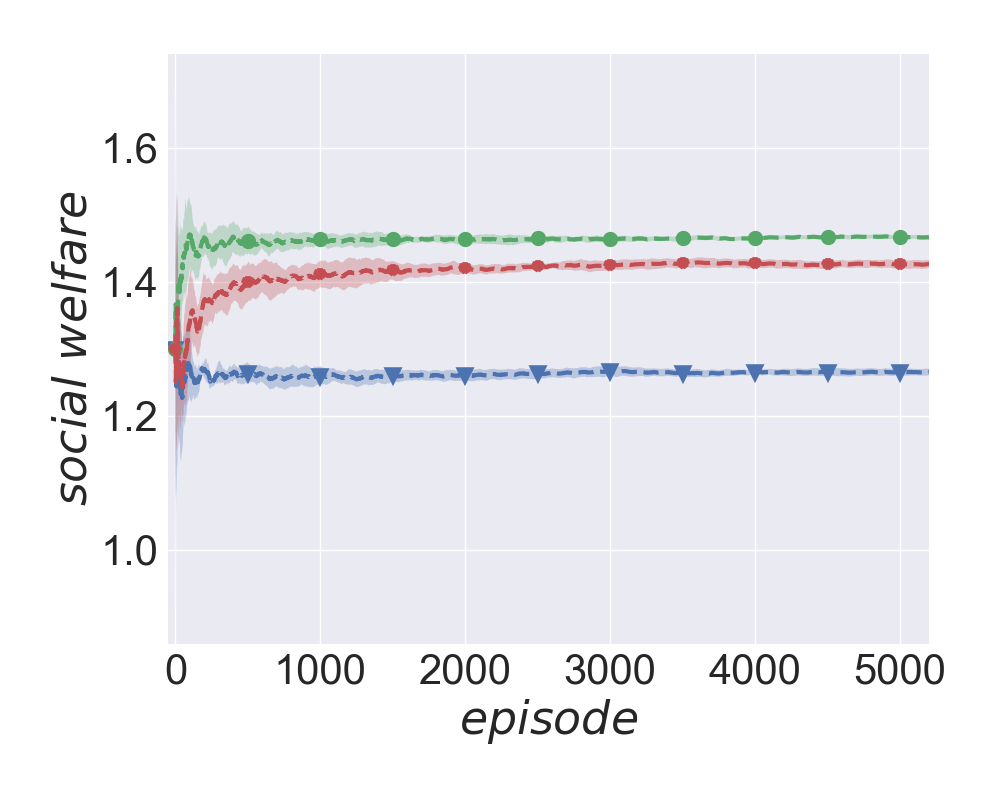}
}
\caption{Apple-Pear game: $agent_2$'s policy varies between $\pi^c$ and $\pi^d$ every 50 episodes.}
\label{achange}
\end{figure}

\begin{figure}[H]
\centering
\subfigure{\includegraphics[height=1.225in,width=1.75in,angle=0]{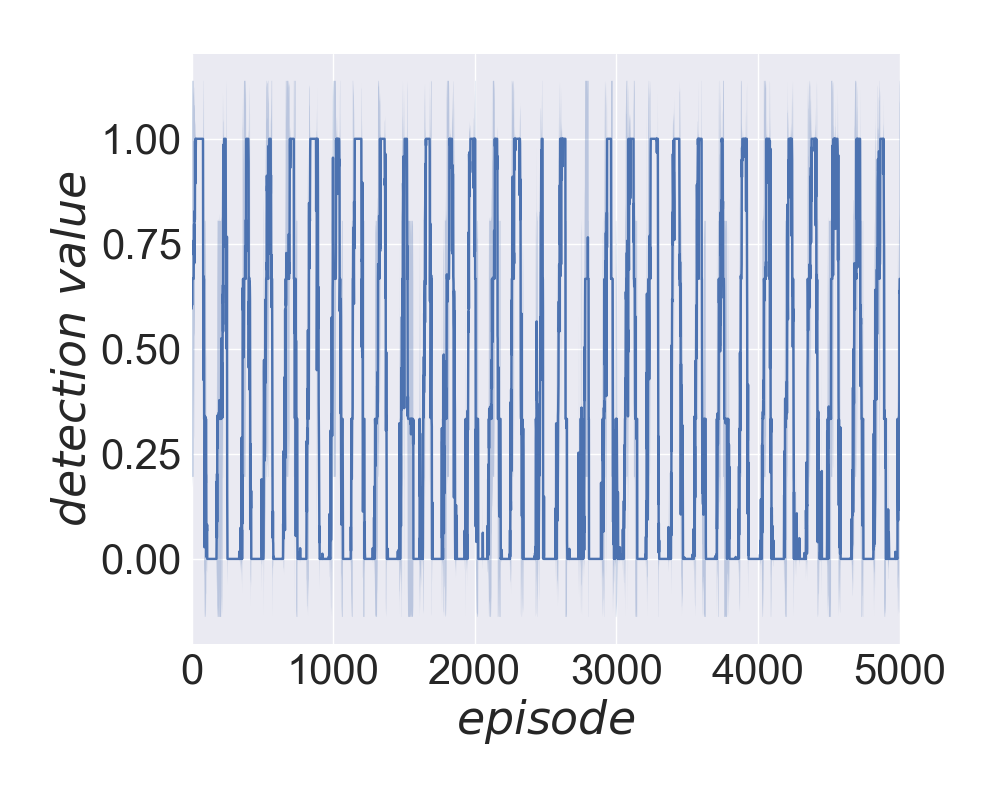}
\includegraphics[height=1.225in,width=1.75in,angle=0]{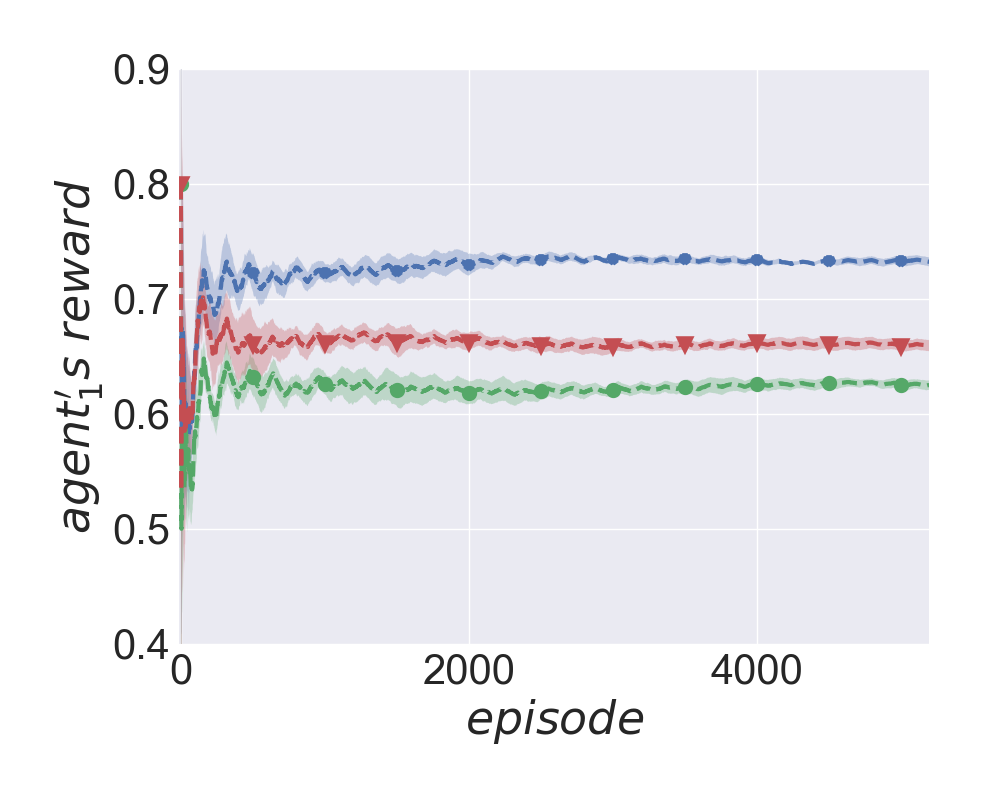}
\includegraphics[height=1.225in,width=1.75in,angle=0]{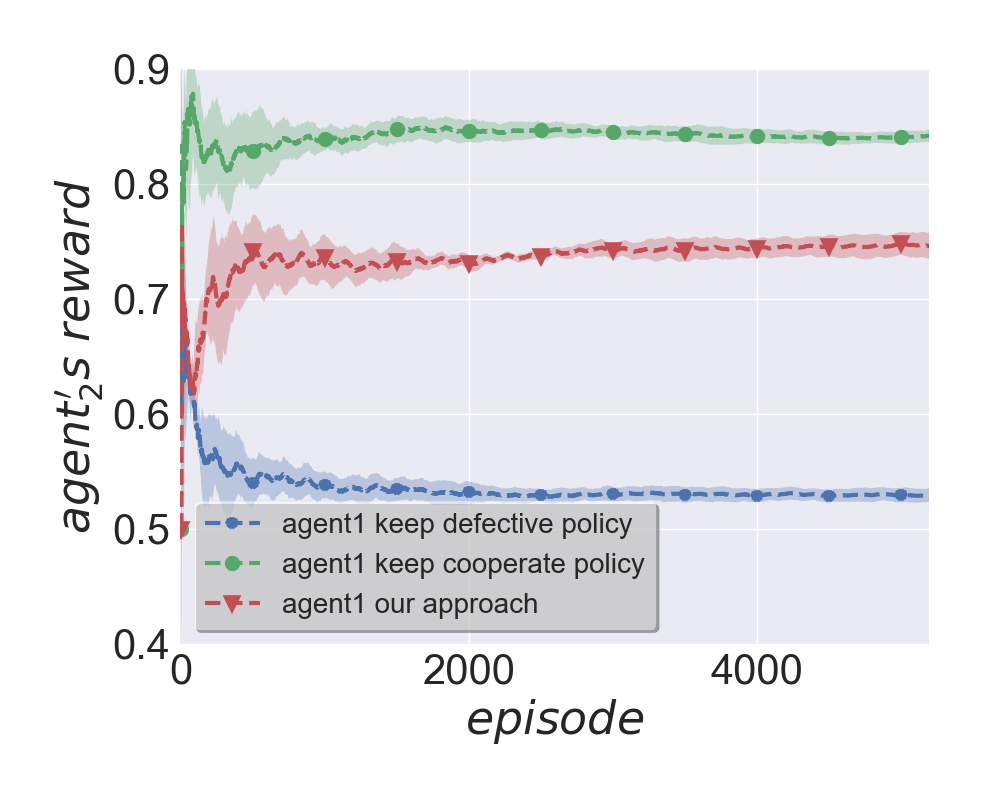}
\includegraphics[height=1.225in,width=1.75in,angle=0]{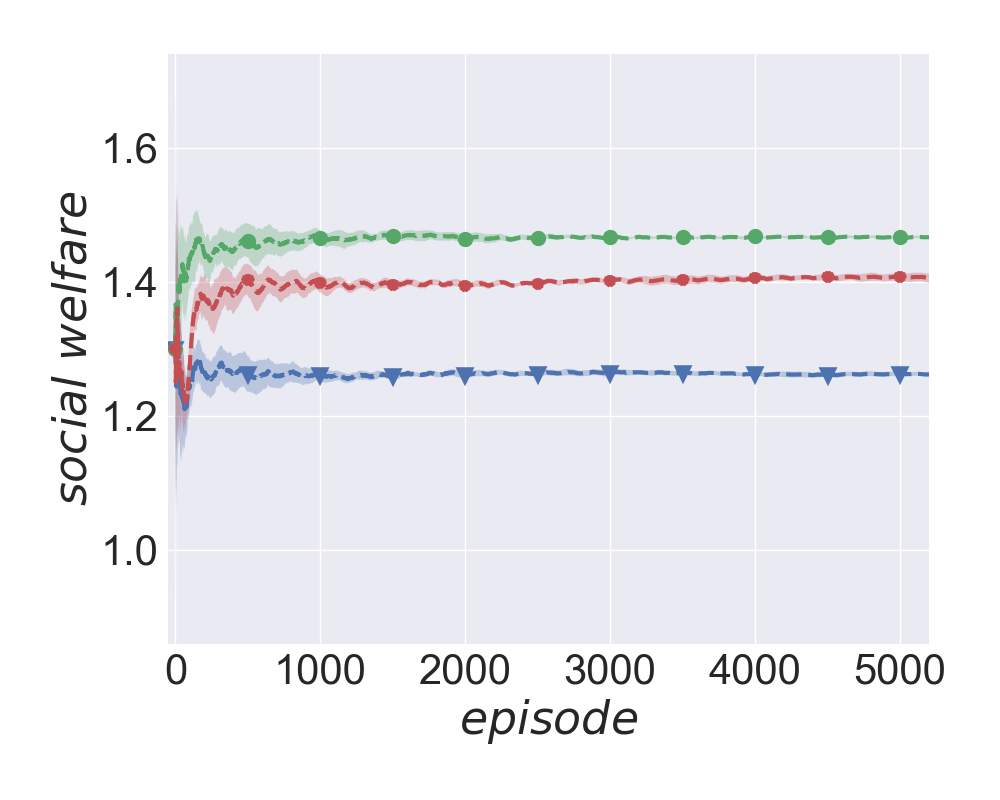}
}
\caption{Apple-Pear game: $agent_2$'s policy varies between $\pi^c$ and $\pi^d$ every 80 episodes.}
\label{achange}
\end{figure}

\begin{figure}[H]
\centering
\subfigure{\includegraphics[height=1.225in,width=1.75in,angle=0]{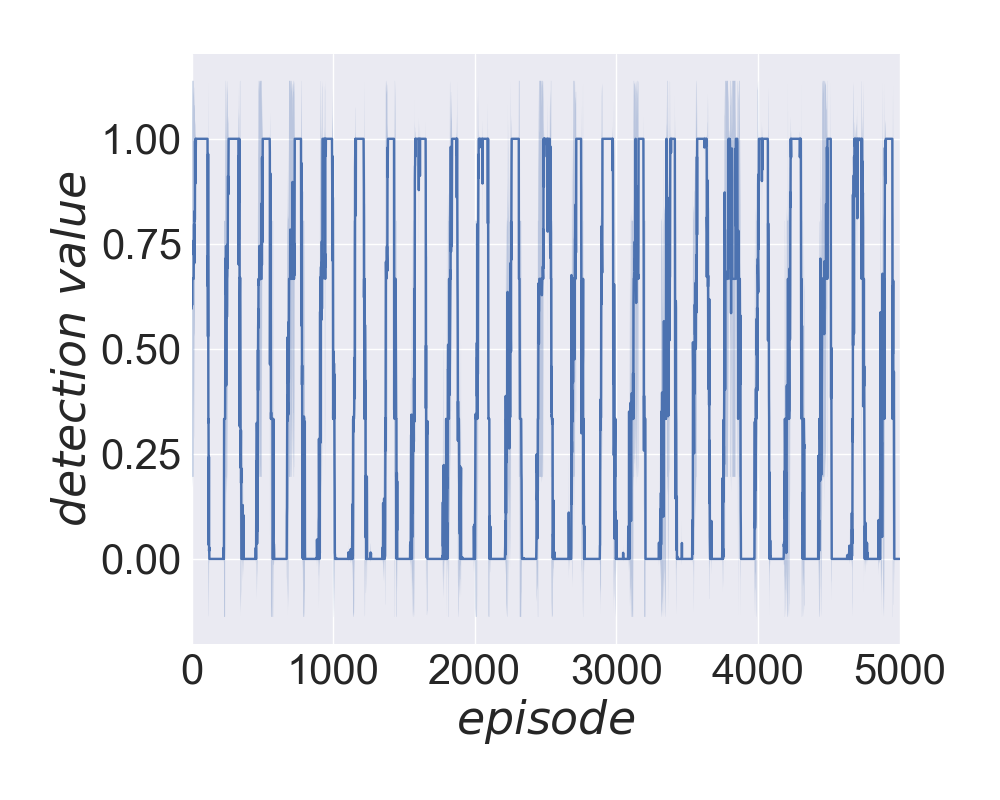}
\includegraphics[height=1.225in,width=1.75in,angle=0]{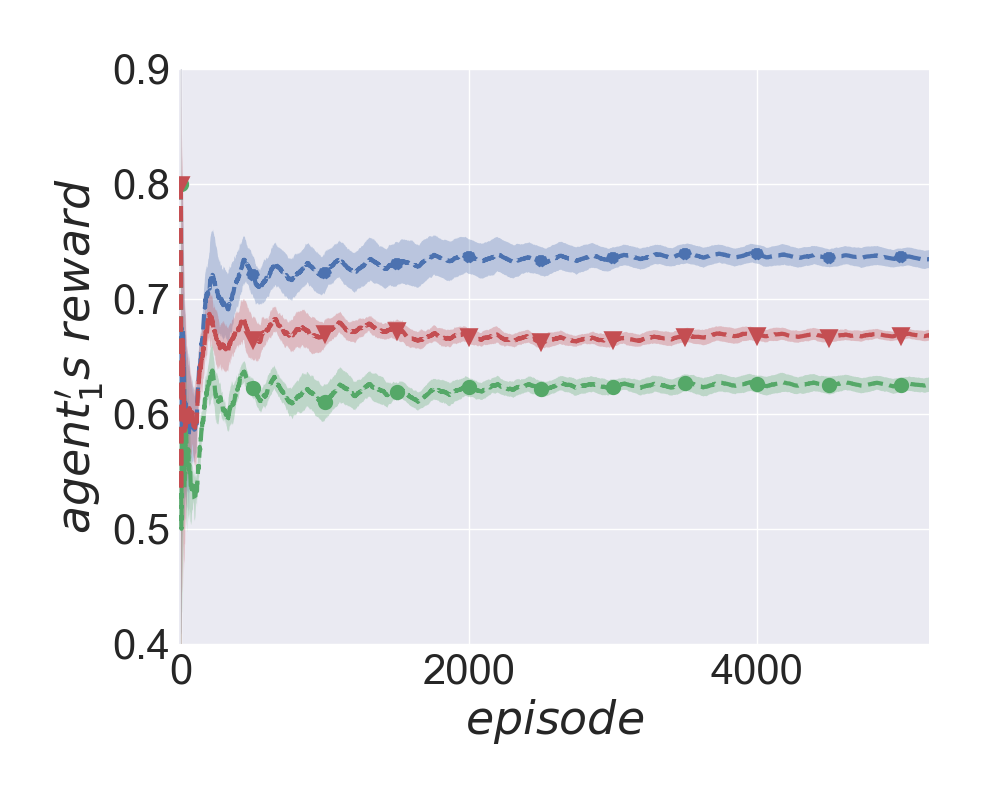}
\includegraphics[height=1.225in,width=1.75in,angle=0]{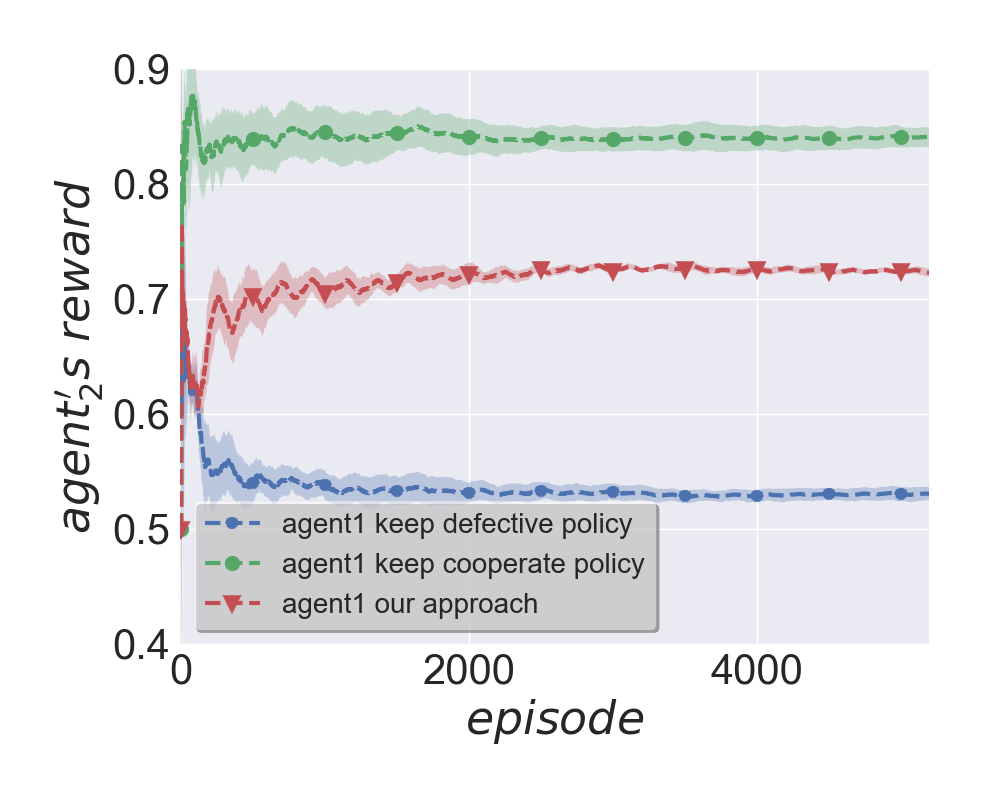}
\includegraphics[height=1.225in,width=1.75in,angle=0]{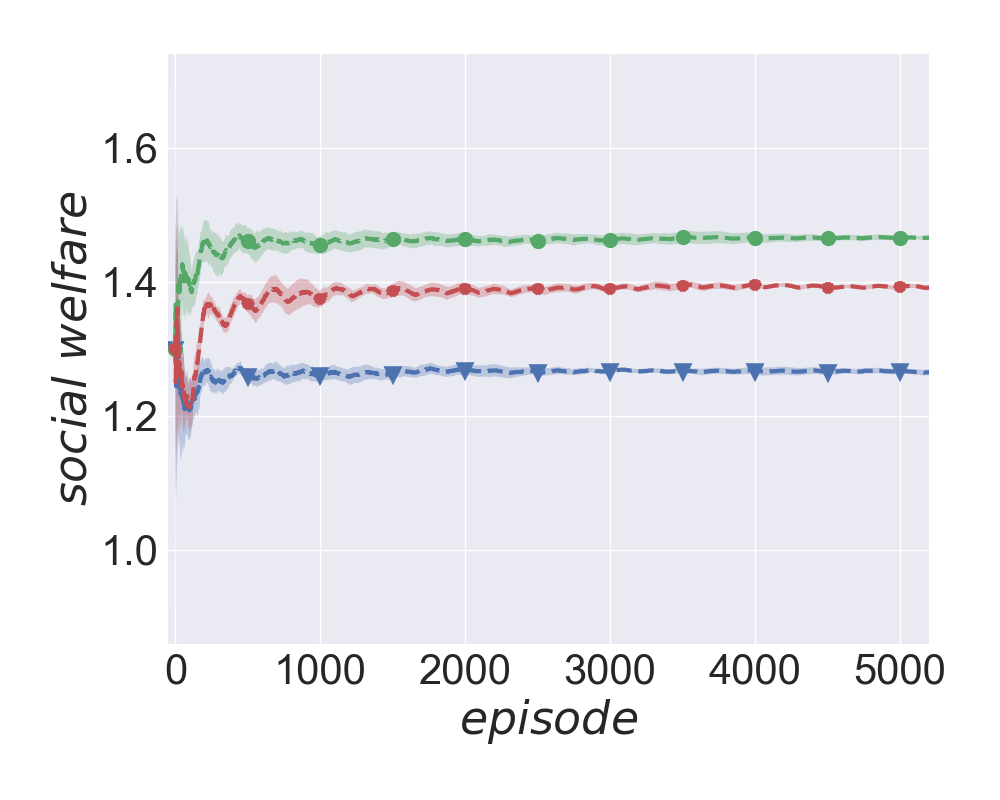}
}
\caption{Apple-Pear game: $agent_2$'s policy varies between $\pi^c$ and $\pi^d$ every 110 episodes.}
\label{achange}
\end{figure}

\begin{figure}[H]
\centering
\subfigure{\includegraphics[height=1.225in,width=1.75in,angle=0]{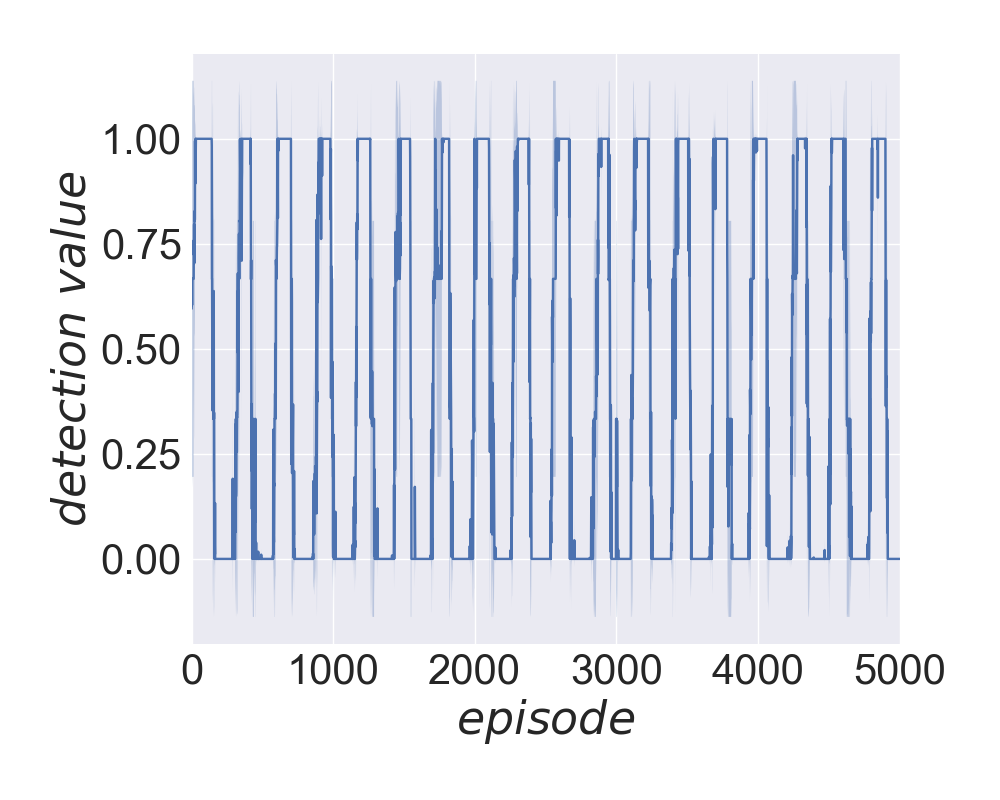}
\includegraphics[height=1.225in,width=1.75in,angle=0]{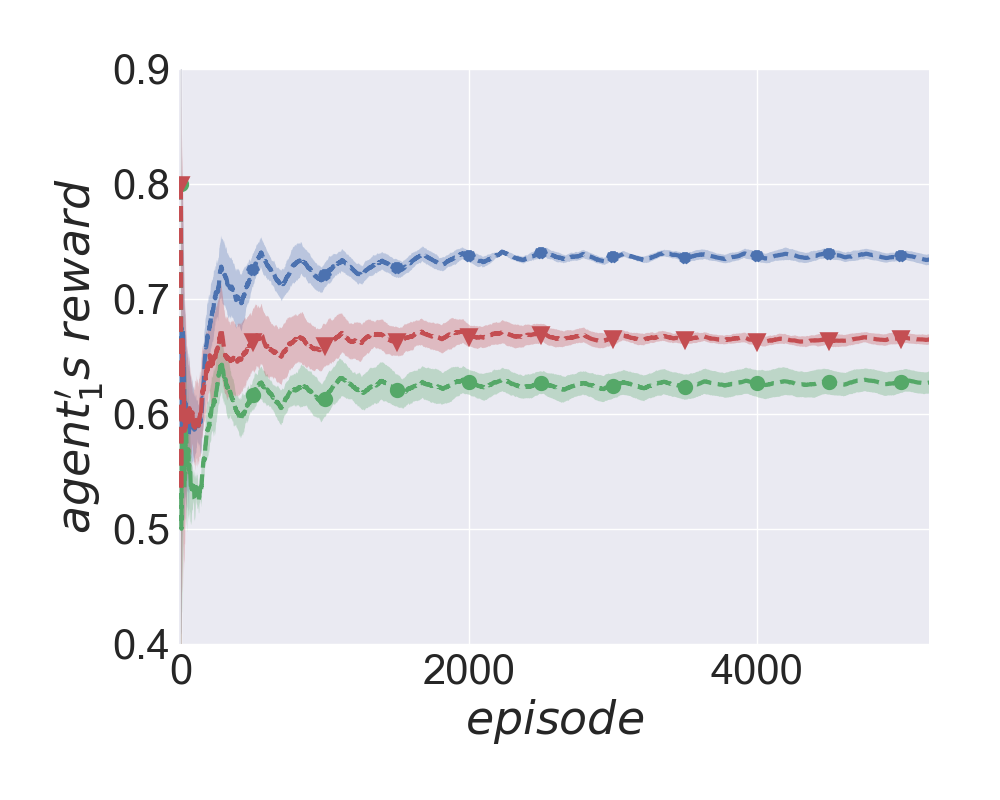}
\includegraphics[height=1.225in,width=1.75in,angle=0]{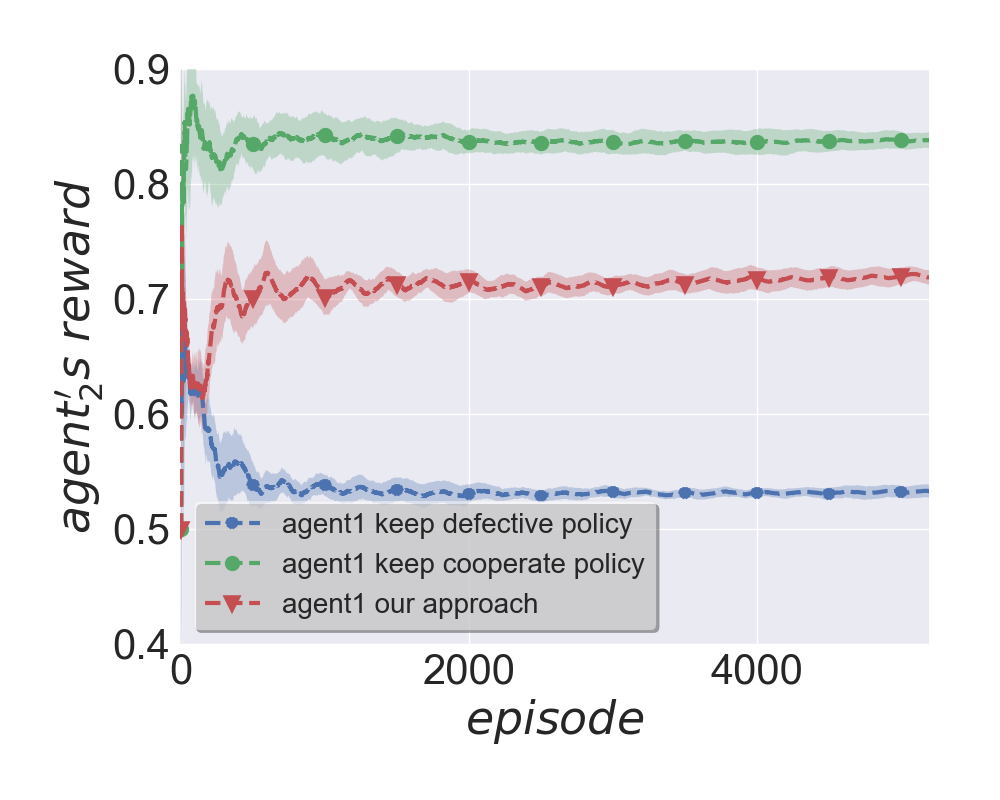}
\includegraphics[height=1.225in,width=1.75in,angle=0]{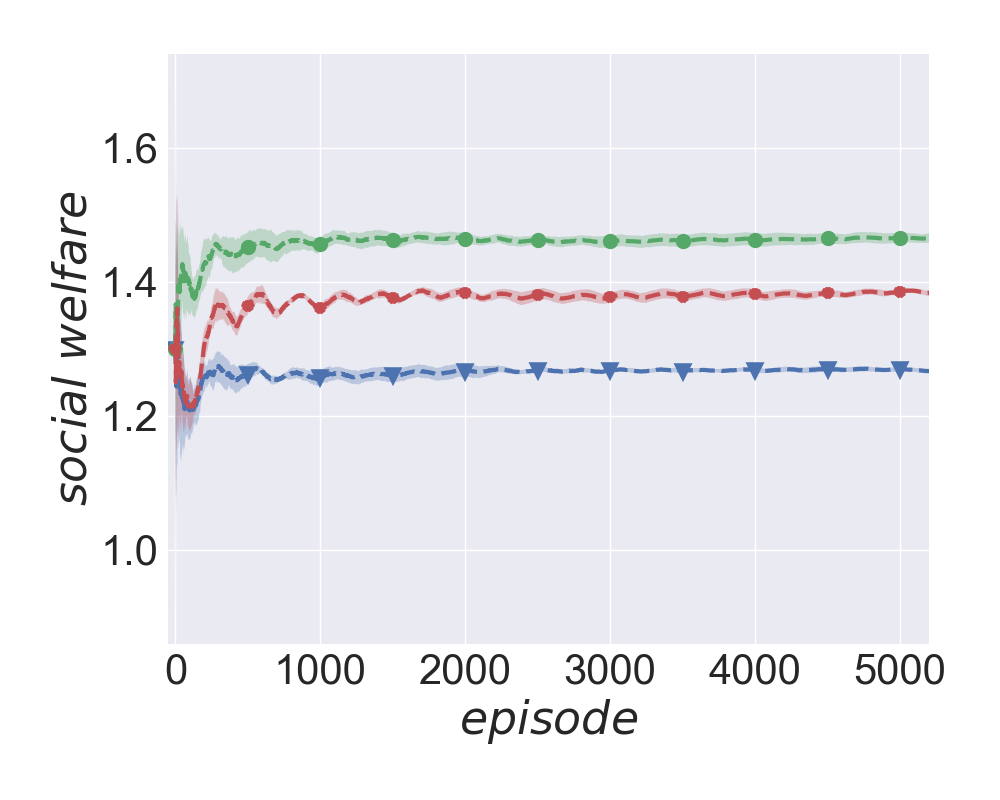}
}
\caption{Apple-Pear game: $agent_2$'s policy varies between $\pi^c$ and $\pi^d$ every 140 episodes.}
\label{achange}
\end{figure}

\begin{figure}[H]
\centering
\subfigure{\includegraphics[height=1.225in,width=1.75in,angle=0]{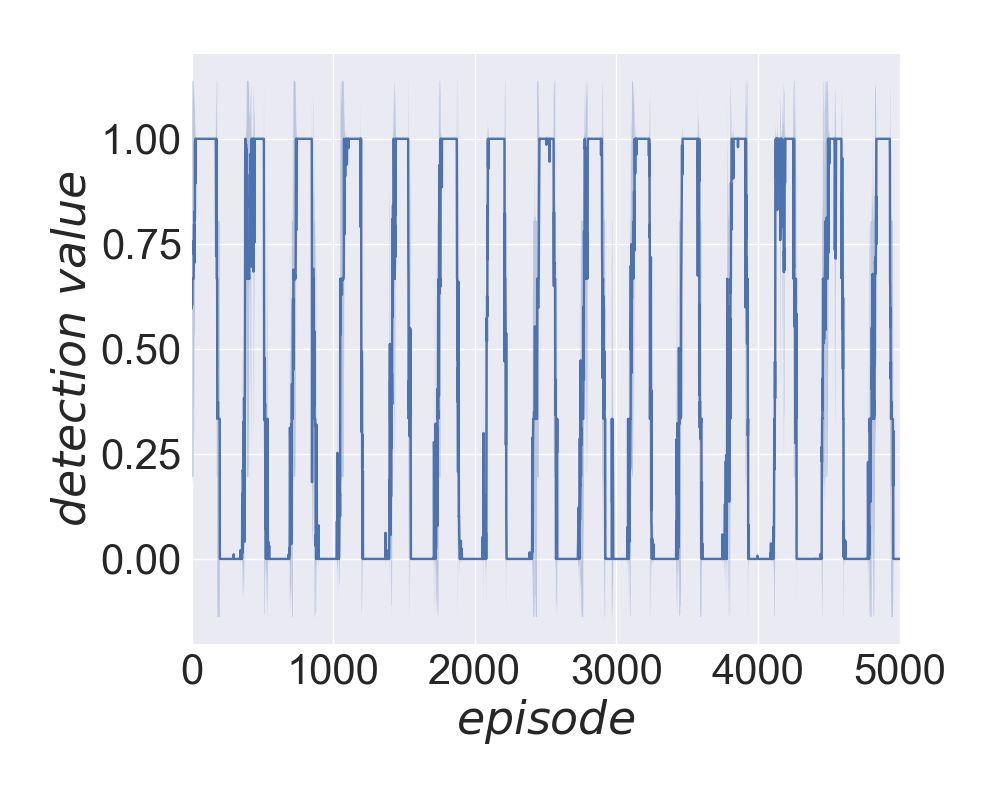}
\includegraphics[height=1.225in,width=1.75in,angle=0]{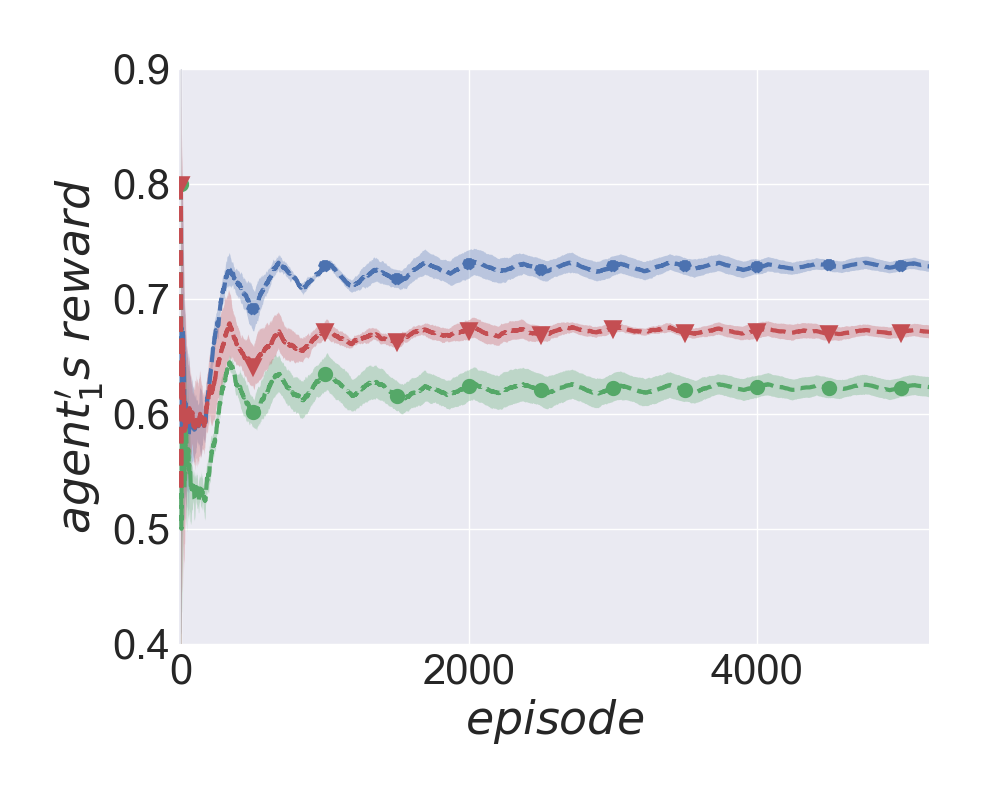}
\includegraphics[height=1.225in,width=1.75in,angle=0]{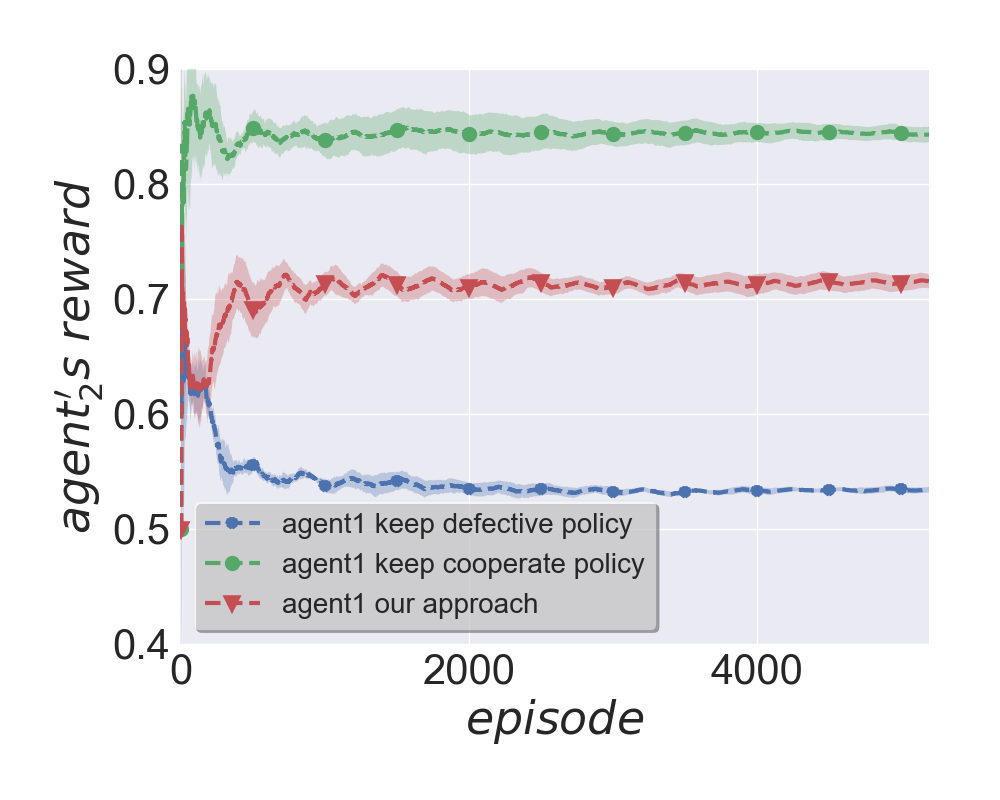}
\includegraphics[height=1.225in,width=1.75in,angle=0]{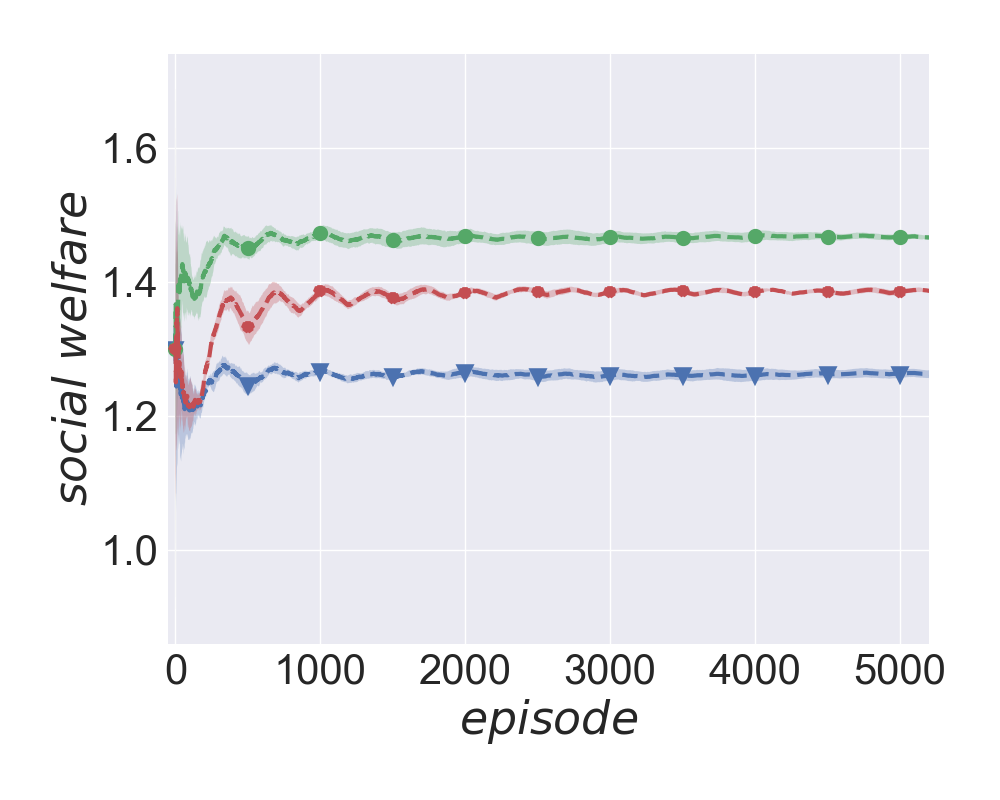}
}
\caption{Apple-Pear game: $agent_2$'s policy varies between $\pi^c$ and $\pi^d$ every 170 episodes.}
\label{achange}
\end{figure}

\begin{figure}[H]
\centering
\subfigure{\includegraphics[height=1.225in,width=1.75in,angle=0]{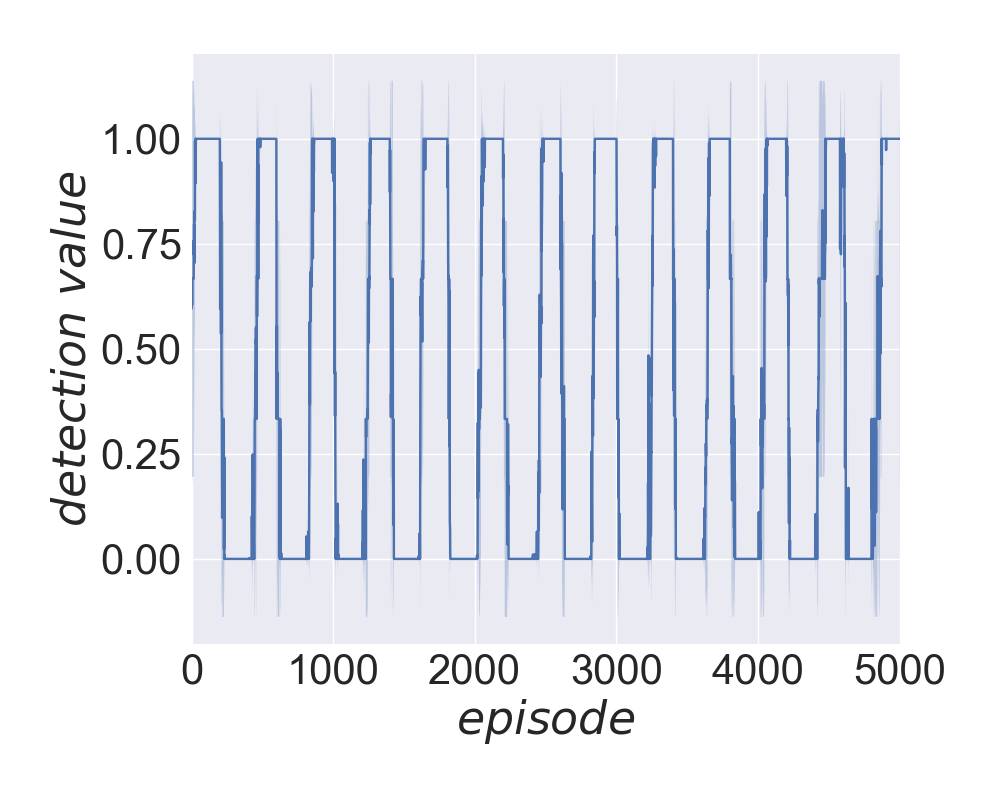}
\includegraphics[height=1.225in,width=1.75in,angle=0]{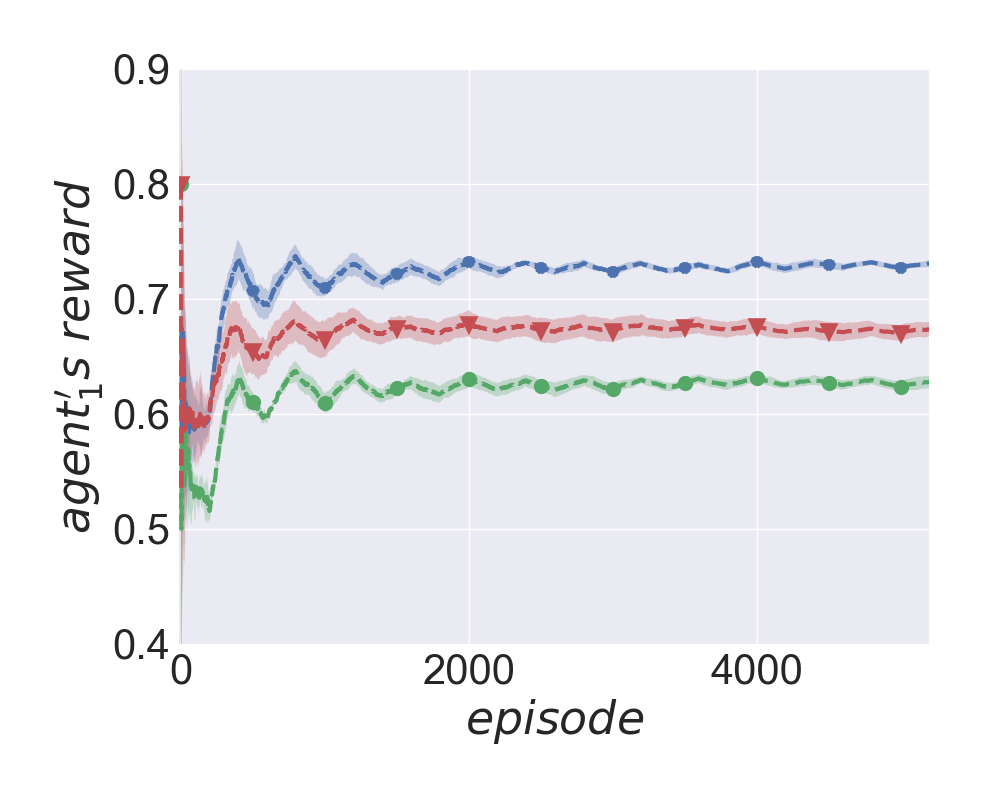}
\includegraphics[height=1.225in,width=1.75in,angle=0]{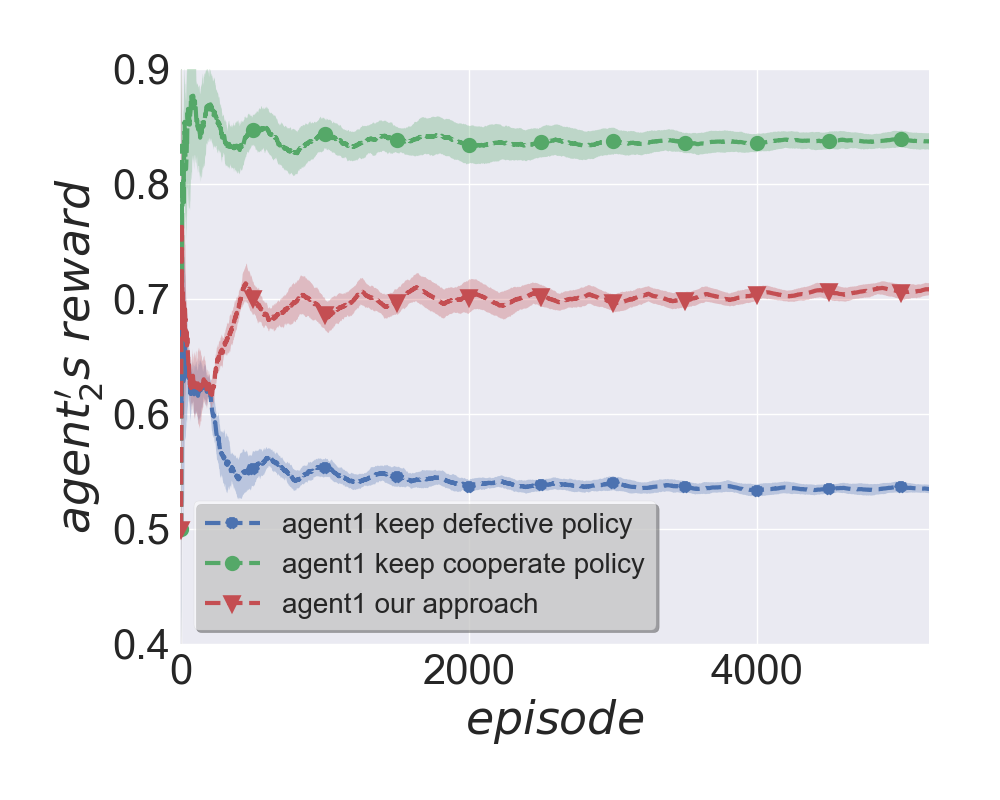}
\includegraphics[height=1.225in,width=1.75in,angle=0]{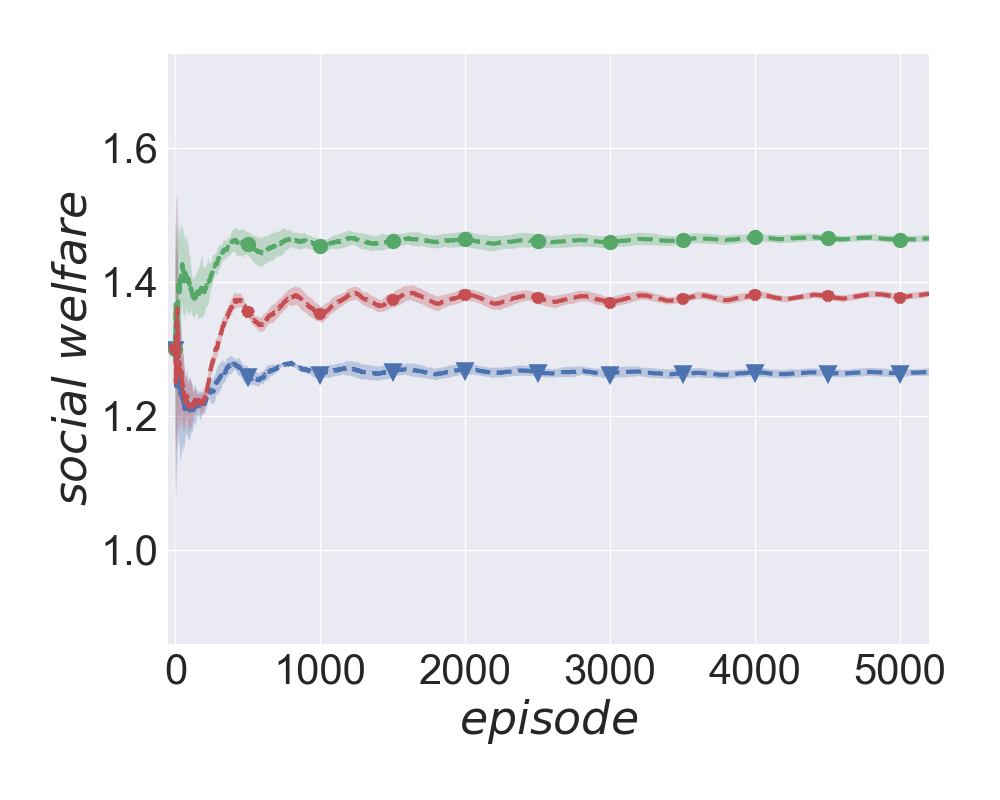}
}
\caption{Apple-Pear game: $agent_2$'s policy varies between $\pi^c$ and $\pi^d$ every 200 episodes.}
\label{achange}
\end{figure}

\subsection{The Gathering Game}

\begin{figure}[H]
\centering
\subfigure{\includegraphics[height=1.225in,width=1.75in,angle=0]{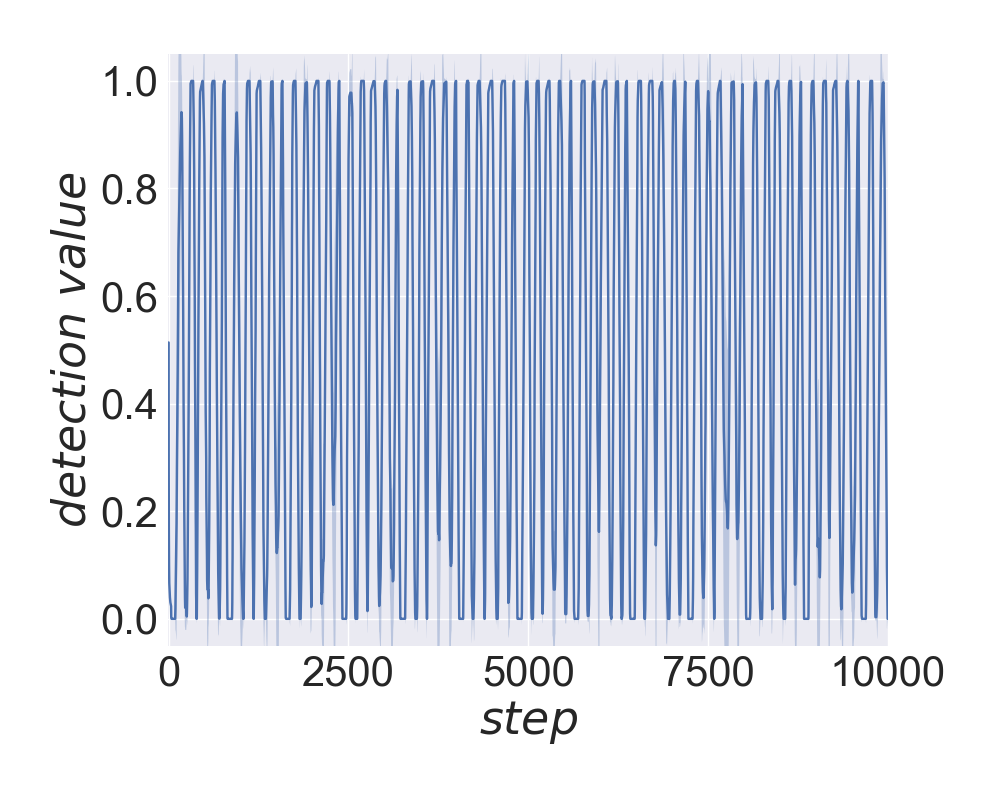}
\includegraphics[height=1.225in,width=1.75in,angle=0]{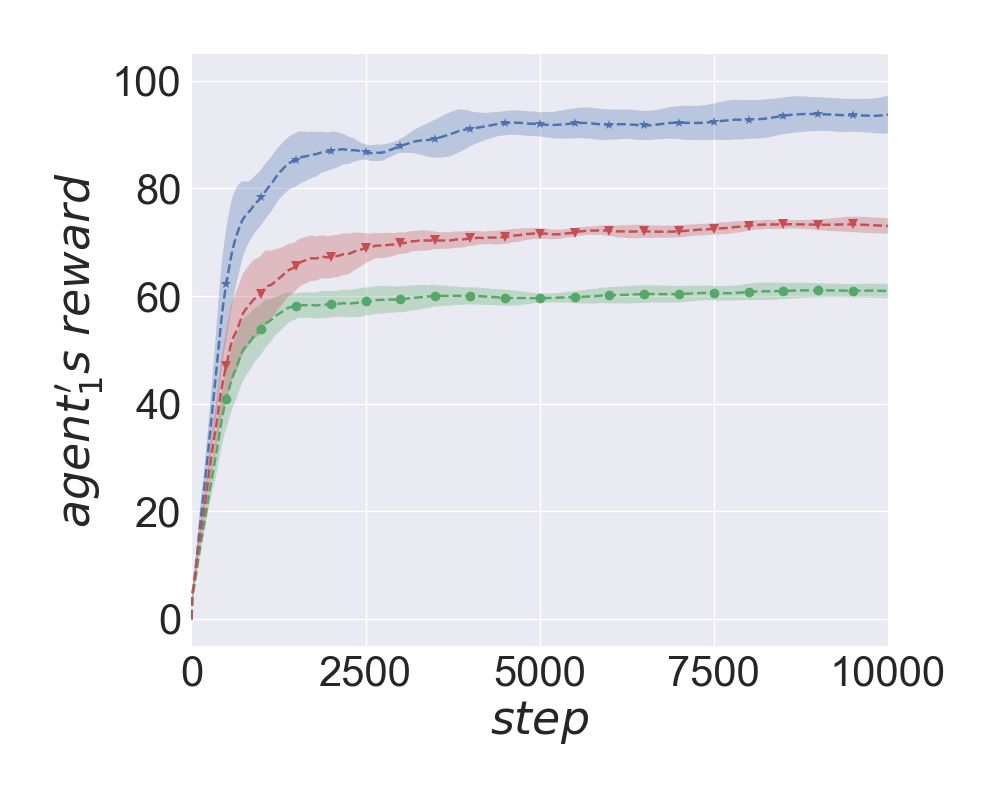}
\includegraphics[height=1.225in,width=1.75in,angle=0]{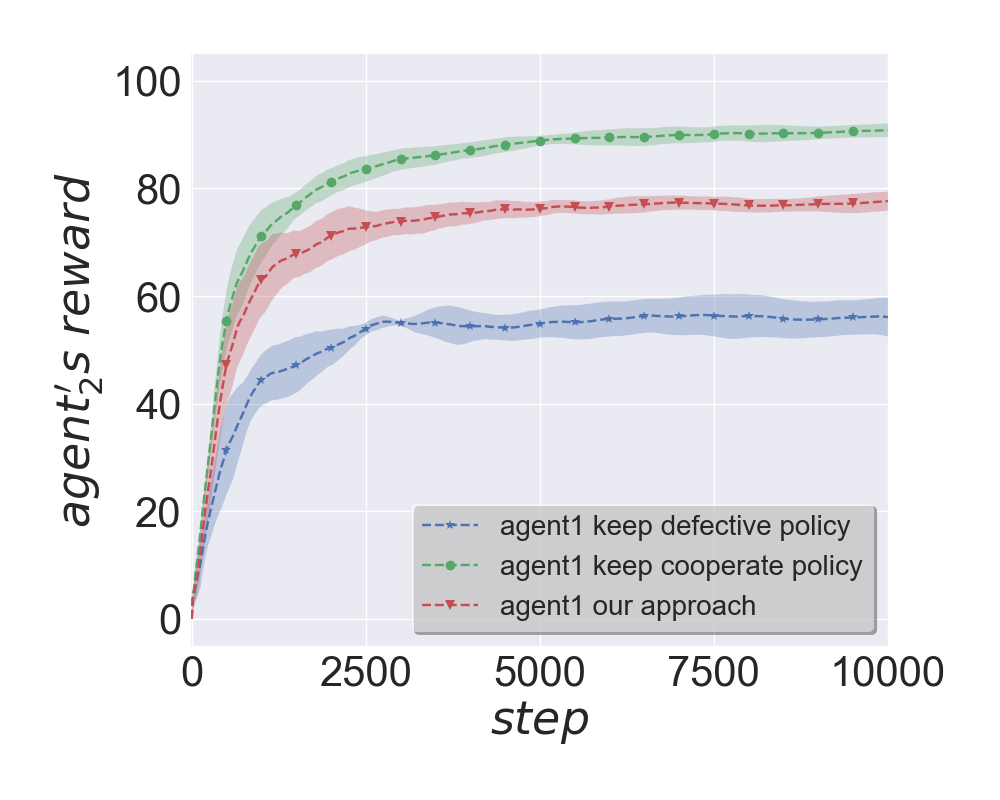}
\includegraphics[height=1.225in,width=1.75in,angle=0]{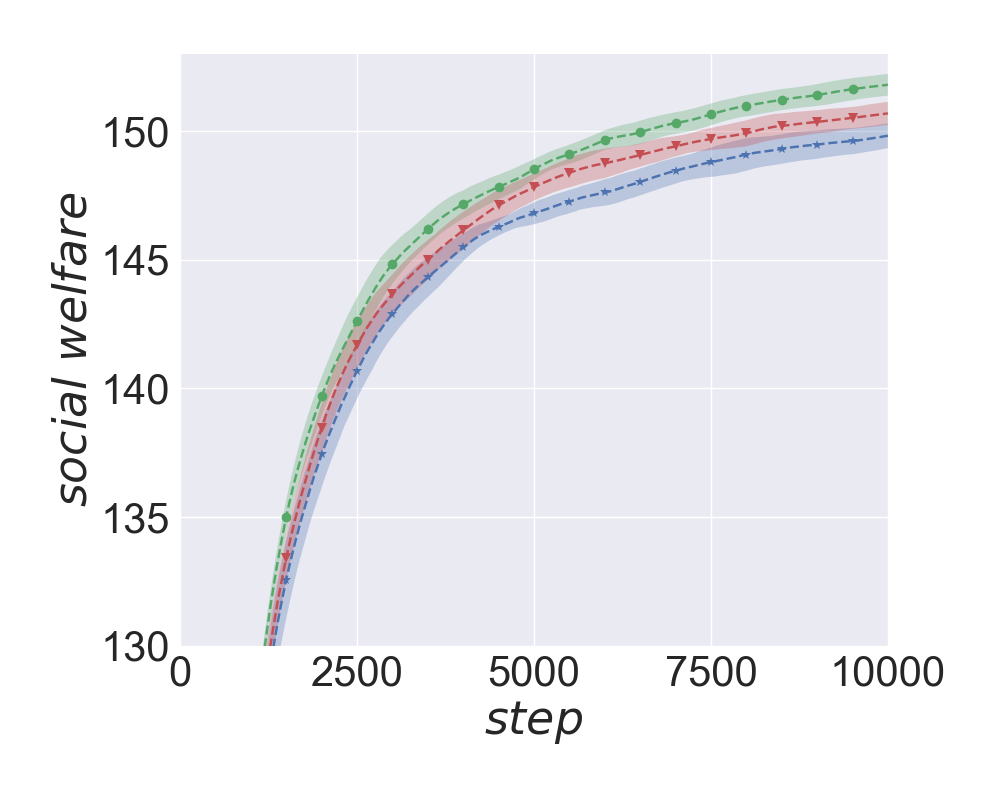}
}
\caption{Fruit Gathering game: $agent_2$'s policy varies between $\pi^c$ and $\pi^d$ every 100 steps.}
\label{gchange}
\end{figure}

\begin{figure}[H]
\centering
\subfigure{\includegraphics[height=1.225in,width=1.75in,angle=0]{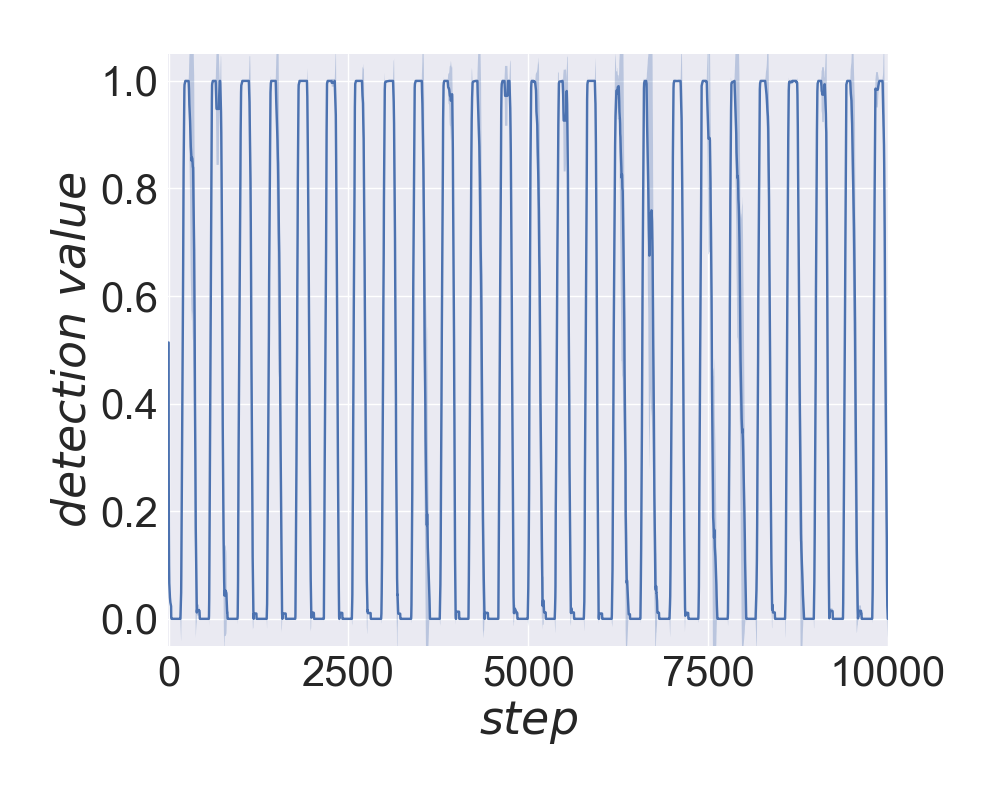}
\includegraphics[height=1.225in,width=1.75in,angle=0]{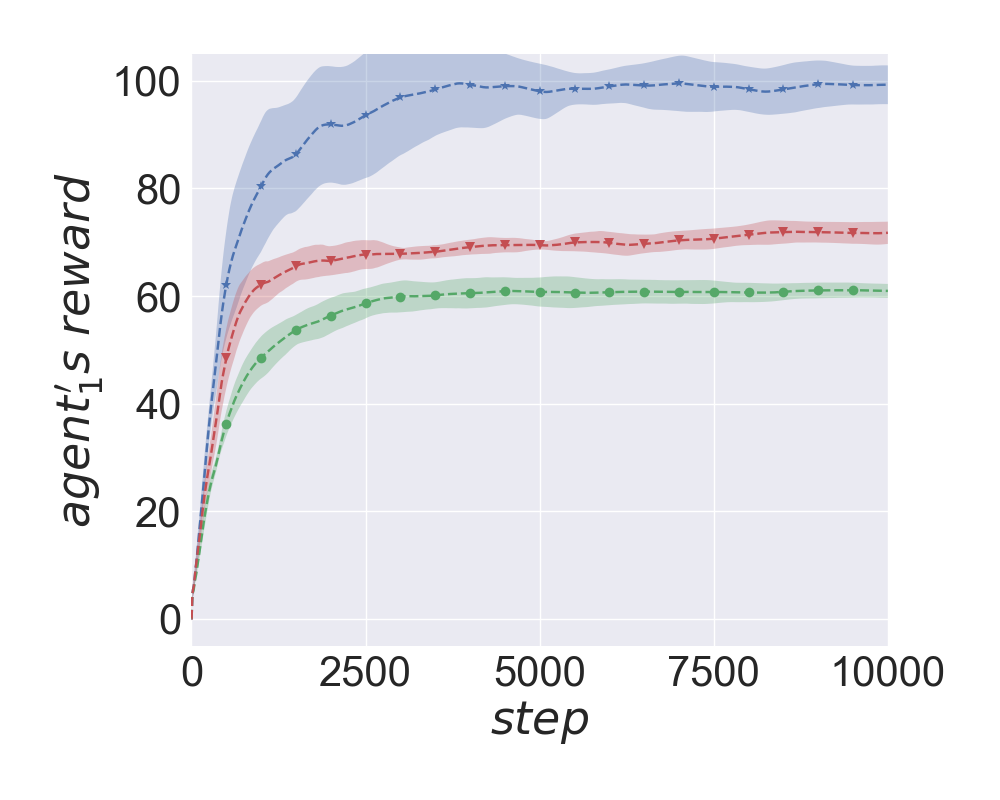}
\includegraphics[height=1.225in,width=1.75in,angle=0]{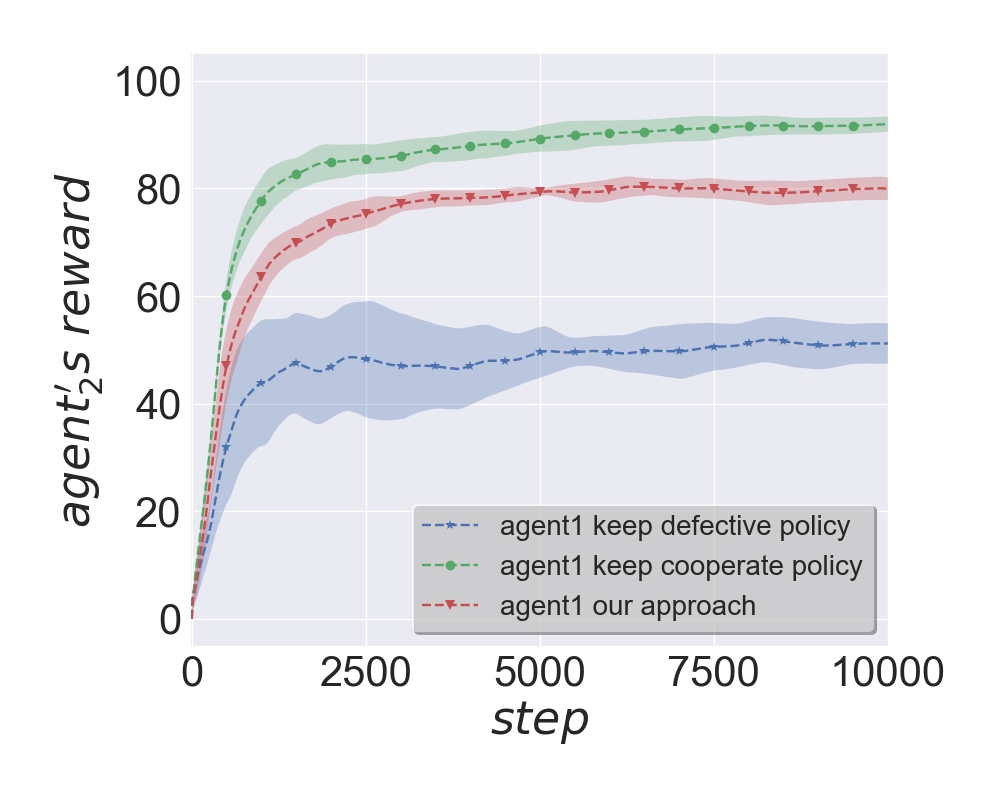}
\includegraphics[height=1.225in,width=1.75in,angle=0]{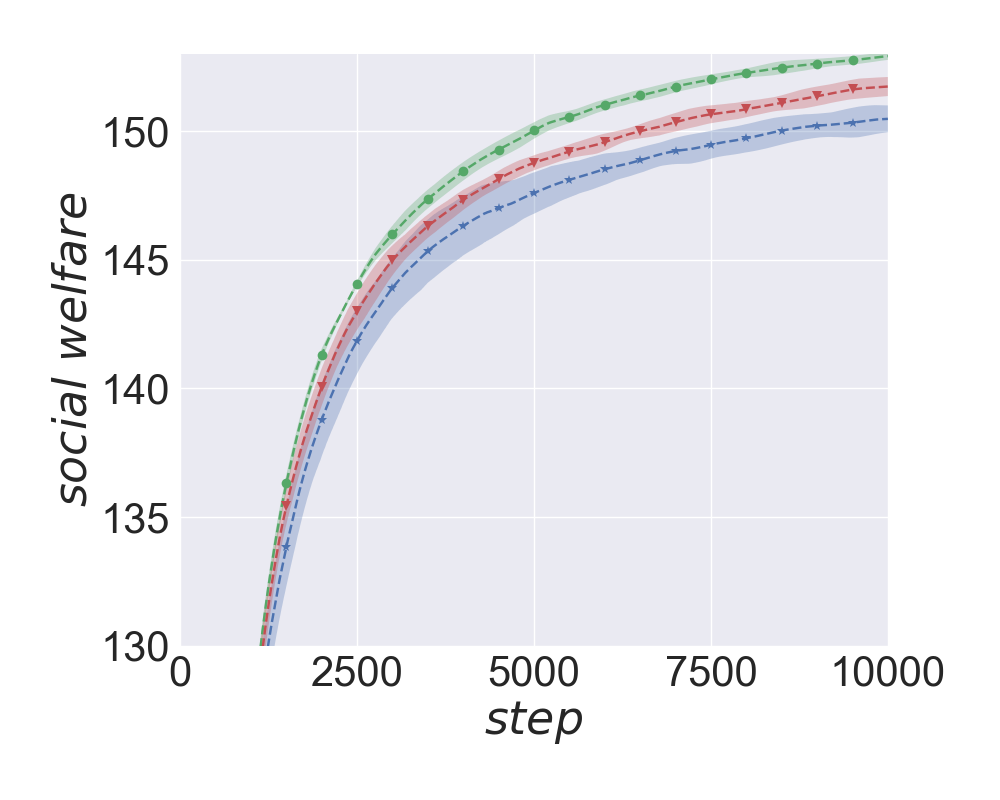}
}
\caption{Fruit Gathering game: $agent_2$'s policy varies between $\pi^c$ and $\pi^d$ every 200 steps.}
\label{gchange}
\end{figure}

\begin{figure}[H]
\centering
\subfigure{\includegraphics[height=1.225in,width=1.75in,angle=0]{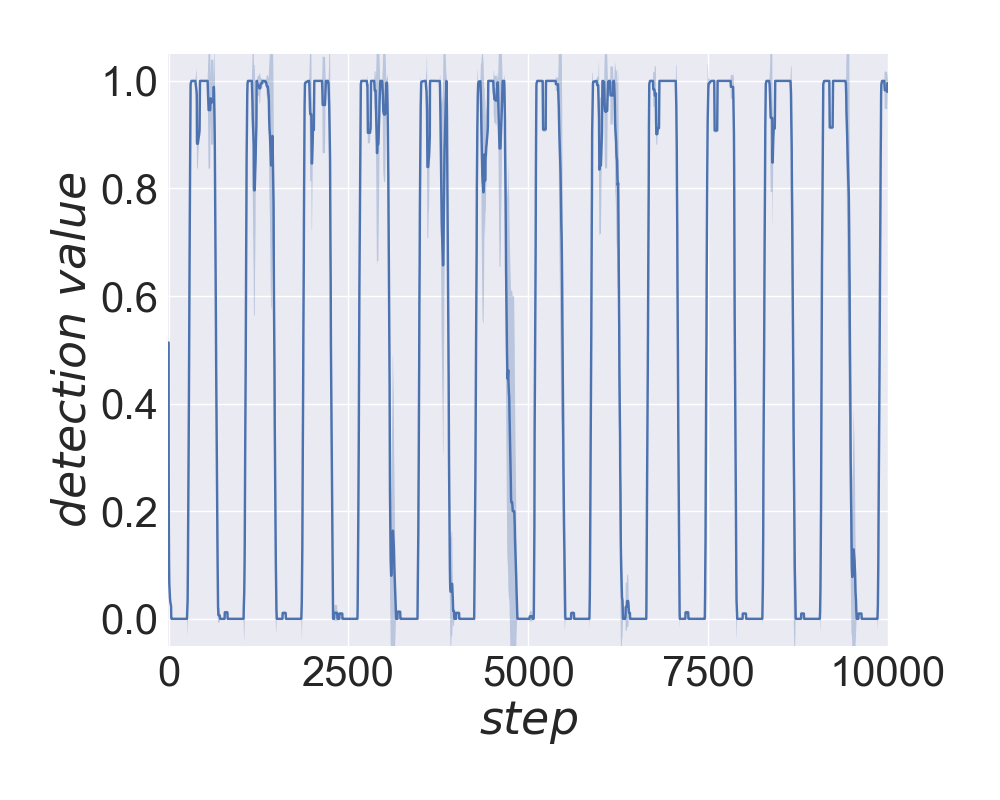}
\includegraphics[height=1.225in,width=1.75in,angle=0]{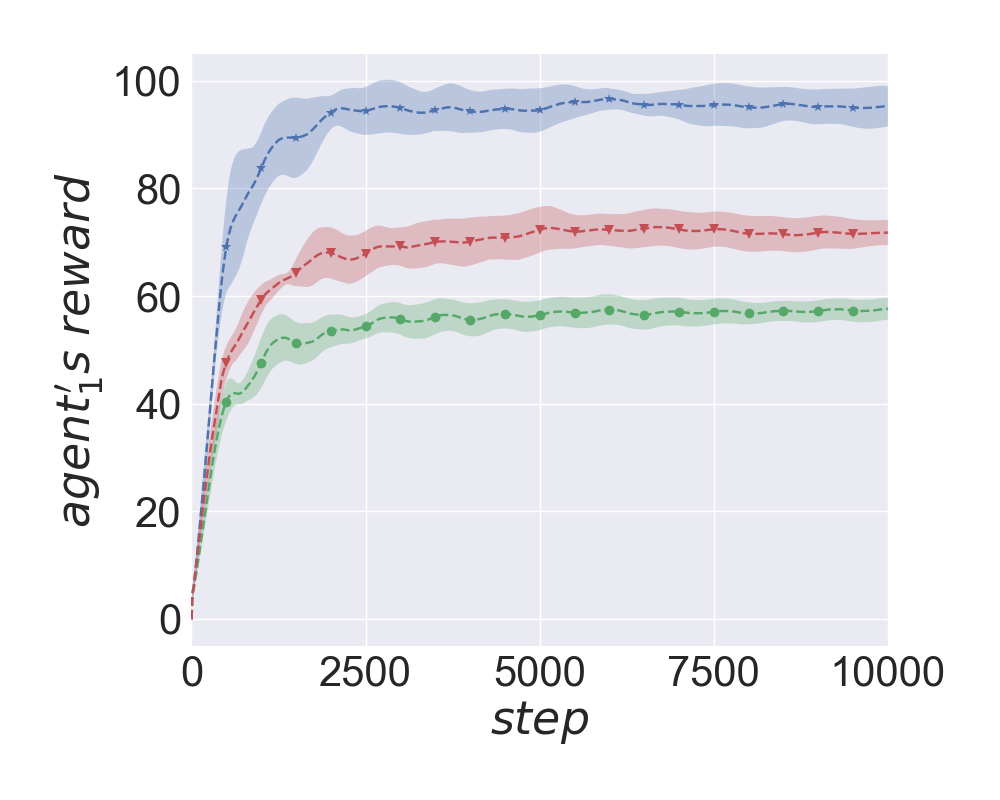}
\includegraphics[height=1.225in,width=1.75in,angle=0]{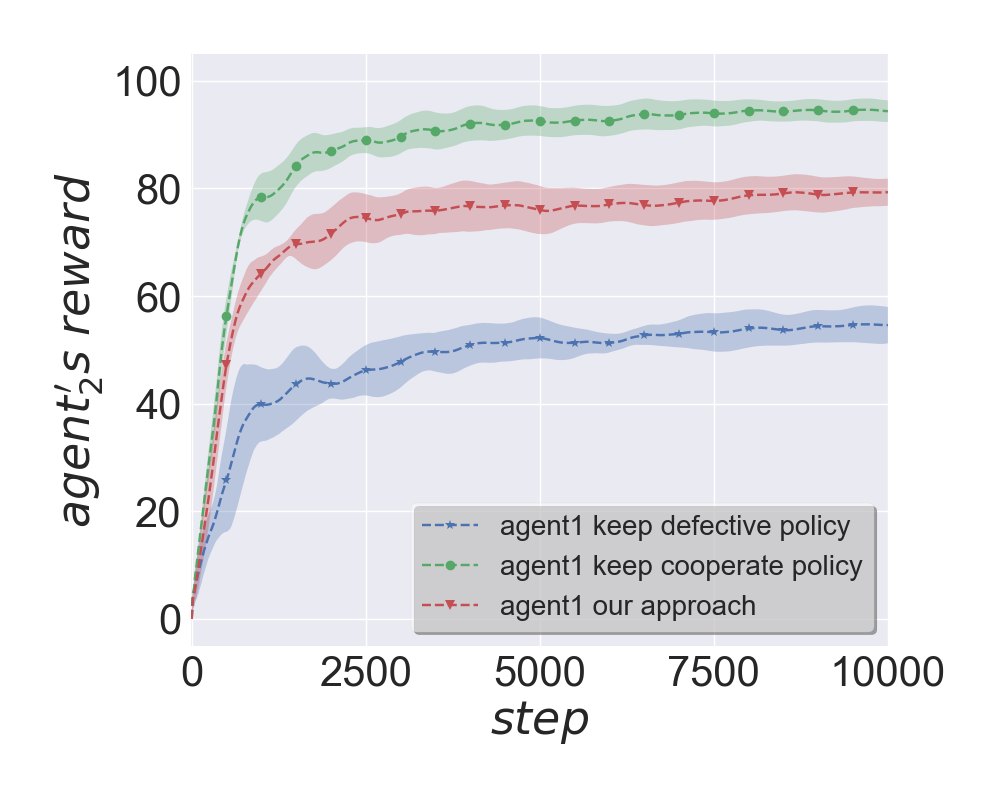}
\includegraphics[height=1.225in,width=1.75in,angle=0]{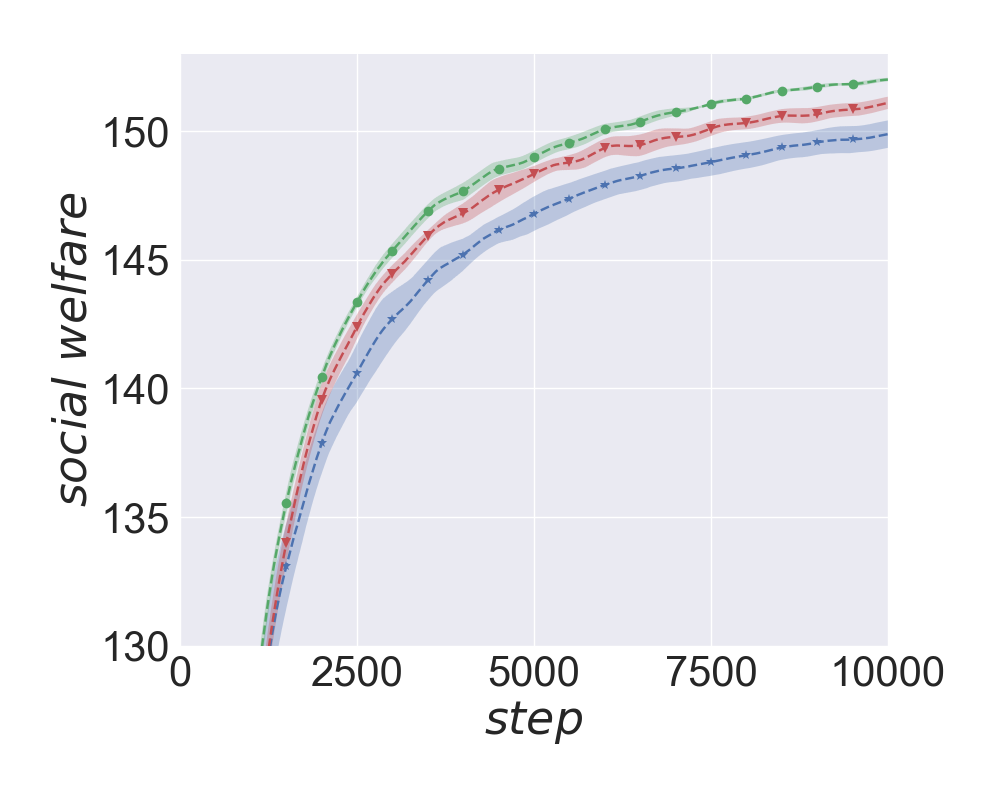}
}
\caption{Fruit Gathering game: $agent_2$'s policy varies between $\pi^c$ and $\pi^d$ every 300 steps.}
\label{gchange}
\end{figure}

\begin{figure}[H]
\centering
\subfigure{\includegraphics[height=1.225in,width=1.75in,angle=0]{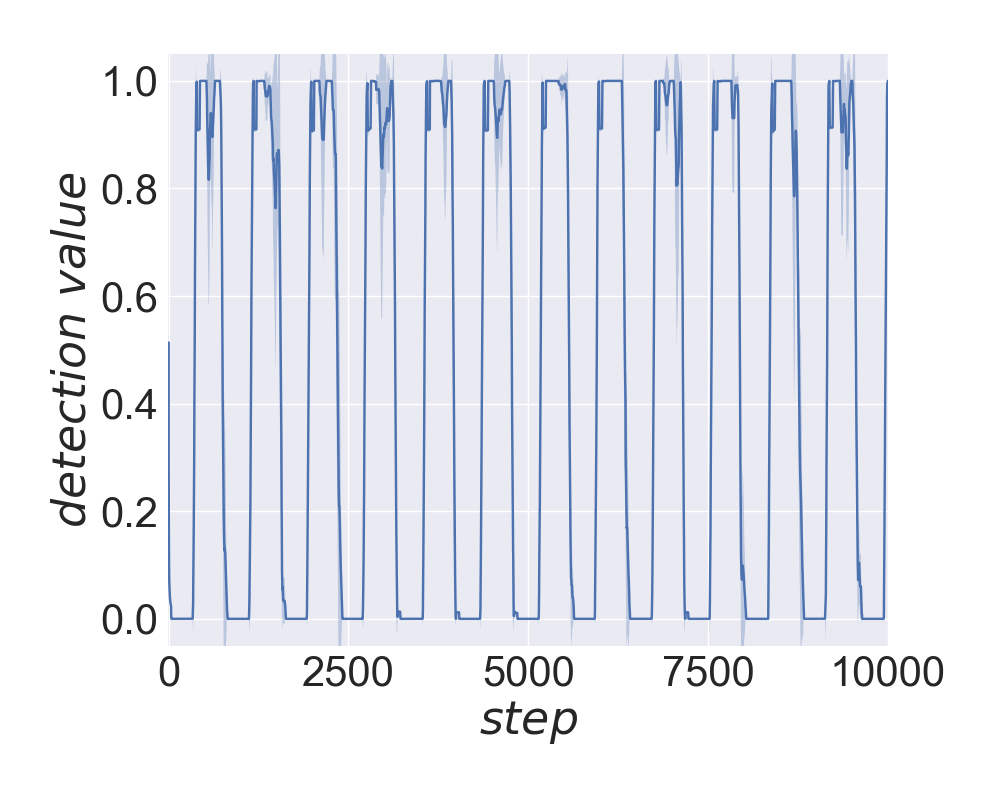}
\includegraphics[height=1.225in,width=1.75in,angle=0]{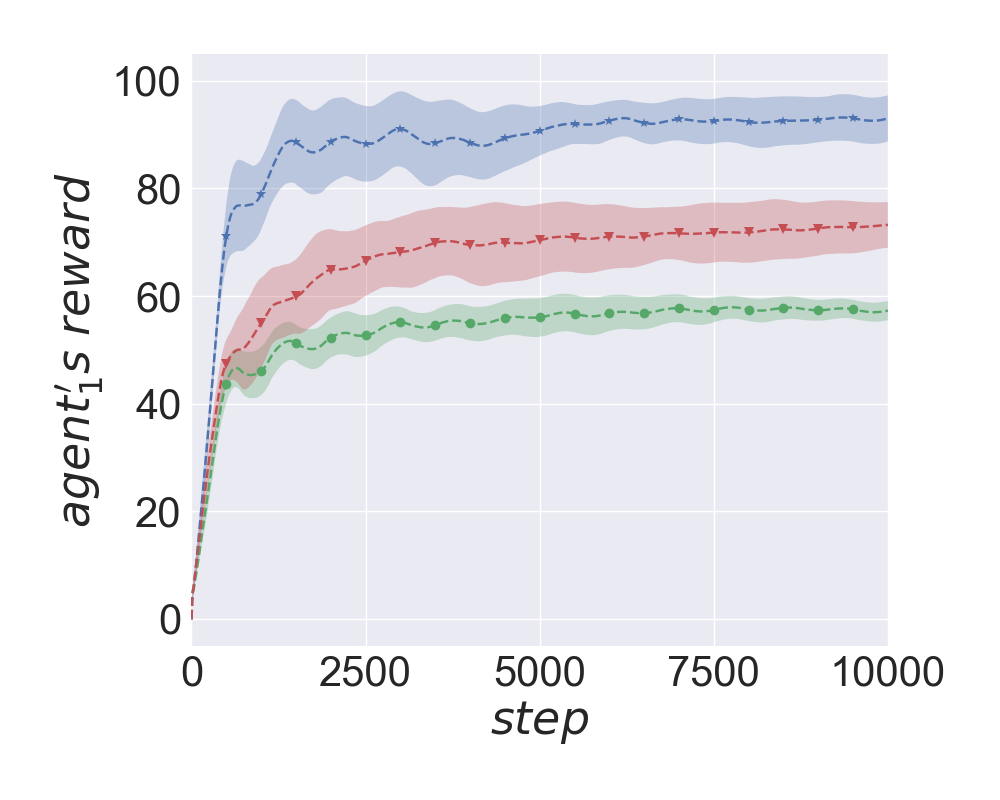}
\includegraphics[height=1.225in,width=1.75in,angle=0]{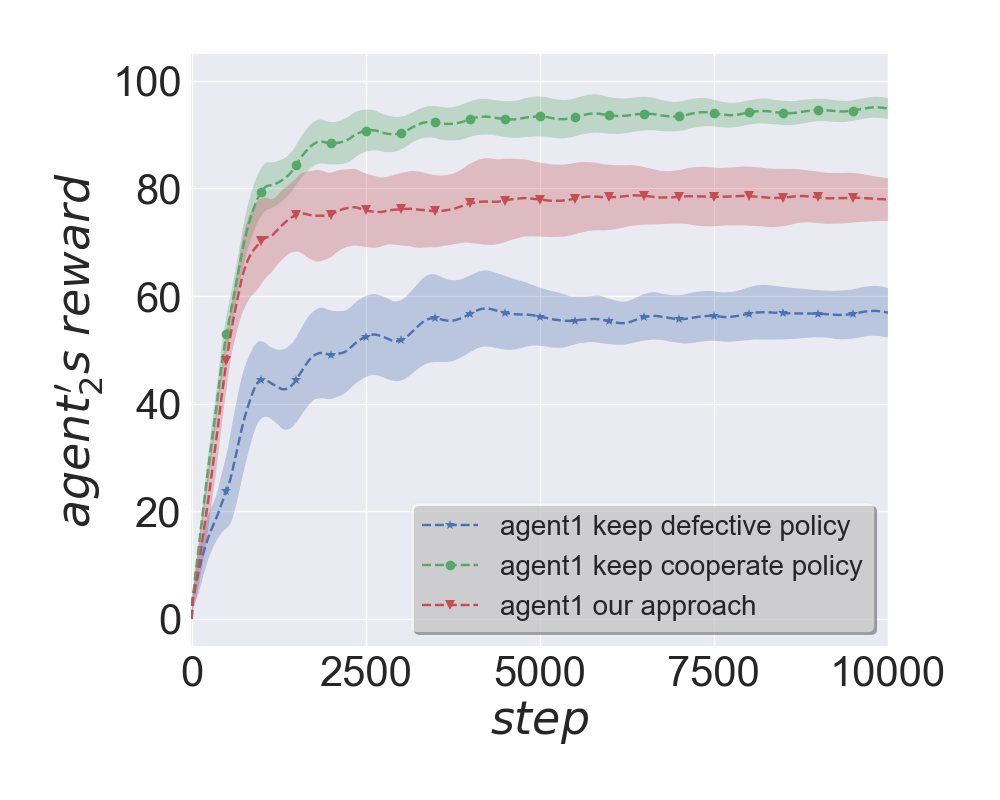}
\includegraphics[height=1.225in,width=1.75in,angle=0]{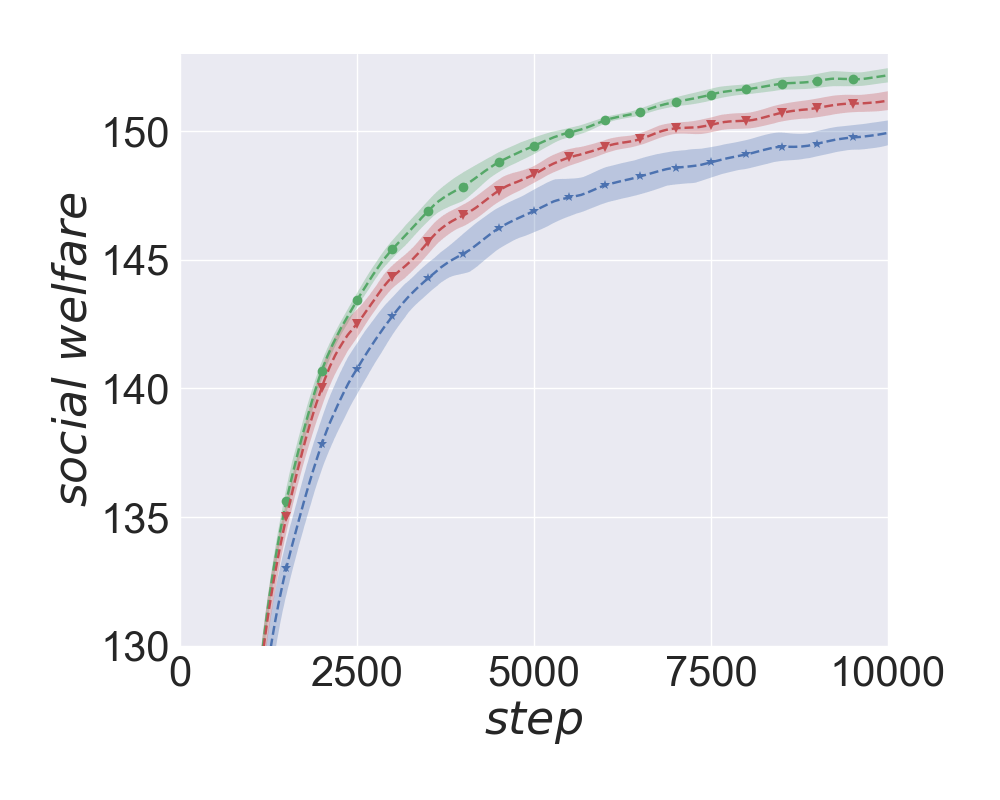}
}
\caption{Fruit Gathering game: $agent_2$'s policy varies between $\pi^c$ and $\pi^d$ every 400 steps.}
\label{gchange}
\end{figure}

\begin{figure}[H]
\centering
\subfigure{\includegraphics[height=1.225in,width=1.75in,angle=0]{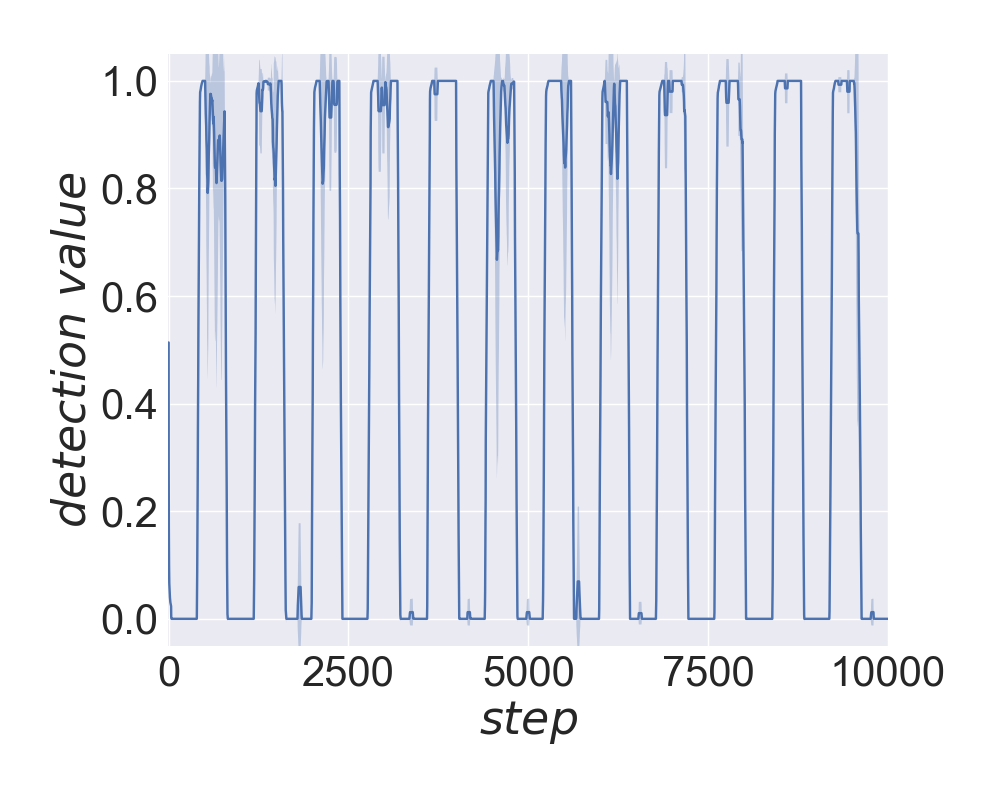}
\includegraphics[height=1.225in,width=1.75in,angle=0]{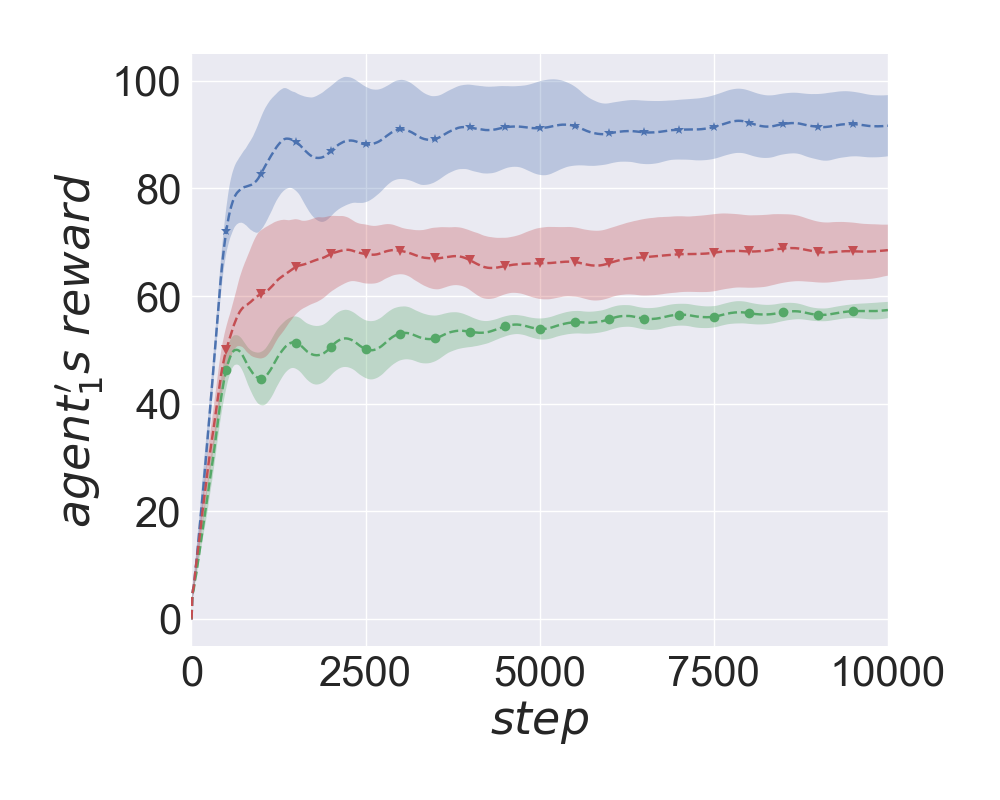}
\includegraphics[height=1.225in,width=1.75in,angle=0]{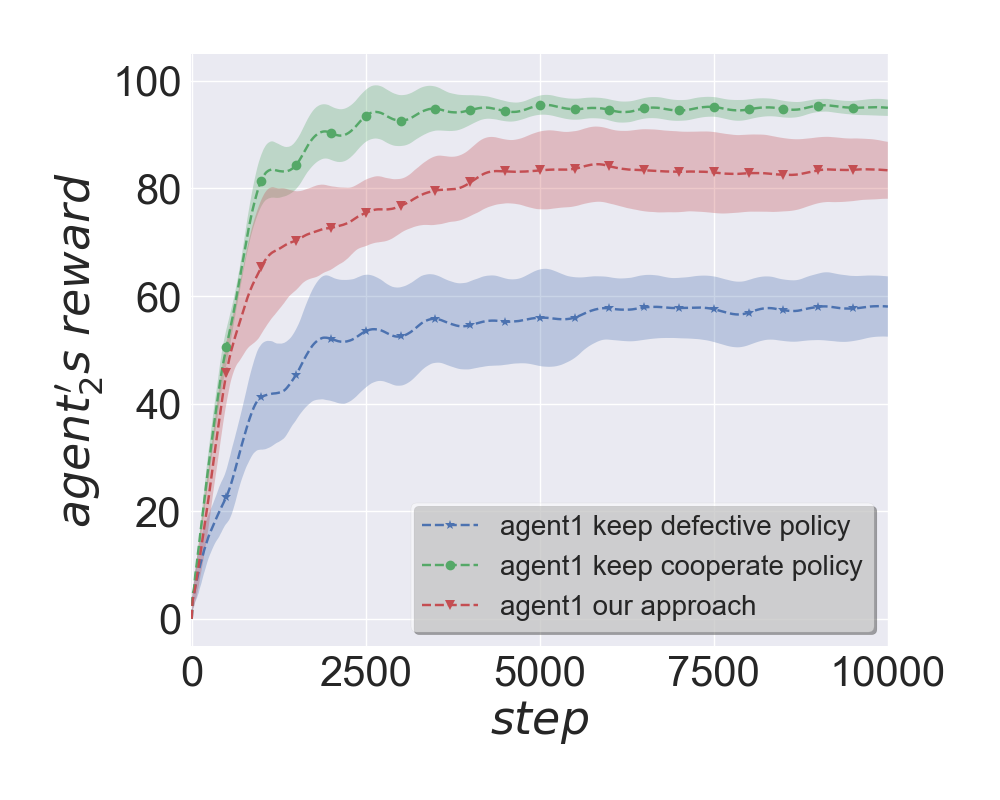}
\includegraphics[height=1.225in,width=1.75in,angle=0]{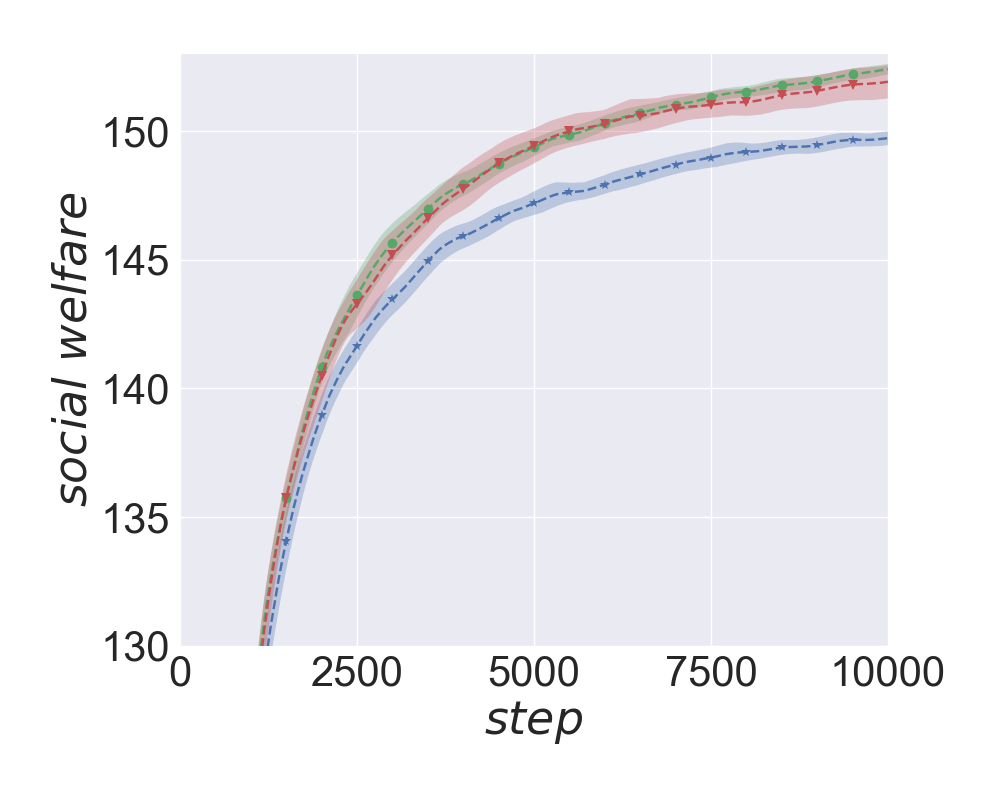}
}
\caption{Fruit Gathering game: $agent_2$'s policy varies between $\pi^c$ and $\pi^d$ every 500 steps.}
\label{gchange}
\end{figure}

\end{document}